\documentclass[review]{elsarticle}

\usepackage{hyperref}

\journal{Artificial Intelligence}

\usepackage{times}
\usepackage{latexsym}
\usepackage{amsmath}
\usepackage{multirow}
\usepackage{url}
\usepackage{graphicx}
\usepackage{mathrsfs}
\usepackage{subfigure}
\usepackage[table]{xcolor}
\usepackage{CJK}

\begin{document}

\begin{frontmatter}

\title{Encoding Implicit Relation Requirements for Relation Extraction: A Joint Inference Approach}


\author[mymainaddress]{Liwei Chen}
\ead{chenliwei@pku.edu.cn}

\author[mymainaddress]{Yansong Feng\corref{mycorrespondingauthor}}
\cortext[mycorrespondingauthor]{Corresponding author}
\ead{fengyansong@pku.edu.cn}

\author[mythirdaddress]{Songfang Huang}
\ead{huangsf@cn.ibm.com}

\author[mymainaddress]{Bingfeng Luo}
\ead{bf\_luo@pku.edu.cn}

\author[mymainaddress]{Dongyan Zhao}
\ead{zhaodongyan@pku.edu.cn}

\address[mymainaddress]{Peking University, 128 Zhong Guan Cun North Street, Haidian, Beijing, China}
\address[mythirdaddress]{IBM China Research Lab, Haidian, Beijing, China}

\begin{abstract}
Relation extraction is the task of identifying predefined relationship between entities,
and  plays an essential role in information extraction, knowledge base construction, question answering and so on.  Most existing relation
extractors  make predictions for each entity pair locally and individually, while ignoring implicit
global clues available across different entity pairs and in the knowledge base,
which often leads to conflicts among local predictions from different entity pairs. This paper
proposes a joint inference framework that employs such global clues to resolve disagreements among
local predictions.
We exploit two kinds of clues
to generate constraints which can capture the implicit type and cardinality requirements of a relation.
Those constraints  can be examined in either hard style or soft style, both of which can be effectively explored
 in an integer linear program formulation.
Experimental results on 
both English and Chinese datasets show that our proposed framework can 
effectively utilize those two categories of global clues
 and resolve the disagreements among local predictions, thus improve various relation extractors when such
clues are applicable to the datasets.
Our experiments also indicate that the clues learnt automatically from existing knowledge bases perform comparably to or better than those refined by human.

\end{abstract}


\end{frontmatter}


\section{Introduction}\label{sec_intro}

Identifying predefined relationship between pairs of entities
is crucial for many knowledge base related tasks, such as automatic knowledge base construction/population, factoid question answering, information extraction and so on, all of which have been popular topics in the field of artificial intelligence~\cite{AKBCworkshop}.
In the literature, the relation extraction task (RE) is usually investigated
in a classification style, where relations are simply treated as isolated class labels, while
their definitions or background information are sometimes ignored. Take the relation \textit{Capital} as an example,
we usually hold an agreement implicitly that this relation will expect a country as its subject and a city as object,
and in most cases, a city can be the capital of only one country.
In a knowledge base,  such clues about a relation's requirements are usually defined as the domain and range
of a relation, which  are no doubt helpful for improving relation extraction.
For instance, \cite{Yao:2010,zhang2013,koch-EtAl:2014:EMNLP2014} all explicitly modeled
the expected types of a relation's arguments with the help of a knowledge base's type taxonomy
and obtained promising results for various information extraction tasks.

However, properly capturing and utilizing such typing clues are not trivial.
One of the  hurdles here is the lack of off-the-shelf resources
and such clues often have to be coded by human experts.
Many knowledge bases do not have a well-defined typing
system, let alone fine-grained typing taxonomies with corresponding type recognizers,
which are crucial to explicitly model the typing requirements for arguments of a relation,
but rather expensive and time-consuming to collect.
Similarly, the cardinality requirements of arguments, e.g., a person can have
only one birthdate and a city can only be
labeled as capital of only one country, should be considered as a strong indicator
to eliminate wrong predictions, but has
to be coded manually as well. Therefore, except expensive explicit treatment, it would
  be a better choice to implicitly mine such requirements from data.

On the other hand, most previous relation extractors process each entity pair\footnote{we will use \textit{entity pair} and
\textit{entity tuple} exchangeably in the rest of the
paper} locally and individually, i.e., the extractor makes
decisions solely based on the sentences
containing the current entity pair and ignores other related pairs, therefore has difficulties to
capture possible disagreements among different entity pairs. 
However, when looking at the output of a multi-class relation
predictor globally, we can easily find possible incorrect predictions such as a university locates
in two different cities, two different cities have been labeled as capital for one country, a country
locates in a city and so on.
Suppose we have two entity pairs,  $<$Richard Fuld, USA$>$ and $<$USA, Washington$>$.
The extractor predicts that the relation between the first pair is $Nationality$ and the relation between the
second one is $LocationCity$. It is easy to find that one of the two predictions should be incorrect,
since the first prediction considers USA as a country but the second claims that USA locates in a city. We may also find such
disagreements inside an entity pair, e.g., the extractor may predict a relation $Capital$ for the
second entity pair according to another instance, which obviously disagrees with the previous prediction $LocationCity$.

Resolving those disagreements mentioned above can be considered as a kind of constrained optimization tasks,
e.g., optimizing the local predictions with global constraints among them, which can be captured by mining relation
background related clues from the knowledge base.
Based on those clues, we can discover the disagreements among the local predictions and generate constraints to resolve those conflicts.
In this paper, we will address how to derive and exploit two categories of clues:
\textbf{the expected types and the cardinality requirements of a
relation's arguments}, to improve the performance of relation extraction. 
We propose to perform joint inference upon multiple local predictions by leveraging implicit
clues that are encoded with relation specific requirements and can be learnt from existing knowledge bases. Specifically,
the joint inference framework operates on the output of a sentence level relation extractor as input,
derives different types of constraints from an existing KB to implicitly capture the expected
type and cardinality requirements for a relation's arguments, and
jointly resolve the disagreements among candidate predictions.
We obtain 5 types of constraints in total, where three of them correspond to the type requirements and the rest ones relate to the cardinality requirements. 
We formalize this procedure
as a constrained optimization problem, which can be solved by many
optimization frameworks.
The constraints we generated are also flexible to be utilized
in either hard style or soft style.
We use integer linear programming (ILP) as the solver and evaluate our framework on both English and Chinese datasets.
The experimental results show that our framework
can improve various relation extraction models, not only traditional feature based ones but also modern neural networks based.
We find that the global clues can be learnt automatically from a knowledge base with better or  comparable performance to
those refined manually, which can be effectively investigated in either hard style or soft style formulations.
Our framework
can outperform 
the state-of-the-art approaches when such clues are applicable to the datasets.

In the rest of the paper, we first review related work in Section~\ref{sec_relate}, and
in Section~\ref{sec_framework}, we describe our framework in detail. Experimental setup and
results are discussed in Section~\ref{sec_experiment}. We conclude this paper in Section~\ref{sec_conclusion}.

\section{Related Work}\label{sec_relate}
The task of relation extraction can be divided into two major categories: sentence-level (or mention-level) and entity pair level (or bag-level).
The former predicts what relation a pair of entities may hold within the given sentence, and
the latter  focuses on identifying the relationship between an entity pair based on one or multiple sentences
containing that entity pair. The entity pair level relation extraction task is more common in the literature, especially for
the knowledge base construction related tasks, such as knowledge base population (KBP) \cite{TAC-KBP} or automatically knowledge
base construction (AKBC) \cite{AKBCworkshop}.

Since traditional supervised relation extraction methods
\cite{Soderland:1995,Zhao:2005:ERI:1219840.1219892,chen2015ai} require manual annotations
and are often domain-specific, nowadays many efforts focus on
open information extraction, which can extract hundreds of thousands of relations from large scale of web texts
using semi-supervised or unsupervised methods\cite{Banko:2007:OIE:1625275.1625705,ReVerb,Wu:2010:OIE:1858681.1858694,qiu-zhang:2014:EMNLP2014,xu-EtAl:2013:NAACL-HLT2,Carlson10}.
However, these relations are often not canonicalized, therefore are difficult to be mapped to an existing KB.

Distant supervision (DS) is a semi-supervised relation extraction framework, which can automatically construct training data
by aligning the triples in a KB to the sentences which contain their
subjects and objects, and this learning paradigm has attracted much attention in information extraction tasks
\cite{Bunescu07learningto,Mintz:2009:DSR:1690219.1690287,Yao:2010,surdeanu2010tackbp,Hoffmann:2011:KWS:2002472.2002541,Mihai2012,zengEMNLP2015,linACL2016}.
DS approaches can predict canonicalized relations (predefined in a KB) for large amount of data and
do not need much human involvement. Since the automatically generated training datasets in
DS often contain noises, there are also research efforts focusing on
reducing the noisy labels in the training data \cite{Takamatsu:2012,xu-EtAl:2013:Short2}, or utilizing human annotated data to
improve the performance \cite{pershina-EtAl:2014:P14-2,angeli-EtAl:2014:EMNLP2014}.
Most of the above works put their emphasis on resolving or reducing the noises in the DS training data,
but mostly focus on the extraction models themselves, i.e., improving the local extractors,
while ignoring the inconsistencies among many local predictions.


As far as we know, few works have managed to take the relation specific requirements for arguments into
account implicitly, and most existing works make predictions locally and individually, lacking in global considerations
for inconsistency among local extractors.
The MultiR system
allows entity tuples to have more than one relations, but still predicts
each entity tuple locally \cite{Hoffmann:2011:KWS:2002472.2002541}.
\cite{Mihai2012} propose a two-layer multi-instance multi-label (MIML)
framework to capture the dependencies among relations. The first layer is
a multi-class classifier making local predictions for single sentences, the output of which
are aggregated by the second layer into the entity pair level. Their
approach only captures relation dependencies,
while we learn implicit relation backgrounds from knowledge bases,
including argument type and cardinality requirements.
\cite{wang-EtAl:2011:EMNLP2} construct a set of relation topics, 
and integrate them into a relation detector for better relation predictions. 
\cite{riedel13relation} propose to use latent vectors to estimate the
preferences between relations and entities. These can be considered
as the latent type information about the relations' arguments, which is learnt from various data sources.
In contrast, our approach can learn implicit clues from existing KBs, and jointly optimize
local predictions among different entity tuples to capture both relation argument type clues
and cardinality clues. \cite{zhang2013} utilize relation cardinality to create negative
samples for distant supervision while we use both implicit type clues and relation cardinality
expectations to discover possible inconsistencies among local predictions.
\cite{Yao:2010,zhang2013,koch-EtAl:2014:EMNLP2014} propose to explicitly use the type information
in distantly supervised relation extraction, which rely on existing typing resources and may have difficulties when the knowledge bases do not have fine types or
sophisticated named entity taggers. In contrast, we try to implicitly mine both type and cardinality clues from $<$subj, relation, obj$>$ triples without using fine typing tools or resources, which are then used to
resolve the disagreements among local predictions.
%

In recent years, neural networks (NN) based models, such as PCNN \cite{zengEMNLP2015}, have been utilized
in the relation extraction task, and the attention mechanism is also adopted to further reduce the noises within a sentence
bag (that is, all the sentences containing an entity pair) \cite{linACL2016}.
\cite{ye-EtAl:2017:Long1} exploit class ties between relations within one entity tuple, 
and obtain promising results.
However, those approaches
still pay less attention to exploiting the possible dependencies between relations globally among all entity pairs.
In contrast, our framework learns implicit clues from existing KBs, and jointly
 optimizes local predictions among different entity tuples to capture
both relation argument type clues and cardinality clues.
Specifically, this framework can lift various existing extractors,
including traditional extractors and NN extractors.

There are also works which first represent relations and entities as embeddings in a KB, 
and then utilize those embeddings to predict missing relations between any pair of entities  in the KB \cite{conf/semweb/KrompassBT15,DBLP:conf/aaai/XieLJLS16}. 
This task setup is different from ours, since we focus on extracting relations between entity pairs from the text resources, 
while they mainly make use of the structure information and descriptions of a KB to learn latent representations.

The idea of global optimization over local predictions has been proven to be helpful  in other  information extraction tasks.
\cite{liji2011} and \cite{liji2013} use co-occurrence statistics among relations or events to
jointly improve information extraction performances in ACE tasks, while we mine existing knowledge bases to collect
global clues to solve local conflicts and find the optimal aggregation assignments, regarding
existing knowledge facts. There are also works which encode general domain knowledge as first order logic rules
in a topic model \cite{conf/emnlp/LacalleL13}. The main differences between their approach and our work are that our global clues can be collected  from knowledge bases and
our instantiated constraints are directly operated in an ILP model.

Most existing works in distantly supervised relation extraction focus on improving the performance locally, including
designing sophisticated traditional or neural network models, incorporating explicit type information, 
reducing  noisy instances in the training data, or utilizing extra
human annotated data, etc.
Different from previous works, we propose to implicitly exploiting the
relation requirements to discover the disagreements among the unreliable local predictions, and formalize the procedure
as a constrained optimization problem, which can resolve those disagreements and achieve a globally optimal assignment.

\section{The Framework}\label{sec_framework}
\begin{figure}[htp]
\centering
\includegraphics[width=12cm]{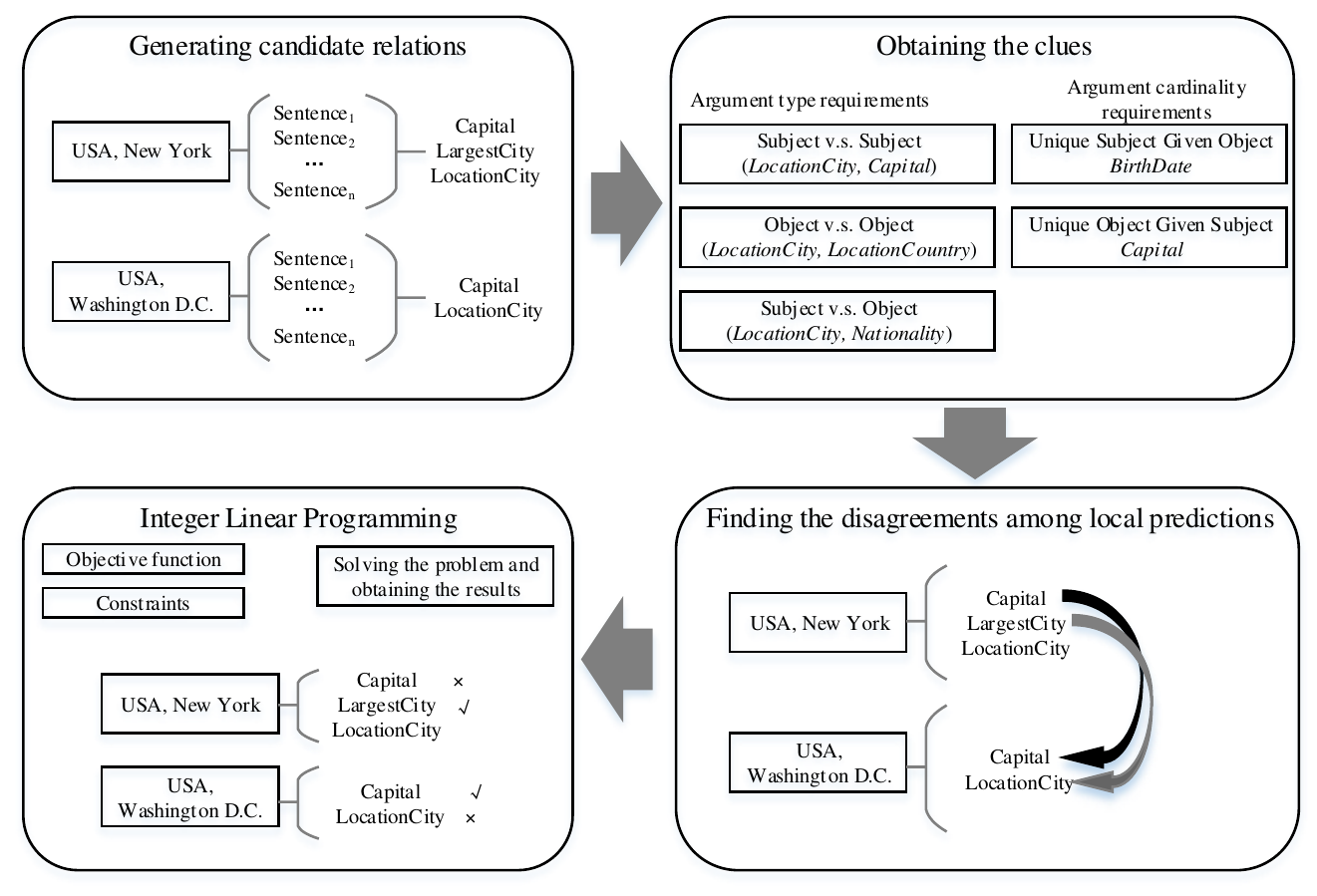}
\caption{An overview illustration of our framework. The framework includes four main steps: generating the candidate relations, obtaining the clues, finding the disagreements among the local predictions,
and using an integer linear programming framework to solve the problem and obtain the final results.}\label{fig:framework}
\end{figure}

Our framework takes a set of entity pairs and their supporting
sentences as its input.
We first train a preliminary sentence level extractor which can output confidence
scores for its predictions, e.g., a maximum entropy or a neural network based model,
and use this local extractor to produce local predictions.
In order to implicitly capture the expected type and cardinality requirements for a relation's arguments,
we derive two kinds of clues from an existing KB, which are further
utilized to discover the disagreements among local candidate predictions.
Our objective is to maximize the overall confidence of
all the selected predictions, as well as to minimize the
inconsistencies among them. Figure \ref{fig:framework} shows an overview of our proposed framework.
%

\subsection{Obtaining Candidate Relations} \label{sec_candidate}
Since we will focus on extracting  relations predefined in a structured KB, we
follow the distant supervision paradigm to collect our training data
guided by this KB, and train a local extractor accordingly,
e.g.,  using an maximum entropy model, a neural network model,
or other existing extractors. 
Given a sentence containing an entity pair $t$ (from a set of entity tuples $\mathcal{T}$),
the model will output the confidence
of this sentence representing certain relationship $r$ (from a predefined relation set $\mathcal{R}$)
between the entity pair.
%
Note that, due to the noisy nature of the training data, we admit that
the resulting prediction is not fully reliable. But we still have a good chance to
find the correct prediction ranked in the second or third,
we thus select \textbf{top three} predictions as the
candidate relations for each entity pair in order to introduce more
potentially correct output.


On the other hand, we should discard the predictions whose
confidences are too low to be true. We set up a threshold of 0.1.
For an entity tuple $t$, we obtain its candidate relation set by combining
the candidate relations of all its sentence-level mentions, and represent it as $R^t$.
For a candidate relation $r \in R^t$ and a tuple $t$,
we define $M^r_t$ as all $t$'s sentence-level
mentions whose candidate relations contain $r$.
Now the confidence score of a relation $r \in R^t$ being
assigned to tuple $t$ can be calculated as:
\begin{equation}\label{eq:confidence}
conf(t,r) = \sum\limits_{m \in M^r_t} {score(m,r)}
\end{equation}
where $score(m,r)$ is the confidence of a mention $m$ representing relation $r$
output by our preliminary extractor.

Traditionally, both lexical features and syntactic features have been investigated
 in the relation extraction task.
Lexical features are the word chains between the subjects and
objects in the sentences, which are usually too specific to frequently appear in
the test data.
For instance, a long lexical feature "\textit{PERSON was an American politician who was born
on August 19, 1946 in LOCATION}" extracted from the training set may be unlikely
to exist in any sentence of the testing set, thus might be useless in predicting
the relation \textit{BirthPlace} in the future. On the other hand, the reliability of syntactic
features depends heavily on the quality of dependency parsing tools, which
may limit the usage of such kind of features in the languages which do not have
high quality parsing toolkits.

Generally speaking, we expect more potentially correct relations to be put into
the candidate relation set for further consideration, i.e., we expect
recall-oriented preliminary local extractors.
So,
in addition to lexical and syntactic features,
traditional feature based extractors can benefit from incorporating n-gram features,
which are considered as more ambiguous and bring higher recall.
In the neural network category, 
 the NN extractors\cite{zengEMNLP2015,linACL2016} take advantage of word embeddings and convolutional architectures
to naturally support recalling more potentially correct results.

\subsection{Disagreements among the Candidates} \label{sec_constraint}
The candidate relations we obtained from local extractors inevitably include
incorrect predictions, since they are often predicted individually, e.g., within one entity pair. 
One way to identify those incorrect predictions is to find any
 disagreements among local predictions,
which could be resolved by discarding the local predictions that lead to the disagreements. 
 This can help us to obtain more accurate predictions, with more global coherence. 
Therefore, our clues should be in a form of \textit{A-and-B-should-not-happen-together}. Theoretically, our framework 
can deal with any sort of clues in such a form. In this paper, we will discuss two kinds of 
them in detail.

As discussed earlier, we will exploit from the knowledge base two categories of clues that implicitly capture
relations' backgrounds: their expected argument types and argument cardinalities, based on which
we can discover two categories of disagreements among the candidate predictions,
summarized as \textbf{argument type inconsistencies} and \textbf{violations of arguments'
uniqueness}, which have been rarely investigated before.
Next, we will discuss them in detail, and describe how to
learn the clues from a KB afterwards.

\begin{figure}
\centering
\includegraphics[width=0.9\textwidth]{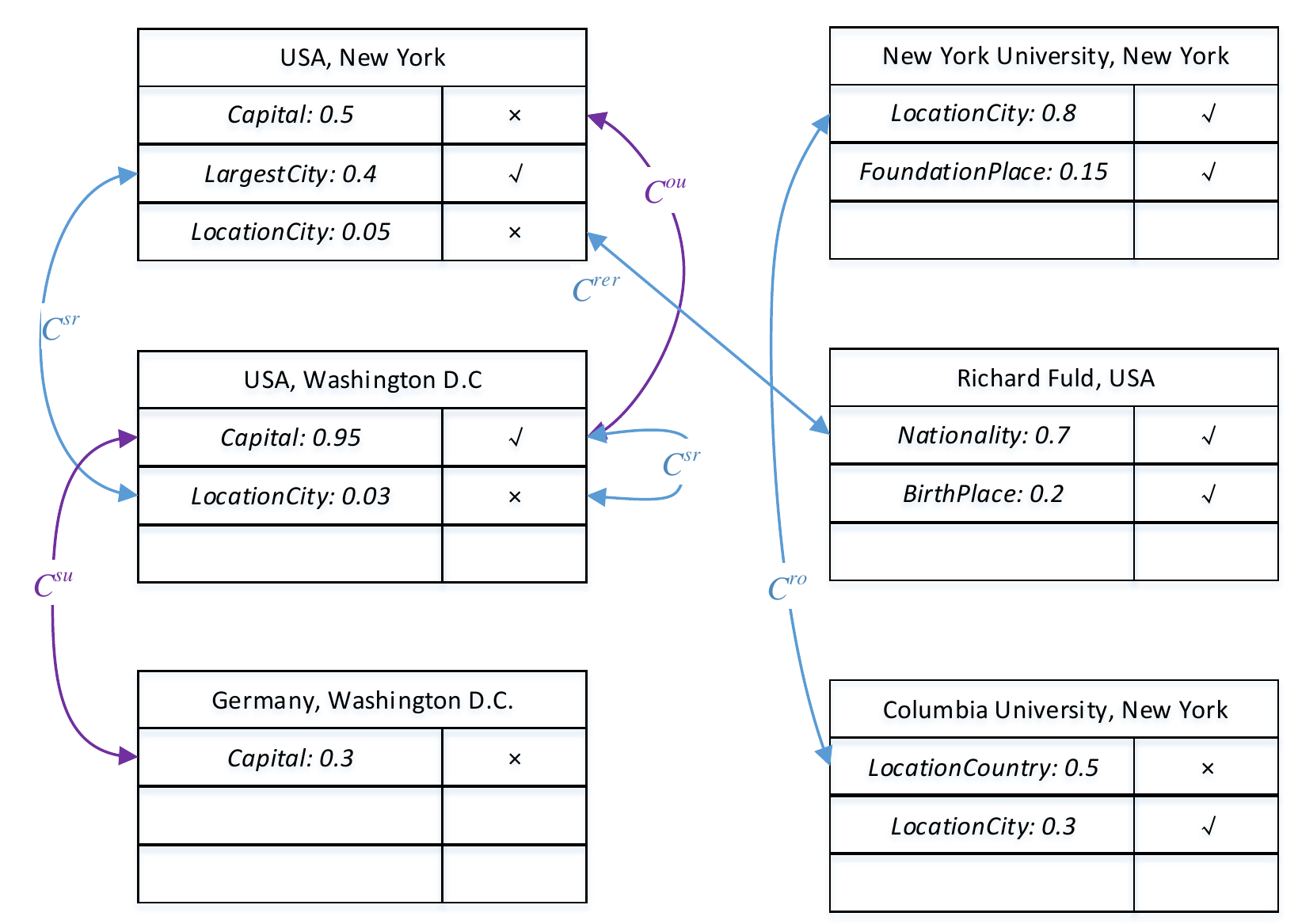}
\caption{Different types of disagreements we investigate among the candidate relations. Each arrow link two local relations that conflict with each other. The color of an arrow indicates the type of the clue we use to discover the disagreement, and the caption of an arrow shows its subtypes.}
\label{fig:conflicts}
\end{figure}

\subsubsection{Implicit Argument Types Inconsistencies: }
Generally, the argument types of correct predictions should be consistent with each other. Given a
relation, its arguments sometimes are required to be certain types of entities.
If the local predictions among different
entity tuples require the same entity to belong to different and \textit{contradictory} types, we
call this situation as argument type inconsistency. Take $<$USA, New York$>$
and $<$USA, Washington D.C.$>$ as an example. In Figure~\ref{fig:conflicts},
$<$USA, New York$>$ has a candidate relation \textit{LargestCity} which restricts \textit{USA} to be
either countries or states, while $<$USA, Washington D.C.$>$ has a prediction
\textit{LocationCity} which requires an organization as its subject.
These two local predictions lead to a disagreement in terms of \textit{USA}'s
type because the latter expects \textit{USA} to be an organization located in a city,
which warns that at least one of the two candidate relations is incorrect.

Besides the disagreement between the subjects of two candidate relations,
from Figure~\ref{fig:conflicts},
we can observe two more situations: the first one is that the objects
of the two candidate relations are inconsistent with each other, for example
$<$New York University, New York$>$ with the prediction \textit{LocationCity}
and $<$Columbia University, New York$>$ with the prediction \textit{LocationCountry}.
The second one is that the subject of one candidate relation does not agree with another
prediction's object, for example $<$Richard Fuld, USA$>$ with the prediction
\textit{Nationality} and $<$USA, New York$>$ with the prediction \textit{LocationCity}.
Although we have not assigned explicit types to these entities, we can still exploit
the inconsistencies implicitly with the help of shared entities.
Note that the implicit argument typing clues here mean whether two relations
can share arguments, but \textbf{not} enumerate what types explicitly their arguments should have.

Formally, these problematic
relation pairs can be divided into the following three subcategories.
We represent the relation pairs $(r_i,r_j)$ that are inconsistent in terms of subjects
as $\mathcal{C}^{sr}$, the relations pairs that are inconsistent in terms of objects
as $\mathcal{C}^{ro}$, the relation pairs that are inconsistent in terms of one's subject
and the other one's object as $\mathcal{C}^{rer}$.

It is worth mentioning that
disagreements inside an entity tuple are also considered here. For instance,
an entity tuple $<$USA, Washington D.C.$>$ in Figure~\ref{fig:conflicts} has two candidate relations,
\textit{Capital} and \textit{LocationCity}. These two predictions are inconsistent with each other
with respect to the type of \textit{USA}. They implicitly consider \textit{USA} as ``country'' and
``organization'', respectively.

\subsubsection{Violations of Arguments' Uniqueness: }
The second kind of disagreement regards the argument cardinality requirements for a relation.
Given a subject, some relations should have unique objects. For
example, in Figure~\ref{fig:conflicts}, given \textit{USA} as the
subject of the relation \textit{Capital}, we can only accept one possible object,
because there is great chance that a country only have one capital. On the other hand,
given \textit{Washington D.C.} as the object of the relation \textit{Capital}, we can only accept
one subject, since usually a city can only be the capital of one country or
state. 
If these are violated in the candidates, we could know that there may be some
incorrect predictions.
We represent the relations expecting unique objects as $\mathcal{C}^{ou}$, and
the relations expecting unique subjects as $\mathcal{C}^{su}$.

\subsubsection{Other Types of Clues}
 In addition to the two types of clues discussed above, there could be some other types 
of clues, in the form of  \textit{A-and-B-should-not-happen-together}, applicable to this task. 
For example, one may design a kind of clues with respect to the numerical values of different
relations' arguments, e.g., one person's \textit{BirthDate} should be earlier than
his \textit{DeathDate}, which can be easily transformed to a constraint regarding 
whether two local predictions can be correct at the same time. 


Such type of clues are surely useful if they are applicable to the datasets.
However, there are few such cases in our current datasets,
we will leave further investigations as an interesting direction for future work.

\subsection{Obtaining the Global Clues}\label{conflict_gen}
Now, the issue is how to obtain the clues used in the previous subsection.
That is, how we determine which relations
expect certain types of subjects, which relations expect certain types of objects, etc.
These knowledge can be definitely coded by human, or learnt from a KB.

Most existing knowledge bases represent their knowledge facts in the form of  ($<$\textit{subject, relation, object}$>$) triple, which
can be seen as relational facts between entity tuples.
Usually the triples in a KB are carefully defined by experts. It
is rare to find inconsistencies among the triples in the knowledge base.
The clues are therefore learnt from KBs, and further refined manually if needed.

Given two relations
$r_1$ and $r_2$, we query the KB for all tuples
bearing the relation $r_1$ or $r_2$. We use $S_i$ and $O_i$ to represent
$r_i$'s ($i \in \{1,2\}$) subject set and object set, respectively.
We adopt a modified Kulczynski similarity coefficient ($\mathcal{K}$)~\cite{kulczynski1927} to
estimate the dependency between the argument sets of two relations:
\begin{equation}
\mathcal{K}(A,B) = \log \frac{1}{2}(\frac{{count(A \cap B)}}{{count(A)}} + \frac{{count(A \cap B)}}{{count(B)}})
\end{equation}
where $count(A \cap B)$ is number of the entities in both $A$ and $B$,
$count(A)$ and $count(B)$ are the numbers of the entities in $A$ and $B$, respectively.
For any pair of relations from $\mathcal{R} \times \mathcal{R}$, we calculate four scores:
$\mathcal{K}(S_1,S_2)$, $\mathcal{K}(O_1,O_2)$, $\mathcal{K}(S_1,O_2)$ and $\mathcal{K}(S_2,O_1)$.
To make more stable estimations, we set up a threshold $\kappa$ for $\mathcal{K}()$.
If $\mathcal{K}(S_1,S_2)$ is lower than the threshold, it means there is a
good chance that $r_1$ and $r_2$ cannot share a subject, and we should generate a clue for them.
Otherwise, we will consider that $r_1$ and $r_2$ can share a subject.
Things are similar for the other three scores. The threshold is set to -3 in this paper.

We can also learn the uniqueness of arguments for relations. For each pre-defined
relation in $\mathcal{R}$, we collect all the triples containing this
relation, and count the portion of the triples which only have one object for
each subject, and the portion of the triples which only have one subject for
each object. The relations whose
portions are higher than the threshold will be considered to have unique
argument values. This threshold is set to 0.8 in this paper.

One issue for the learning strategy is  when a relation has only a small
amount of instances in the KB, the learnt clues about that relation can be inevitably noisy,
which may bring errors to the final predictions.
Here, we should take different application scenarios into account, to
 balance the human crafted clues and the automatically obtained ones.
For example, during the initial stage of  knowledge base population, i.e., when the KB is still small,
we can first utilize manually crafted clues for new relations and existing relations
with only a small number of instances. As we accumulate more and more relation instances
and populate them into the KB, we can move onto the automatically learnt clues.

\subsection{Integer Linear Program Formulation}
As discussed above, given a set of entity pairs and their candidate
relations output by a preliminary extractor, our goal is to find an optimal
configuration for all those entities pairs jointly, solving the
disagreements among those candidate predictions and maximizing the overall
confidence of the selected predictions. It has been proven that this problem
is the reduction from an NP-complete problem, minimum vertex cover, which means it is an NP-hard problem\footnote{https://en.wikipedia.org/wiki/Integer\_programming}.  
Many optimization models can be used to obtain the approximate solutions.
In this paper, we propose to solve the problem by using an ILP
tool, IBM ILOG Cplex\footnote{www.cplex.com}.

\subsubsection{Objective Function}\label{sec:objectiveFunc}
For each tuple $t$ and one of its candidate relations $r$,
we define   a binary decision variable $d_t^r \in \{0,1\}$, 
indicating whether the candidate relation $r$ is selected by the solver.
Our objective is to maximize the total confidence of all the selected candidates,
and the objective function can be written as:
\begin{eqnarray}\label{objective}
\max \sum\limits_{t \in {\cal T},r \in {R^t}} {(c} onf(t,r) + \mathop {\max }\limits_{{\rm{m}} \in M_t^r} score(m,r))d_t^r
\end{eqnarray}
where $conf(t,r)$ is the confidence of the tuple $t$ bearing the candidate relation $r$,
$M^r_t$ is the set of $t$'s sentence-level mentions whose candidate relations contain $r$.
The first component is
the original confidence scores of all the selected candidates, and the second one is
the maximal mention-level confidence scores of all the selected candidates.
The latter is designed to
encourage the model to select the candidates with higher individual mention-level confidence scores,
which, to some extent, coincides with the distant supervision assumption. 
For example, in an entity pair $t$, 5 mentions are predicted as $r_1$ 
with 0.2 confidence score by the local extractor, and another mention is 
predicted as $r_2$ with a confidence 0.9, we would like to encourage the model 
to select the latter.

\subsubsection{Constraint Generation}\label{sec_cons_gen}
Now we formulate the clues we have collected into constraints to help resolve the disagreements
among the candidate relations. Note that the formulation can be in either hard style or soft style. 

The hard-style formulation will generate hard constraints based on those clues, which means
if there are two predictions violating the constraints, we will have to discard the one
which may result in a smaller objective, e.g., the one with lower confidence.
For example, given two predictions which indicate that a country is also an organization
locating in a city, there is a good chance that one of the predictions is incorrect and
should be eliminated.

On the other hand, the soft-style formulation allows a constraint to be violated, to some extent, with a
continuous value as a kind of penalty. This is designed under the observation
that  some of the clues we have generated can be violated in certain situations. For example,
in most cases an actor will not be a politician, but \textit{Arnold Schwarzenegger} is both an actor and a politician.
Thus, allowing the constraints to be violated with certain penalty can deal with such rare situations,
and potentially avoid eliminating true positive samples according to one strict hard constraint.
Intuitively, one should expect a higher penalty for very rare situations, and a smaller penalty for a constraint with several exceptions.
%

\subsubsection{Hard Style Formulation}\label{sec:hard}
Here we describe how to formally take into consideration the two kinds of disagreements, i.e., \textit{implicit argument types inconsistencies}
and \textit{violations of arguments' uniqueness}, in a hard style.
For the sake of clarity, we describe the
constraints derived from each scenario of these two disagreements separately.

The subject-relation constraints avoid the disagreements
between the predictions of two tuples sharing a subject.
These constraints can be represented as:
\begin{eqnarray}\label{constraint_subjrel}
d_{{t_i}}^{{r^{t_i}}} + d_{{t_j}}^{{r^{t_j}}} \le 1 \\
\forall t_i,t_j: subj(t_i) = subj(t_j) \wedge (r^{t_i}, r^{t_j}) \in \mathcal{C}^{sr} \nonumber
\end{eqnarray}
where $t_i$ and $t_j$ are two tuples in $\mathcal{T}$, $subj(t_i)$
is the subject of $t_i$, $r^{t_i}$ is a candidate relation of $t_i$,
$r^{t_j}$ is a candidate relation of $t_j$.

The object-relation constraints avoid the inconsistencies
between the predictions of two tuples sharing an object.
Formally we add the following constraints:
\begin{eqnarray}\label{constraint_subjrel}
d_{{t_i}}^{{r^{t_i}}} + d_{{t_j}}^{{r^{t_j}}} \le 1 \\
\forall t_i,t_j: obj(t_i) = obj(t_j) \wedge (r^{t_i}, r^{t_j}) \in \mathcal{C}^{ro} \nonumber
\end{eqnarray}
where $t_i \in \mathcal{T}$ and $t_j \in \mathcal{T}$ are two tuples, $obj(t_i)$
is the object of $t_i$.

The relation-entity-relation constraints ensure that if an entity
works as subject and object in two tuples $t_i$ and $t_j$, respectively,
their relations should agree with each other. The constraints can be designed as:
\begin{eqnarray}\label{constraint_subjrel}
d_{{t_i}}^{{r^{t_i}}} + d_{{t_j}}^{{r^{t_j}}} \le 1 \\
\forall t_i,t_j: obj(t_i) = subj(t_j) \wedge (r^{t_i}, r^{t_j}) \in \mathcal{C}^{rer} \nonumber
\end{eqnarray}

The object uniqueness constraints ensure that the relations requiring
unique objects do not bear more than one object given a subject.
\begin{eqnarray}\label{constraint_subjrel}
\sum\limits_{t \in Tuple(r),subj(t) = e} {d_t^r}  \le 1 \\
\forall e, r:  r \in \mathcal{C}^{ou} \nonumber
\end{eqnarray}
where $e$ is an entity, $Tuple(r)$ are the tuples whose candidate relations contain $r$.

Similarly, the subject uniqueness constraints ensure that given an object,
the relations expecting unique subjects do not bear more than one subject.
\begin{eqnarray}\label{constraint_subjrel}
\sum\limits_{t \in Tuple(r),obj(t) = e} {d_t^r}  \le 1 \\
\forall e, r: r \in \mathcal{C}^{su} \nonumber
\end{eqnarray}

\subsubsection{Soft Style Formulation}

To utilize the constraints in a soft style, we give each constraint a non-negative
penalty of violating it. The higher the penalty is, the more confident we are that the corresponding
constraint should not be violated. An extreme case is that the constraint has a penalty of infinite, which means this constraint should not be violated at all, working as a hard constraint.
Again, our aim is to find an
optimal solution which maximizes the objective function (Equation~\ref{objective}) as defined in Section~\ref{sec:objectiveFunc},
and minimizes the penalty of violating the constraints at the same time.

We then rewrite the  objective function (Equation~\ref{objective}) to accommodate constraints in a soft style. 
For each hard constraint $d_{{t_i}}^{{r^{t_i}}} + d_{{t_j}}^{{r^{t_j}}} \le 1$
discussed in Section~\ref{sec:hard}, we define a new decision variable, ${d_{{t_i},{t_j}}^{{r^{{t_i}}},{r^{{t_j}}}}}$,
which will be 1 if the constraint is violated (that is, when both $d_{{t_i}}^{{r^{t_i}}}$ and $d_{{t_j}}^{{r^{t_j}}}$ are 1),
and 0 otherwise. Note that each ${d_{{t_i},{t_j}}^{{r^{{t_i}}},{r^{{t_j}}}}}$ is still a binary decision
variable, but each one will be associated with a continuous penalty of violating the corresponding constraint.
However, for the constraints generated from the second category of clues (\textit{violations of
argument uniqueness}), it is difficult to fit them linearly into the objective function\footnote{For a constraint like $\sum\limits_{t \in Tuple(r),obj(t) = e} {d_t^r}  \le 1$, it will
be violated when more than one $d_t^r$ equal to 1. We need a decision variable similar with ${d_{{t_i},{t_j}}^{{r^{{t_i}}},{r^{{t_j}}}}}$ to represent the
violation of the constraint. However, it is impossible to use a linear combination
of several linear constraints  to make the decision variable act as we expect.}.

Therefore, in this paper, we only transform the constraints from \textit{implicit argument types inconsistencies} 
into soft constraints. Starting from Equation~\ref{objective}, we take into account the penalty values of violating the
soft constraints, and formulate the total penalty as a new component of the objective, which can be defined as:
\begin{eqnarray*}\label{soft_objective}
\max (\sum\limits_{t \in {\cal T},r \in {R^t}} {(c} onf(t,r) + \mathop {\max }\limits_{{\rm{m}} \in M_t^r} score(m,r))d_t^r - \sum {d_{{t_i},{t_j}}^{{r^{{t_i}}},{r^{{t_j}}}}p({r^{{t_i}}},{r^{{t_j}}})} ) \\
\begin{array}{l}
 \forall {t_i},{t_j}:subj({t_i}) = subj({t_j}) \wedge ({r^{{t_i}}},{r^{{t_j}}}) \in {{\cal C}^{sr}}\\
\forall {t_i},{t_j}:obj({t_i}) = obj({t_j}) \wedge ({r^{{t_i}}},{r^{{t_j}}}) \in {{\cal C}^{ro}}\\
\forall {t_i},{t_j}:obj({t_i}) = subj({t_j}) \wedge ({r^{{t_i}}},{r^{{t_j}}}) \in {{\cal C}^{rer}}
\end{array}
\end{eqnarray*}
The penalty values $p({r^{{t_i}}},{r^{{t_j}}})$ of violating the constraints should be calculated based
on the clues from which they are generated. However,
it is impossible to manually assign proper penalty values for each of those clues, so we stick on
the automatically generated clues for the soft constraints.
There could be many ways to automatically obtain the penalty values, and one straightforward
method is to use the $\mathcal{K}$ scores of the clues from which those constraints are generated: 
\begin{eqnarray}\label{penalty}
p({r^{{t_i}}},{r^{{t_j}}}) = -\alpha \mathcal{K}({r^{{t_i}}},{r^{{t_j}}})
\end{eqnarray}
where the parameter $\alpha$ adjusts the weights of the penalty of
violating the constraints, and $\mathcal{K}({r^{{t_i}}},{r^{{t_j}}})$ is calculated
based on the argument sets of the two relations as described in Section \ref{conflict_gen}.
Since the $\mathcal{K}$ scores are negative and the penalty values are usually considered as positive, we thus add a minus sign in the equation. 

In order to restrict the decision variable ${d_{{t_i},{t_j}}^{{r^{{t_i}}},{r^{{t_j}}}}}$ to be 1
only when both $d_{{t_i}}^{{r^{t_i}}}$ and $d_{{t_j}}^{{r^{t_j}}}$ are 1,
we need to add the following new constraints:
\begin{eqnarray}\label{constraint_for_soft}
d_{{t_i},{t_j}}^{{r^{{t_i}}},{r^{{t_j}}}} \le d_{{t_i}}^{{r^{{t_i}}}}\\
d_{{t_i},{t_j}}^{{r^{{t_i}}},{r^{{t_j}}}} \le d_{{t_j}}^{{r^{{t_j}}}}\\
d_{{t_i}}^{{r^{{t_i}}}} + d_{{t_j}}^{{r^{{t_j}}}} \le d_{{t_i},{t_j}}^{{r^{{t_i}}},{r^{{t_j}}}} + 1\\
\begin{array}{l}
\forall {t_i},{t_j}:subj({t_i}) = subj({t_j}) \wedge ({r^{{t_i}}},{r^{{t_j}}}) \in {{\cal C}^{sr}}\\
\forall {t_i},{t_j}:obj({t_i}) = obj({t_j}) \wedge ({r^{{t_i}}},{r^{{t_j}}}) \in {{\cal C}^{ro}}\nonumber \\
\forall {t_i},{t_j}:obj({t_i}) = subj({t_j}) \wedge ({r^{{t_i}}},{r^{{t_j}}}) \in {{\cal C}^{rer}}
\end{array}
\end{eqnarray}

No matter how we formalize those constraints, hard or soft, the ILP based joint optimization framework helps us
 combine the confidence from  local extractors and the implicit relation background encoded in the
global consistencies among entity tuples together. After
the optimization problem is solved, we will
obtain a full configuration per each candidate relation, which will help us eliminate incorrect candidates.





\section{Experiments}\label{sec_experiment}
We will compare our proposed framework with  both traditional and neural networks based state-of-the-art relation extractors on different datasets in different languages. Specifically, these experiments  are designed to address the following questions:  (a) whether our proposed framework can effectively handle the global inconsistency among local predictions? (b) whether our framework can work with automatically obtained clues, except manually designed clues? (c)  how differently the framework performs with hard-style or soft-style constraints?

Next, we will introduce the datasets, describe the baseline models used for comparisons, and present the results in detail.

\subsection{Datasets}
We evaluate our approach on three datasets, including two
English datasets and one Chinese dataset.

The first English dataset, Riedel's dataset, is the one used in
\cite{Sebastian2010,Hoffmann:2011:KWS:2002472.2002541,Mihai2012}, with the same training/test split as previous works. It uses
Freebase as the knowledge base,  covering 52 Freebase relations in total,  and the New York Time corpus~\cite{nyt} as the text corpus,
including about 281,000 entity tuples, 552,000 sentences in the training set, about 96,000
entity tuples, and 172,000 sentences in the testing set.

We construct another English dataset, the DBpedia dataset, by mapping the triples in
DBpedia~\cite{Bizer:2009} to the sentences in the  New York Time corpus.
We map 51 different relations to the corpus and result in about 134,000 sentences with 50,000
entity tuples for training, and 53,000 sentences with 30,000 entity
tuples for testing.

For the Chinese dataset, we derive knowledge facts and construct a Chinese KB
from the Infoboxes of HudongBaike\footnote{www.baike.com}, one of the largest Chinese online encyclopedias.
We collect four national economic newspapers in 2009 as the text corpus.
28 different relations are mapped to the corpus, resulting in 60,000
entity tuples, 120,000 sentences for training, and
40,000 tuples, 83,000 sentences for testing.

We obtain two kinds of global clues, regarding \textit{implicit argument types inconsistencies}
and \textit{violations of arguments' uniqueness}, respectively, for all three
datasets. Each clue in the first category is a pair of relations,
e.g., $<$country, nationality$>$ is a clue from $\mathcal{C}^{sr}$, which means
it is unlikely that the two relations can share subjects. Each clue in the
second category is a single relation who requires unique argument values.
For example, $capital$ is a relation from $\mathcal{C}^{ou}$, indicating that
we can only accept one object given a specific subject for relation $capital$.

The Riedel's dataset is produced from a much earlier version of
Freebase which we cannot access now, thus can not obtain any clues automatically.
We therefore manually collect clues for this dataset only.
The numbers of the clues manually and automatically obtained for those datasets are listed in
Table~\ref{table_clues}. We list the manually obtained clues for the DBpedia and Chinese datasets in the Appendix.
\begin{table*}
\centering \caption{\label{table_clues}The number of the clues on the three datasets.}
\setlength\tabcolsep{4pt}
\begin{tabular}{c|cc|cc}
\hline
\bfseries Dataset & \multicolumn{2}{c|}{\bfseries Manual Clues} & \multicolumn{2}{|c}{\bfseries Automated Clues} \\
 & \small The 1st Category & \small The 2nd Category  & \small The 1st Category & \small The 2nd Category \\
\hline
\itshape Riedel's & 58 & 11 & - & - \\
\itshape DBpedia & 65 & 8 & 676 & 18  \\
\itshape Chinese & 60 & 12 & 1341 & 28  \\
\hline
\end{tabular}
\end{table*}

As for the number of the generated constraints on each dataset, if we consider all
the predefined relations as candidates, there would be a huge number of constraints which only depend on
the number of entity pairs and what clues we have.
However, in practice we could select candidates
from the top predictions of the local extractors as discussed in Section~\ref{sec_candidate},
thus the number of the constraints will also be related to the local extractors. Generally speaking,
the extractors that are more confident with their predictions will result in less candidates,
and also less constraints.


We also notice that, given thousands of different relations in the DBpedia or HudongBaike KB,
we only manage to map dozens of different relations from  the corpora we used. In order to extract other types of relations, one may need to collect different resources in the future.

\subsection{Baselines and Competitors}
For a thorough examination, we compare with both traditional RE models and modern neural networks based RE models.
\begin{enumerate}
\item \textbf{MaxEnt}: Maximum Entropy~\cite{Berger:1996} is a widely used baseline extractor in distantly supervised RE, which usually takes both lexical and syntactic features as input. In our experiments, we also incorporate N-gram based lexical features for an expectation of higher recall.
\item \textbf{MultiR}: is a novel joint model that is designed to deal with the overlap between multiple relations~\cite{Hoffmann:2011:KWS:2002472.2002541}.
\item \textbf{MIML-RE}:  follows a two-step paradigm, which first uses a multi-class classifier to make latent predictions for the mentions, which are then fed to
a bunch of binary classifiers to decide whether the entity pair holds the corresponding relation. MIML-RE is one of the state-of-the-art  traditional
    feature based RE systems~\cite{Mihai2012}.
\item \textbf{NN-avg}:  is a variant of the PCNN model in \cite{zengEMNLP2015}, where the embedding
of an entity pair is obtained by averaging the embeddings of the sentences containing this pair. It is also
used as a baseline extractor in~\cite{linACL2016}. 
\item \textbf{NN-att}: utilizes the attention mechanism to weight each
sentence inside a bag\footnote{We call all the sentences containing an entity pair as a bag for that entity pair.}, while calculating the  embedding for an entity pair~\cite{linACL2016}. \texttt{NN-att} is currently one of the state-of-the-art neural network RE algorithms.
\end{enumerate}
Note that  
all the above models are originally designed to output
entity pair level predictions.

Our general framework will take sentence level predictions as input, resolve the inconsistencies among them and finally output the entity pair level results. We  thus collect the sentence level predictions from \textbf{MaxEnt}, \textbf{MultiR}, \textbf{NN-avg} and \textbf{NN-att}\footnote{We can not find local predictions from MIML-RE},  feed them to our framework, and compare our proposed method with their originally designed integration schemes as well as other existing integration methods.
%
%

\begin{enumerate}
\item \textbf{Mintz++}: is the baseline integration mechanism described in \cite{Mihai2012}, which obtains multi-label outputs for an entity tuple by \textbf{OR-ing}  all its local predictions. A similar approach is also adopted in MultiR.
\item \textbf{NN-avg-bag}: In neural network based models, one simple approach of integration is to average the embeddings
of a bag's all sentences into one bag-level embedding, which is then used to obtain the bag-level predictions~\cite{linACL2016}. 
\item \textbf{NN-att-bag}:  designs an attention mechanism to adaptively assign different weights for different sentences inside a bag, according to their importances for a candidate relation~\cite{linACL2016}.
\end{enumerate}

Following  previous works~\cite{Sebastian2010}, we use the Precision-Recall Curve (P-R Curve) as the evaluation criterion in our experiments. For each model, we rank all final predictions (in the entity pair level) in a descending order according to their  confidence scores.  In the ranked list, we  compute precision/recall at each position, and then plot the P-R Curve for this model.  Usually, the closer the curve is to the upper right corner, the better performance the model has.

\subsection{Applying to Different Extractors}
Our main question is whether our proposed framework can handle the global inconsistencies among
the local predictions from sentence-level relation extractors. 
We examine both traditional feature-based extractors (\textit{MaxEnt} and \textit{MultiR})
and neural networks based extractors (\textit{NN-avg} and \textit{NN-att}) within our framework with
hand crafted global clues. For each dataset we manually
select around 60 relation pairs to capture the clues about \textit{argument types inconsistencies},
and then include all the relations with unique argument values in $\mathcal{R}$.
We first compare our framework with the \textbf{Mintz++} style integration,
which has been proved to be very competitive in \cite{Mihai2012}.

\paragraph{\textbf{The MaxEnt Extractor}}

Firstly, we feed both our framework and a Mintz++ style integrator with the MaxEnt's sentence-level predictions.
\begin{figure*}
\centering
\subfigure[The DBpedia Dataset]{\label{fig:maxentout:a} 
\includegraphics[width=3.9cm]{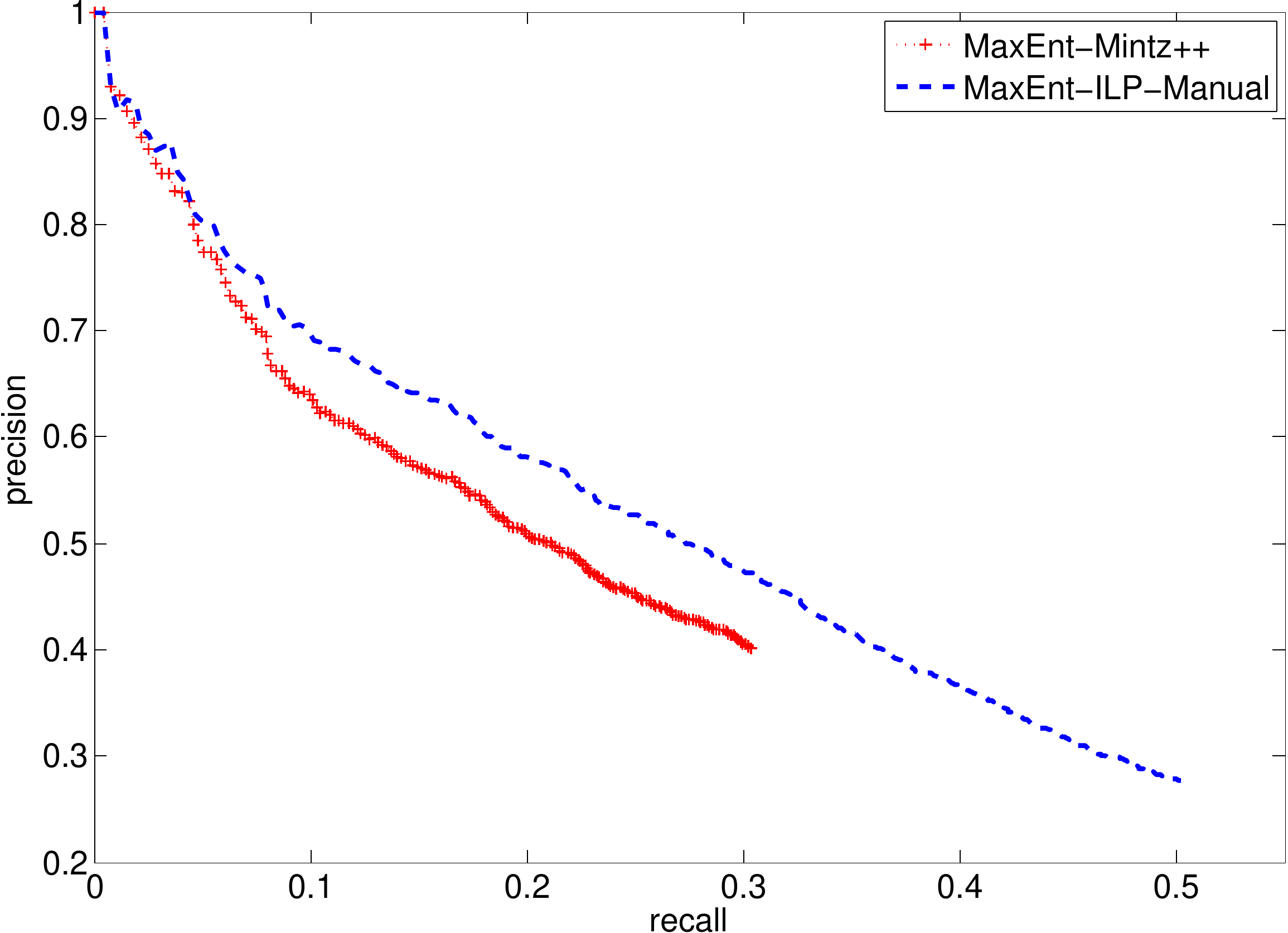}}
\subfigure[The Chinese Dataset]{\label{fig:maxentout:b} 
\includegraphics[width=3.9cm]{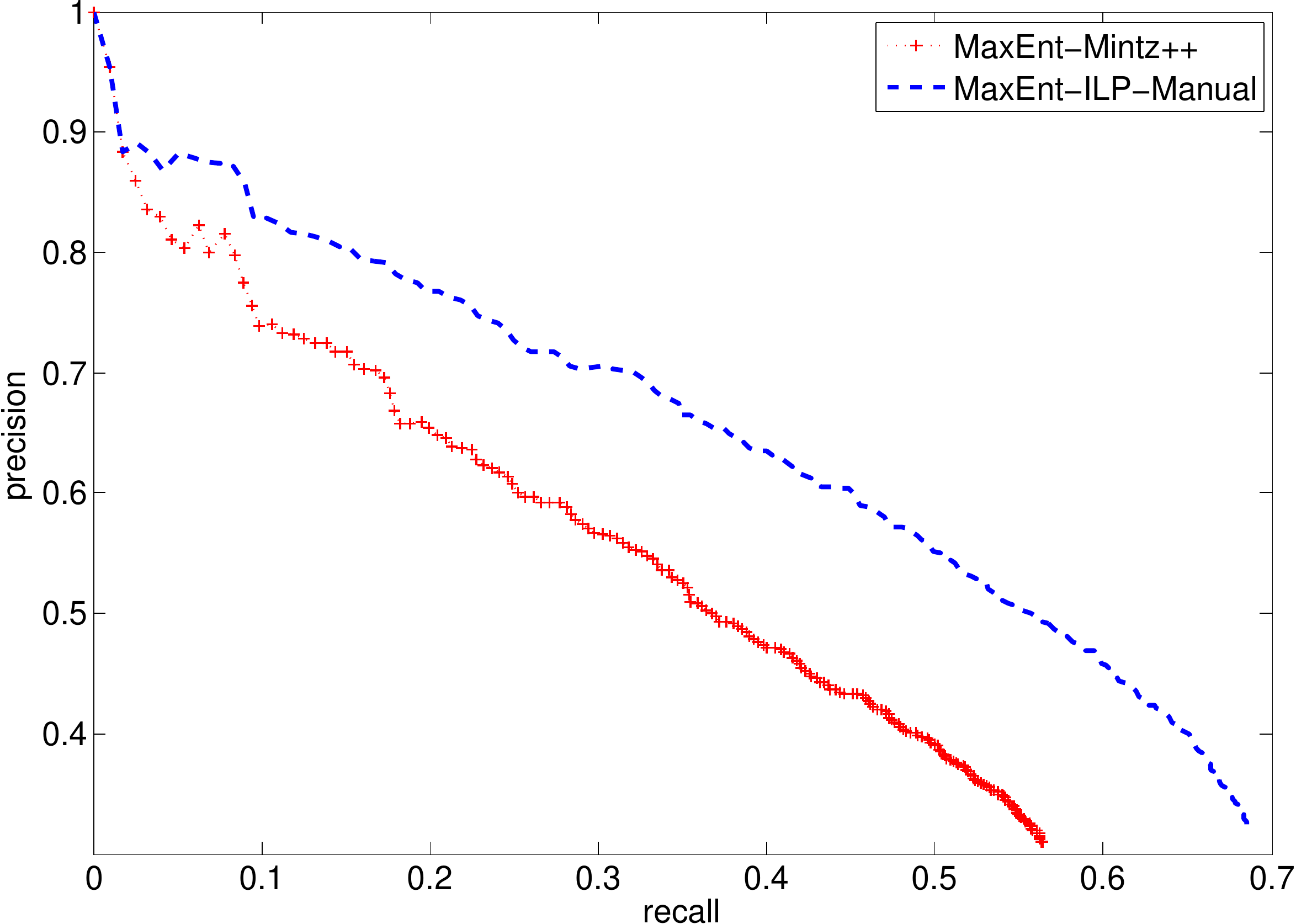}}
\subfigure[The Riedel's Dataset]{\label{fig:maxentout:c} 
\includegraphics[width=3.9cm]{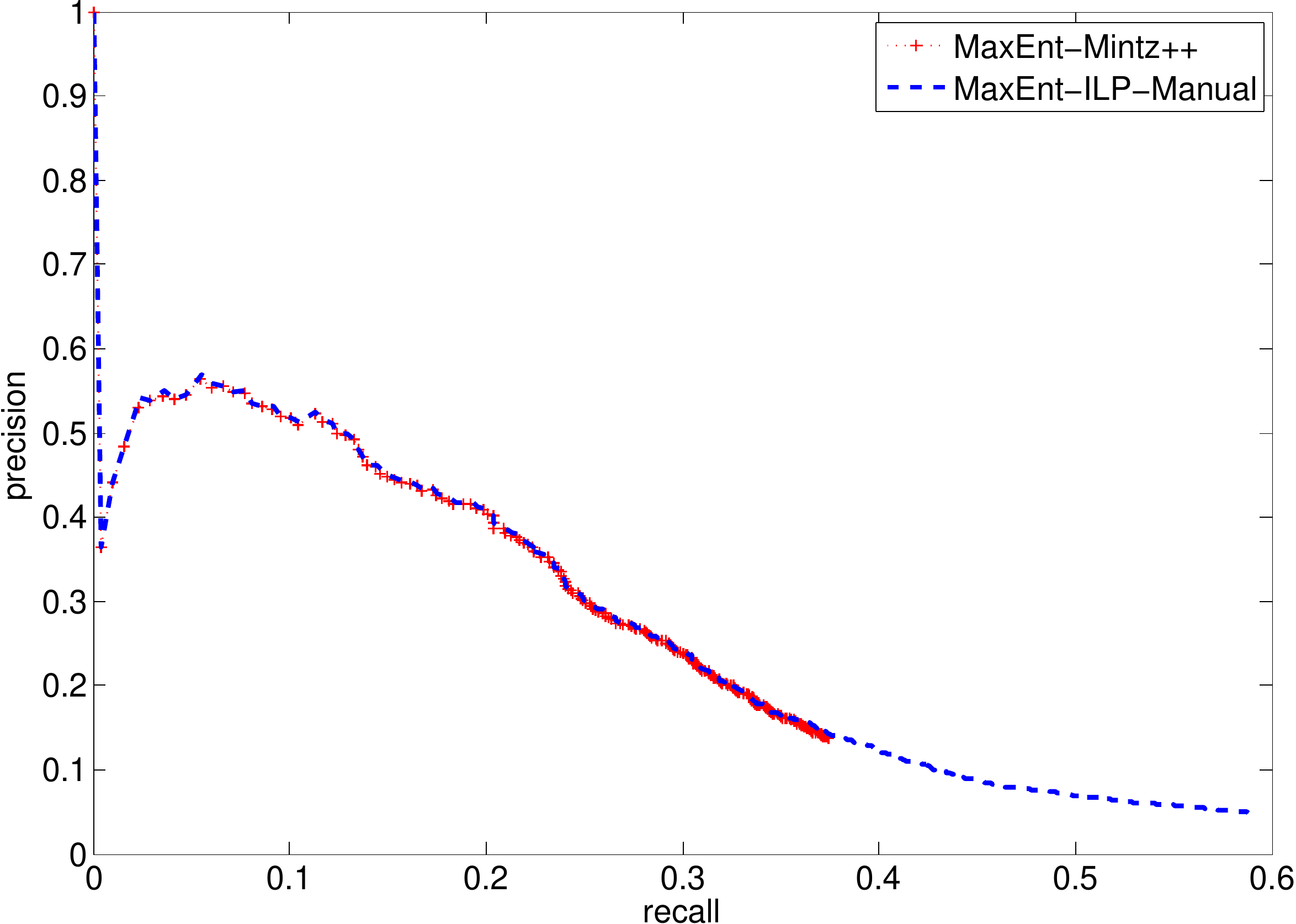}}
\caption{The results of the MaxEnt extractor on three datasets. }\label{fig:maxentout} 
\end{figure*}
As we can see in Figure~\ref{fig:maxentout},  compared with the MaxEnt-Mintz++,
our framework MaxEnt-ILP-Manual wins MaxEnt-Mintz++ at almost every recall point,
performing consistently better in both
the DBpedia dataset and the Chinese dataset. 
Our ILP based framework
takes the same sentence-level local predictions as MaxEnt-Mintz++, but helps filter out incorrect predictions from the local
 MaxEnt extractor, leading to higher precisions.

However, in the Riedel's dataset, our framework cannot improve the performance. 
We manually investigate the dataset, and find that this dataset is dominated by three \textbf{BIG} relations:

\noindent {\small \textit{$/$location$/$location$/$contains}}, {\small \textit{$/$people$/$person$/$nationality}} and {\small \textit{$/$people$/$person$/$place\_lived}},

\noindent which
cover about two-thirds of the entity tuples in total,
and make the output of the local extractor bias even more to these \textbf{BIG} relations.
Specifically, we find that about two-thirds of the local predictions are covered by one relation,
{\small \textit{$/$location$/$location$/$contains}}.
These \textbf{biased} local predictions actually prevent us from collecting
useful constraints from different kinds of clues. The reasons are twofold. Firstly,
the type requirements of this relation are too general, e.g., it is not helpful at all to identify the difference between a city and a country since both of them can act as both subject and object for the relation {\small \textit{$/$location$/$location$/$contains}}. Secondly, most entities are predicted to associate with \textbf{only one} relation, thus there will not be any disagreement among them, and we are not able to correct any wrong predictions, either.
Other relations in this dataset may have some disagreements among local output,
but they are too few to be captured by manually coded clues.
Therefore, in this dataset, we cannot resolve the wrong local predictions by implicitly  capturing the type requirements.

In order to better illustrate how the proposed framework actually works to improve performances,
we compare the outputs of our framework and Mintz++, and
 investigate 
how many incorrect predictions are eliminated 
and corrected
by our framework on the DBpedia and Chinese datasets. 
 We also examine 
how many correct predictions are newly introduced by ILP, which are  predicted as NA by the 
Mintz++ style integration.

\begin{table}
\centering \caption{\label{table_detail}Details of the improvements by our ILP framework in the DBpedia and Chinese datasets. }
\setlength\tabcolsep{4pt}
\begin{tabular}{cccc}
\hline
\bfseries \small Datasets & \small  $\mbox{  }$ \# of eliminated  & $\mbox{  }$\# of corrected & \small $\mbox{  }$ \# of introduced w.r.t. Mintz++  \\
\hline
\itshape \small DBpedia & \small 268 & \small 61 & \small 1426 \\
\itshape \small Chinese & \small 1506 & \small 14 & \small 283 \\
\hline
\end{tabular}
\end{table}

The results summarized in Table~\ref{table_detail} show that our framework can not only eliminate the incorrect
predictions from local extractors, but also introduce more correct predictions,
which are originally ranked the 2nd and 3rd in the local predictions, at the same time.
We also find an interesting result: in the DBpedia dataset, our ILP framework is more likely to introduce
correct predictions, while in the Chinese dataset,
it tends to reduce more incorrect predictions.
This agrees with what is shown in the P-R curves: in the DBpedia dataset, our framework extends longer
along the recall axis, while in the Chinese dataset, it improves the precision significantly.
It is also worth mentioning that the predictions being corrected are much less than the ones being eliminated or newly
introduced  
on both datasets. This is mainly because that 
when correcting a prediction, our framework should first identify an
incorrect prediction for an entity pair, eliminate it, and pick out the next
best prediction with a high enough confidence and mostly compatible with the constraints, at the same time. 

\paragraph{\textbf{The MultiR Extractor}}

MIML-RE and MultiR are two state-of-the-art models among traditional relation extractors.
However, the MIML-RE model can only output entity-pair level results, we thus fit MultiR's
\textbf{sentence-level} extractor into our framework.

\begin{figure*}
\centering
\subfigure[The DBpedia Dataset]{\label{fig:multirout:a} 
\includegraphics[width=3.9cm]{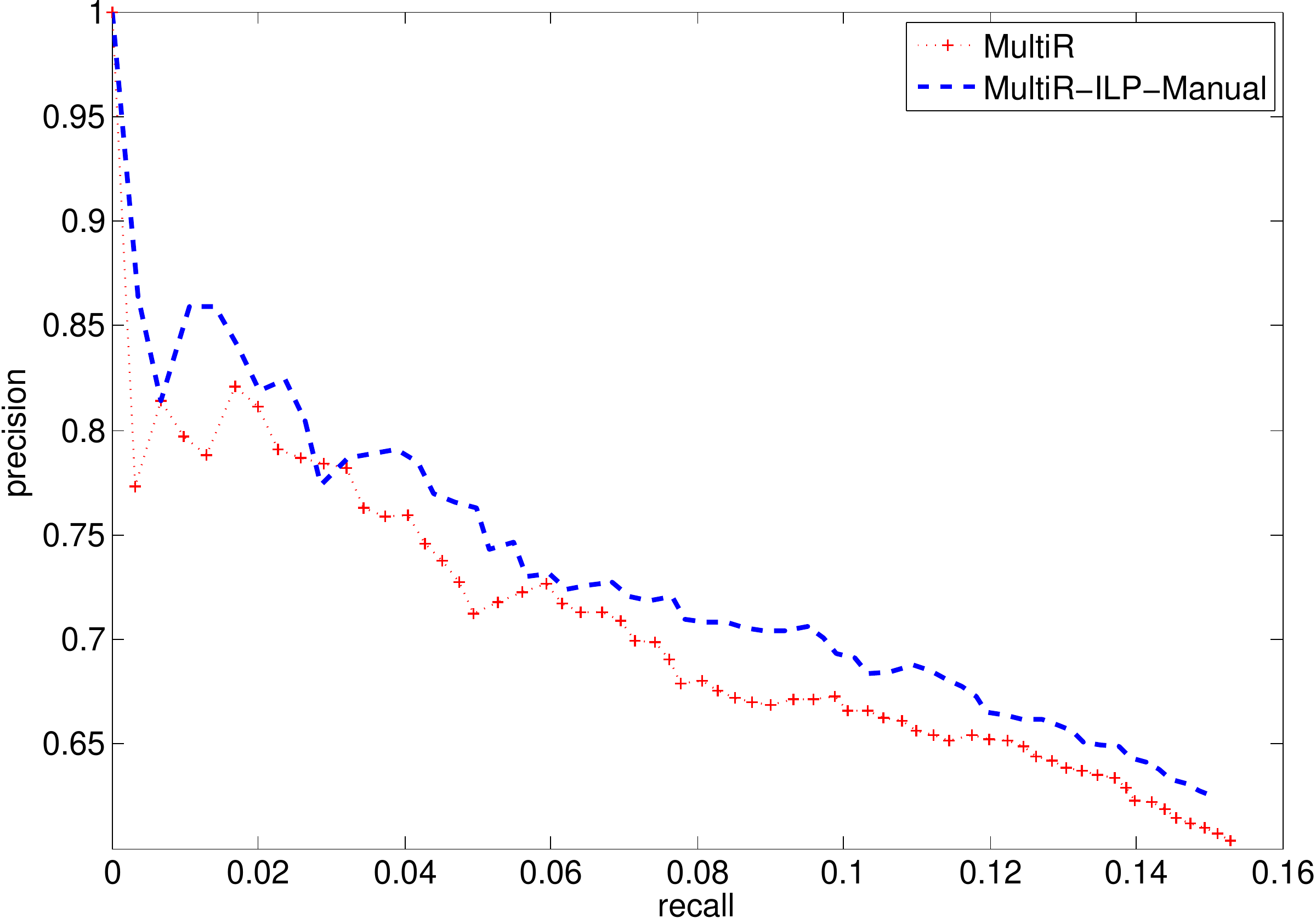}}
\subfigure[The Chinese Dataset]{\label{fig:multirout:b} 
\includegraphics[width=3.9cm]{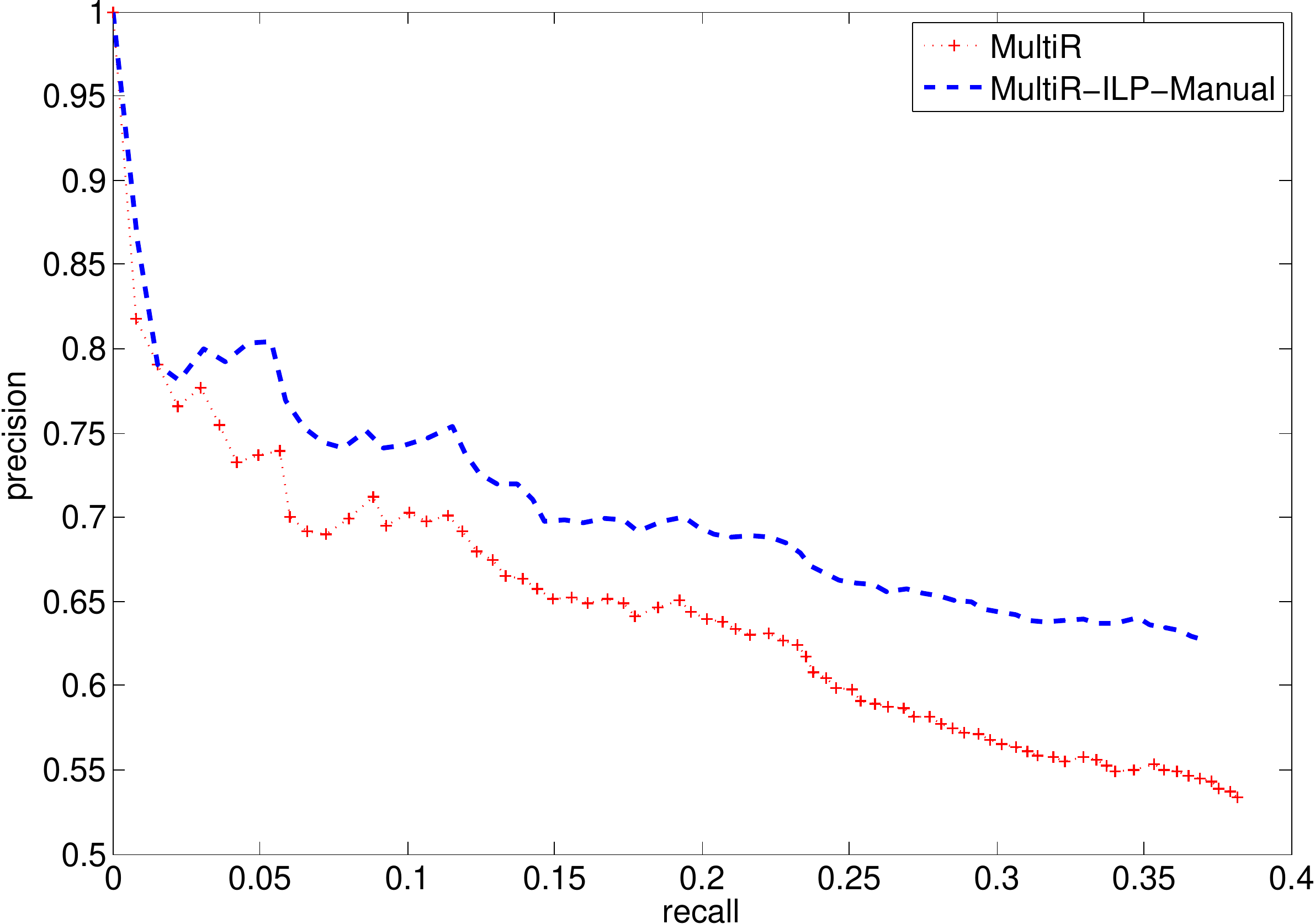}}
\subfigure[The Riedel's Dataset]{\label{fig:multirout:c} 
\includegraphics[width=3.9cm]{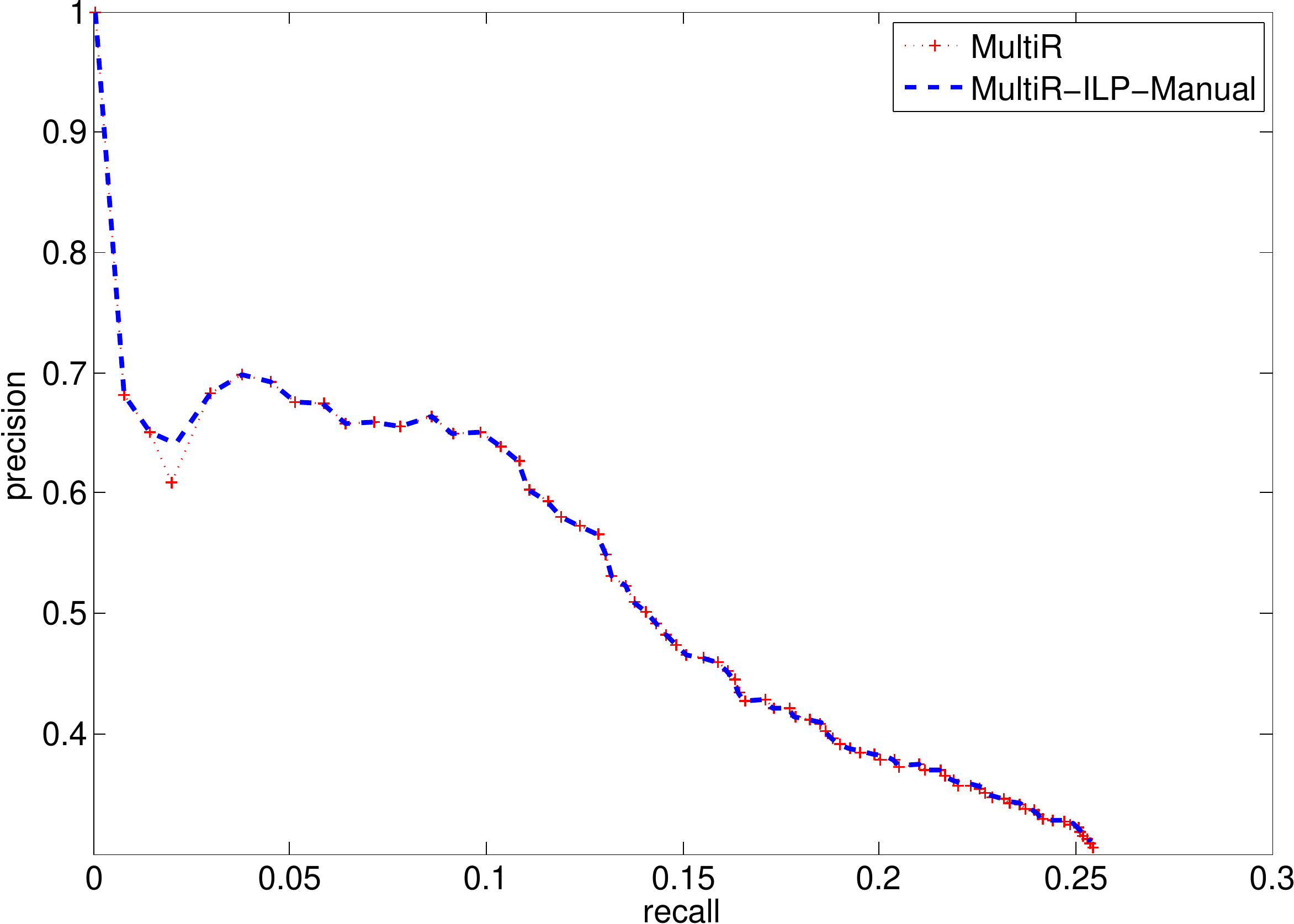}}
\caption{The results of the MultiR extractor on the three datasets. }\label{fig:multirout} 
\end{figure*}

As we can see in Figure~\ref{fig:multirout}, in both the DBpedia and Chinese datasets, our ILP optimized MultiR
outperforms original MultiR in most parts of the curves. Note that,
original MultiR adopts an integration scheme very similar to Mintz++
to merge their sentence-level predictions into entity-pair level.
We think the reason is that our framework makes use of global clues to discard incorrect predictions, while the original MultiR does not.

In addition, the improvements by our framework are not as high as in the MaxEnt case, 
possibly due to the fact that the sentence-level MultiR does not perform well in these two datasets. Furthermore,
the confidence scores output by MultiR
are not normalized to the same scale, which makes it difficult
in setting up a reasonable confidence threshold to select the candidate predictions and contributing to the objective function.
As a result, we only use the top one result as the
candidate, since including top two predictions without thresholding the confidences empirically leads to inferior performances,
indicating that a probabilistic local  extractor
is more suitable for our framework.
We also notice that our framework, again, does not bring significant improvement in the Riedel's dataset, possibly due to the same reasons as discussed in previous section.

\paragraph{\textbf{The Neural Network Extractors}}

Now we feed our framework with the sentence-level output from two neural network relation extractors,
the average based model (\textit{NN-avg}) and the attention based (\textit{NN-att})~\cite{linACL2016}.
\begin{figure*}
\centering
\subfigure[The DBpedia Dataset]{\label{fig:nnout:a} 
\includegraphics[width=3.9cm]{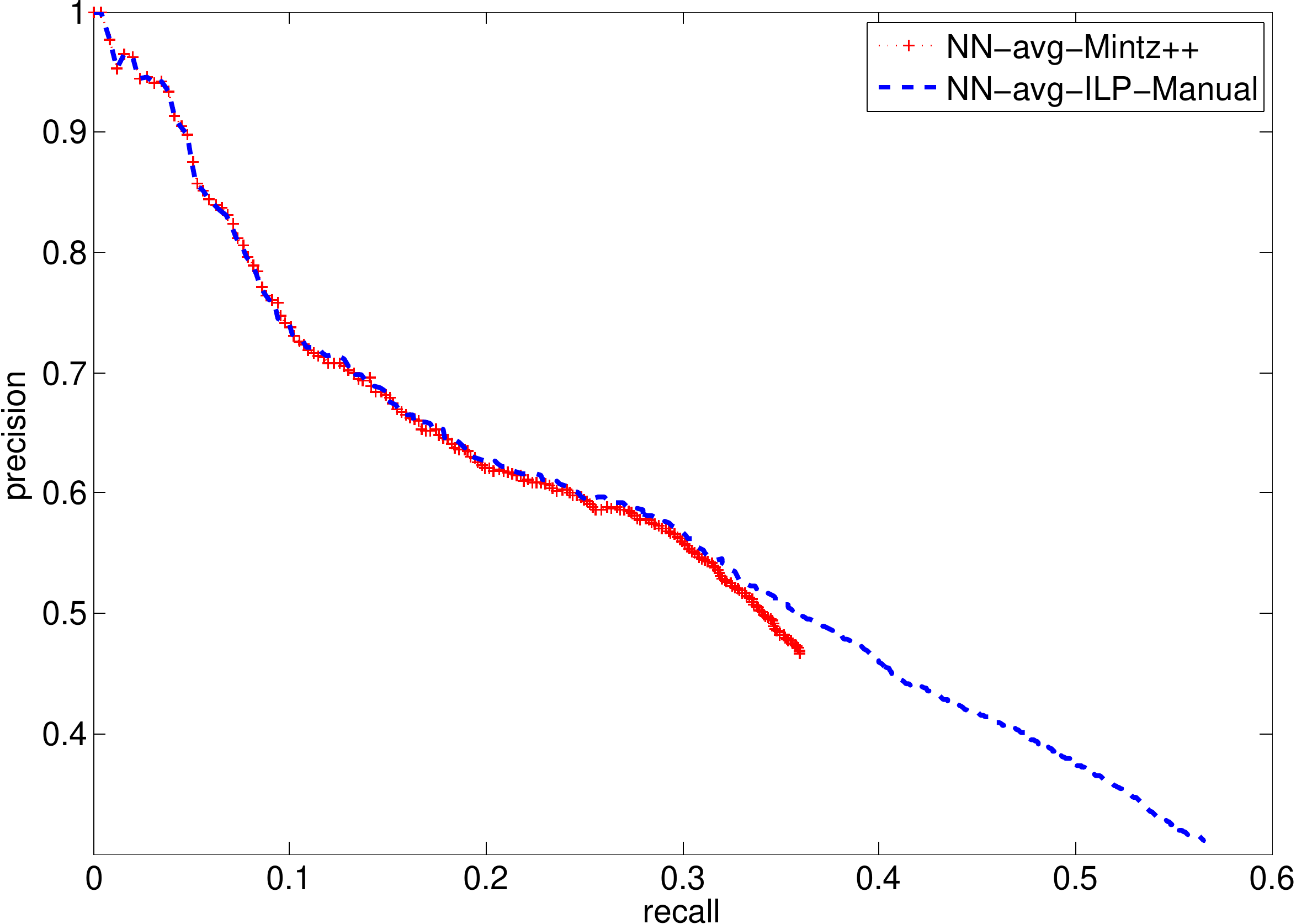}}
\subfigure[The Chinese Dataset]{\label{fig:nnout:d} 
\includegraphics[width=3.9cm]{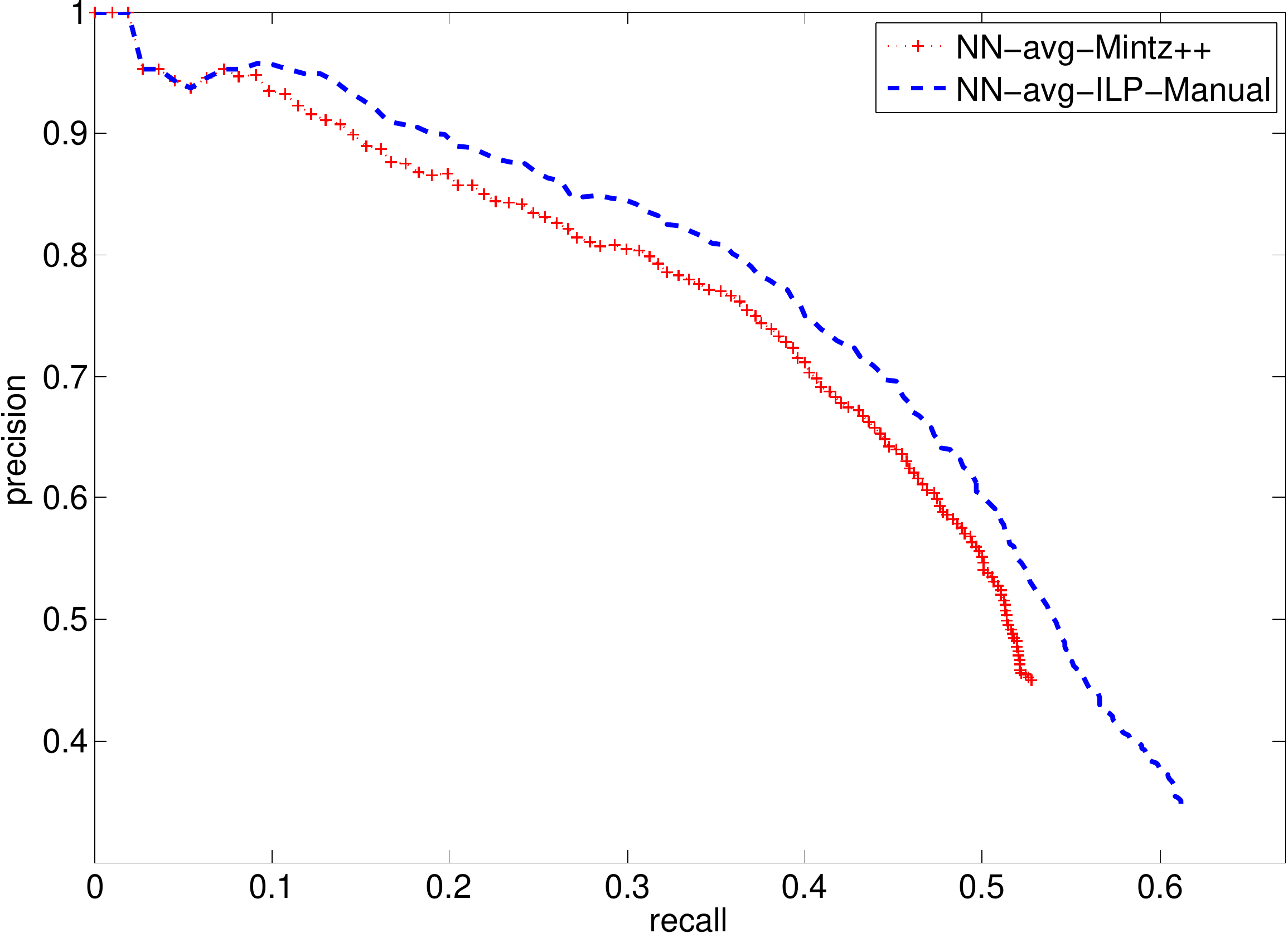}}
\subfigure[The Riedel's Dataset]{\label{fig:nnout:b} 
\includegraphics[width=3.9cm]{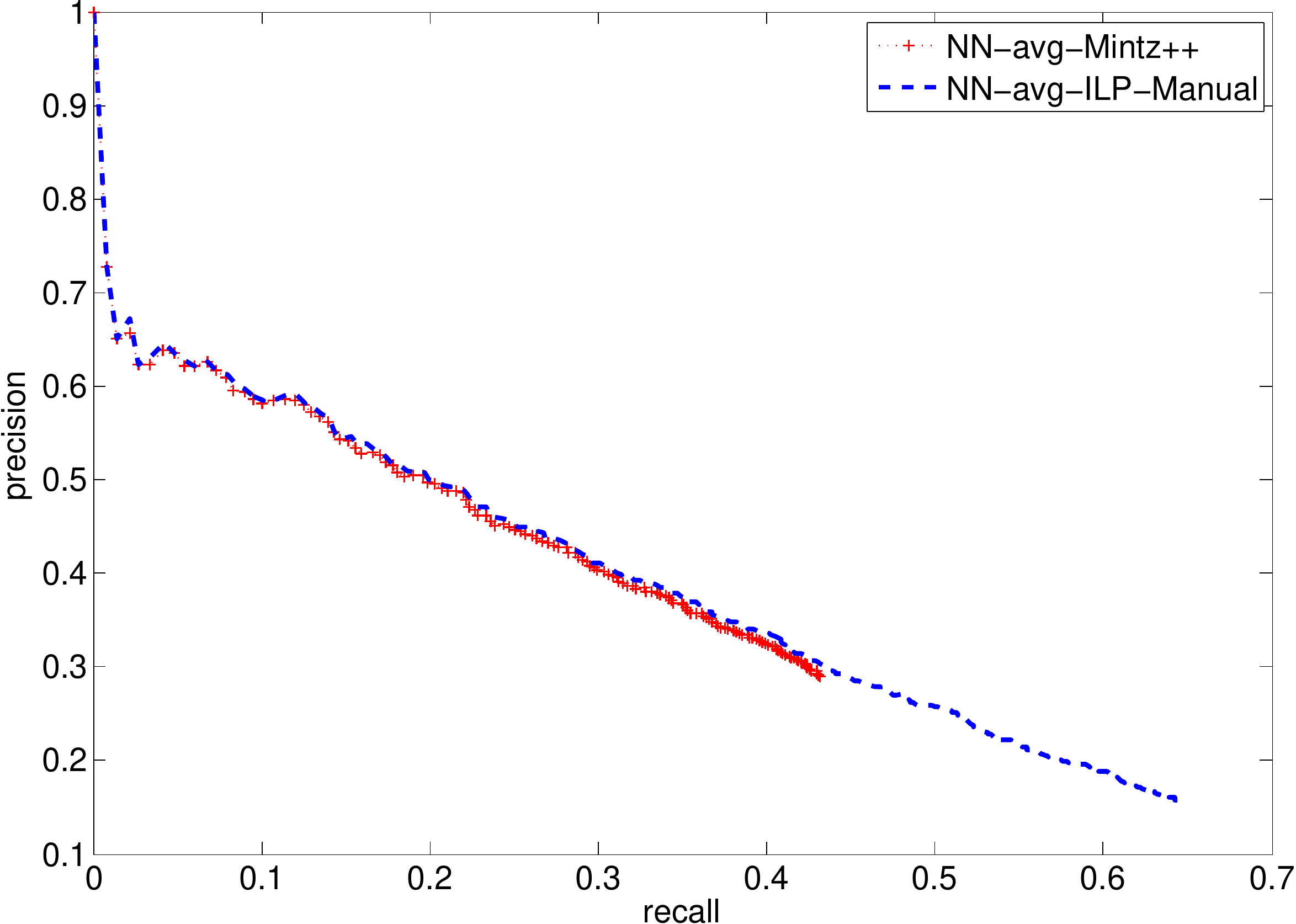}}
\\
\subfigure[The DBpedia Dataset]{\label{fig:nnout:e} 
\includegraphics[width=3.9cm]{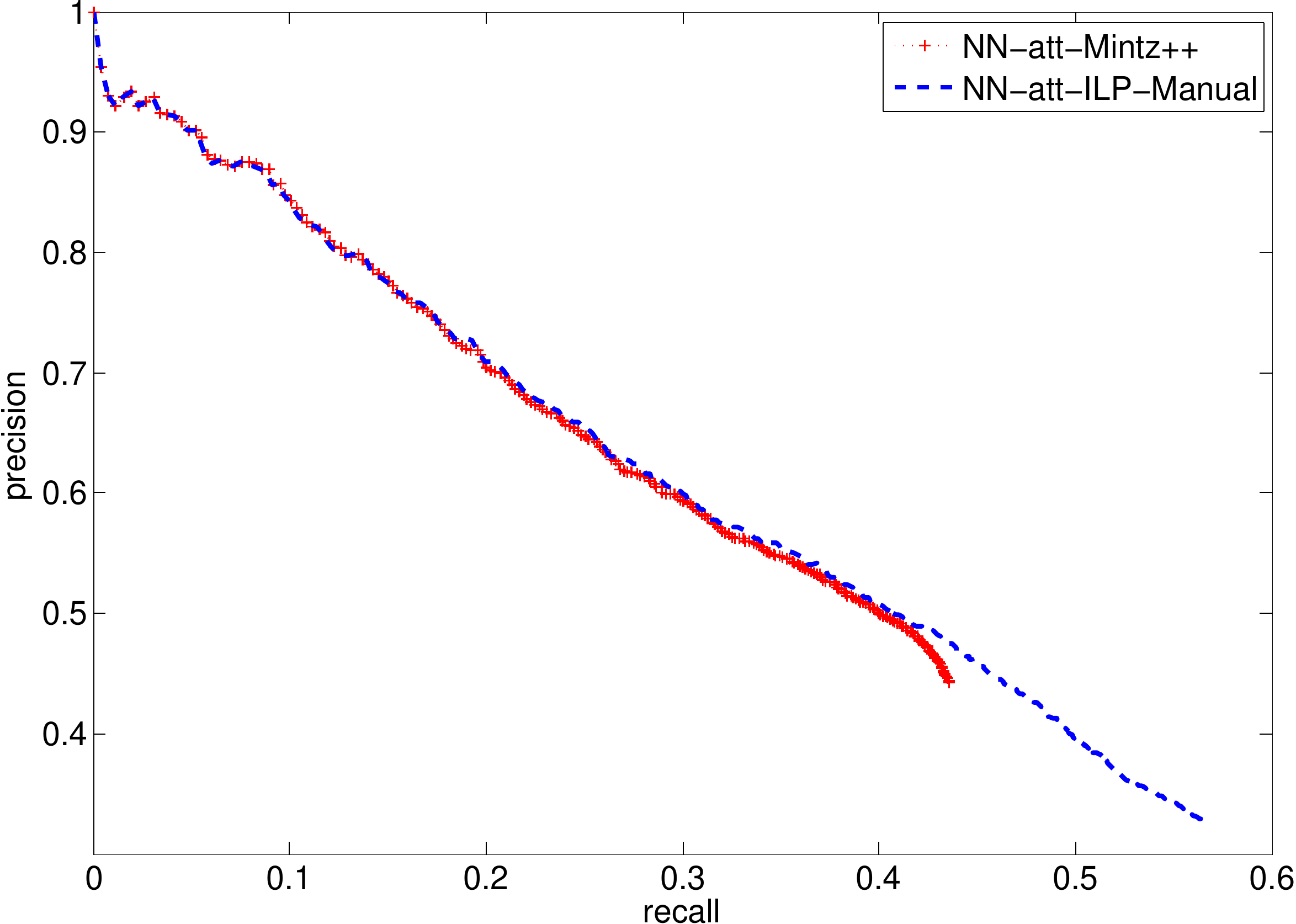}}
\subfigure[The Chinese Dataset]{\label{fig:nnout:c} 
\includegraphics[width=3.9cm]{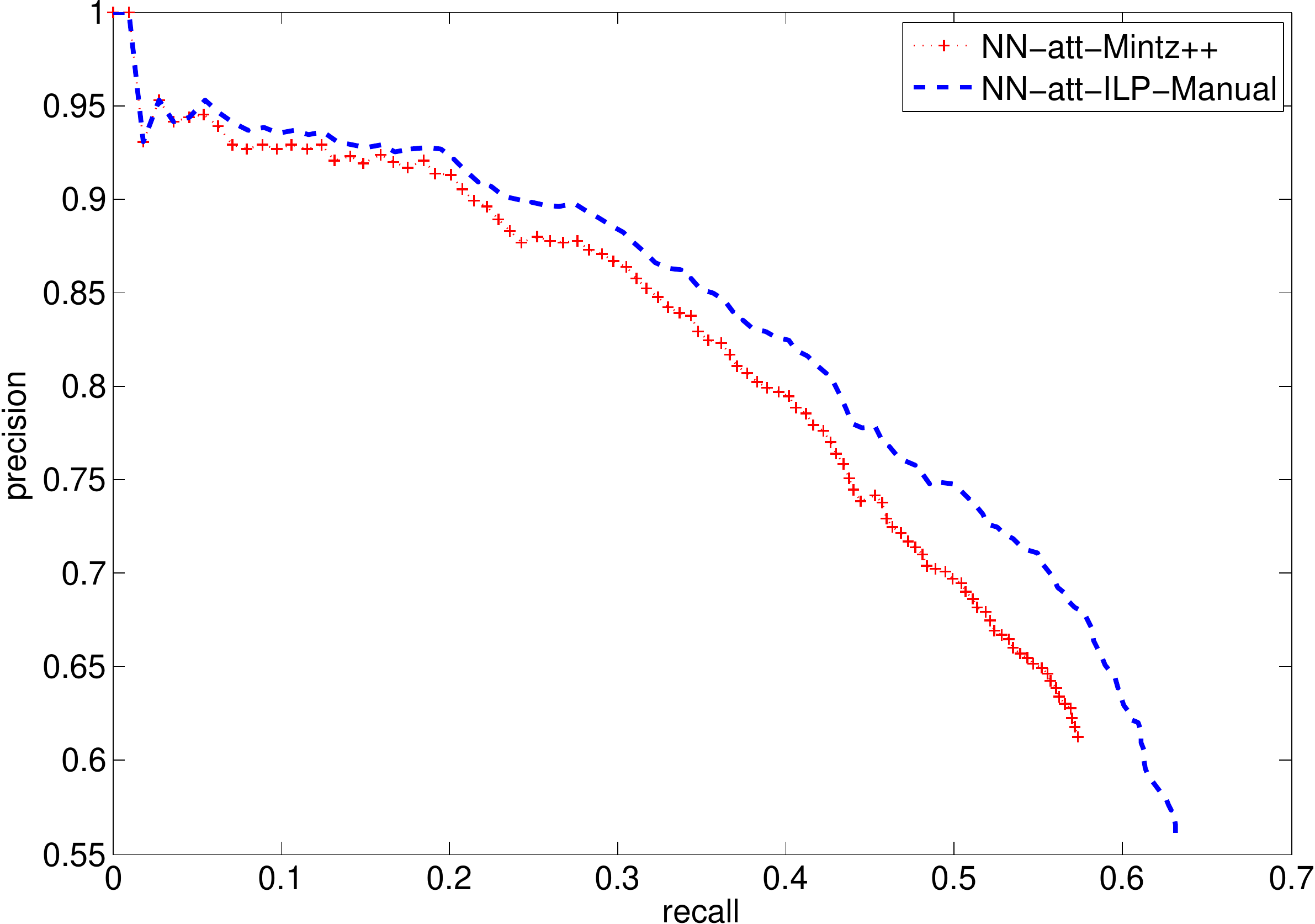}}
\subfigure[The Riedel's Dataset]{\label{fig:nnout:f} 
\includegraphics[width=3.9cm]{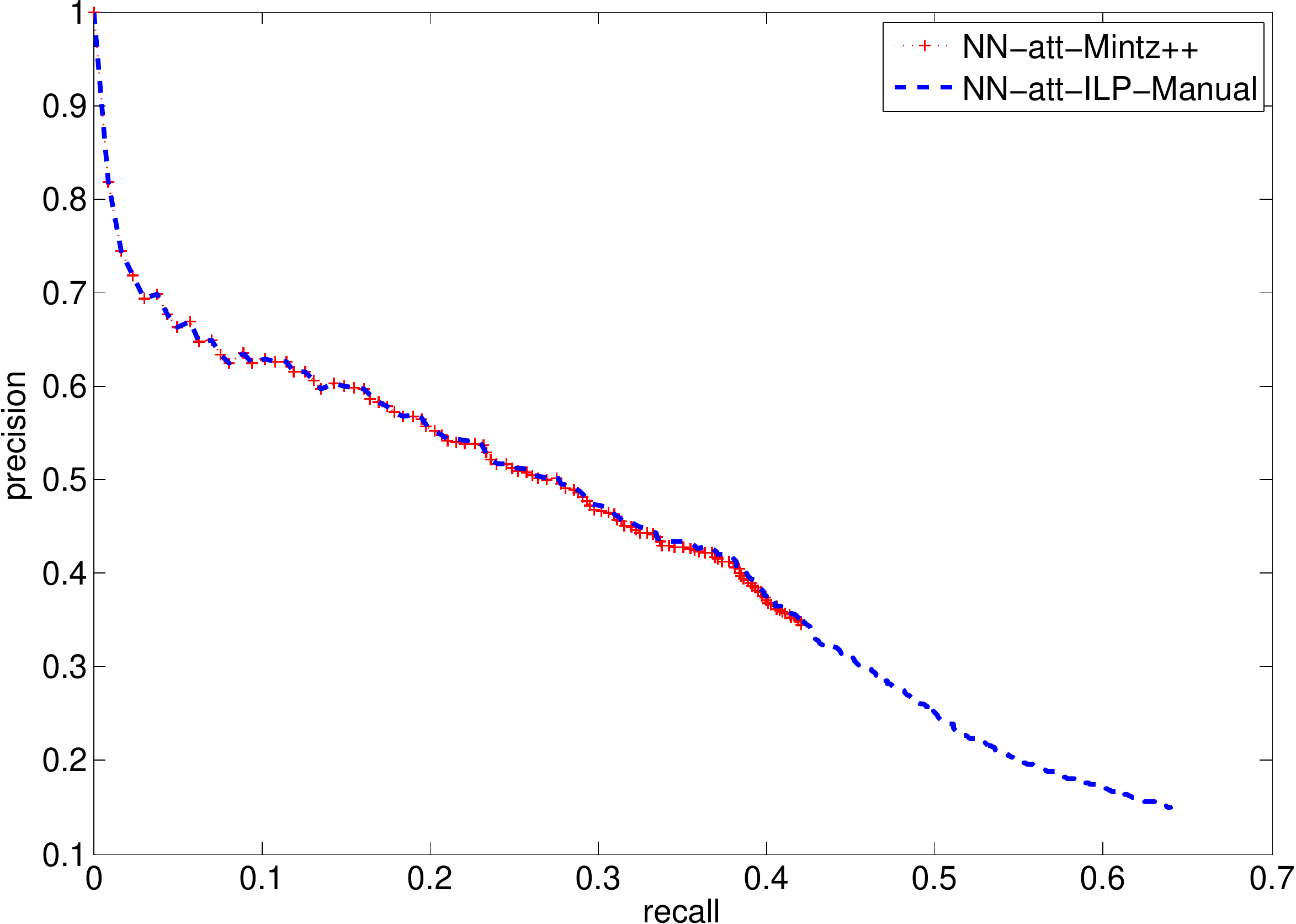}}
\caption{The results of ILP optimized NN models on the three datasets. (a)-(c) are results from \textit{NN-avg}, while (d)-(f) are from \textit{NN-att}. }\label{fig:nnout} 
\end{figure*}

From Figure \ref{fig:nnout} we can observe that, in the Chinese dataset, our ILP framework performs
better than the Mintz++ style integrated NN extractors. However, in the DBpedia dataset, the NN improvement
is not as significant as in the traditional models.
One possible reason is that the result of NN extractors is considered as  more accurate and confident than the traditional models,
resulting in less constraints and less improvements accordingly. And again,
there are still no improvements for both NN models in the Riedel's dataset.

From the above discussion, we find that our proposed framework can effectively work with different local relation extractors, both traditional feature based and neural network based, and improve the overall extraction performance when such global clues  are applicable to the datasets. And for the datasets where we  fail to generate useful constraints, e.g., the Riedel's dataset, our framework does not hurt the performance.


\subsection{Automatically and Manually Obtained Clues}
Now we will investigate the difference between 
the manually obtained clues and automatically generated ones.
The manually obtained clues can only be formulated into constraints in the hard style,
while the automatically generated clues can be used in both hard style and soft style.
Considering that the MultiR extractor does not perform as well as the MaxEnt and NN extractors,
we only list the results of MaxEnt and NN extractors in the following.

\paragraph{\textbf{Results on the MaxEnt Extractor}}

We first compare the performance of the manual clues in the hard style (\textit{Manual}),
automatic clues in the hard style (\textit{Auto-Hard}) and automatic ones in the soft
style (\textit{Auto-Soft}) when they are applied to the output of the MaxEnt extractor.
\begin{figure}
\centering
\subfigure[The DBpedia Dataset]{\label{fig:manvsauto:a} 
\includegraphics[width=5.0cm]{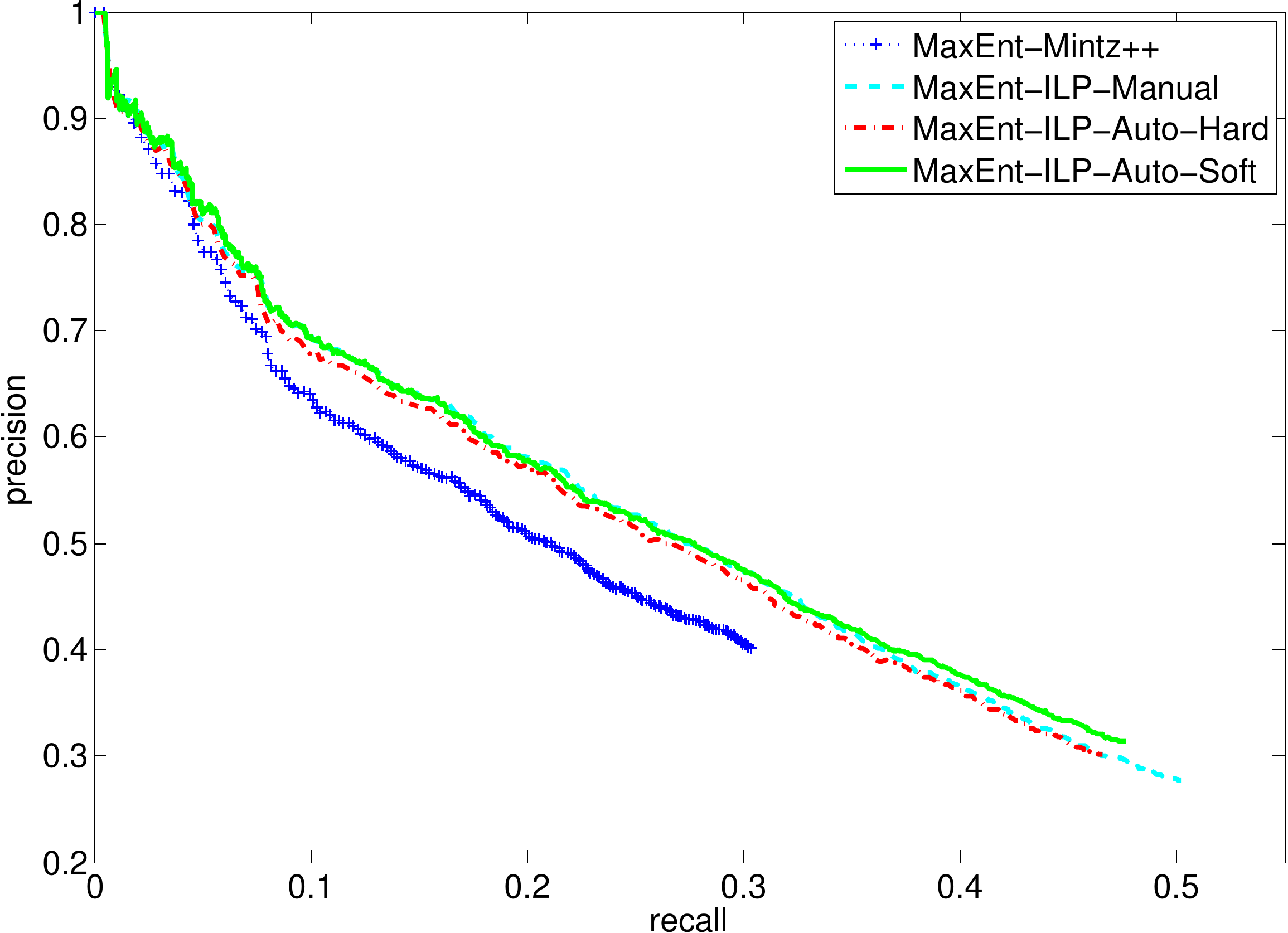}}
\subfigure[The Chinese Dataset]{\label{fig:manvsauto:b} 
\includegraphics[width=5.0cm]{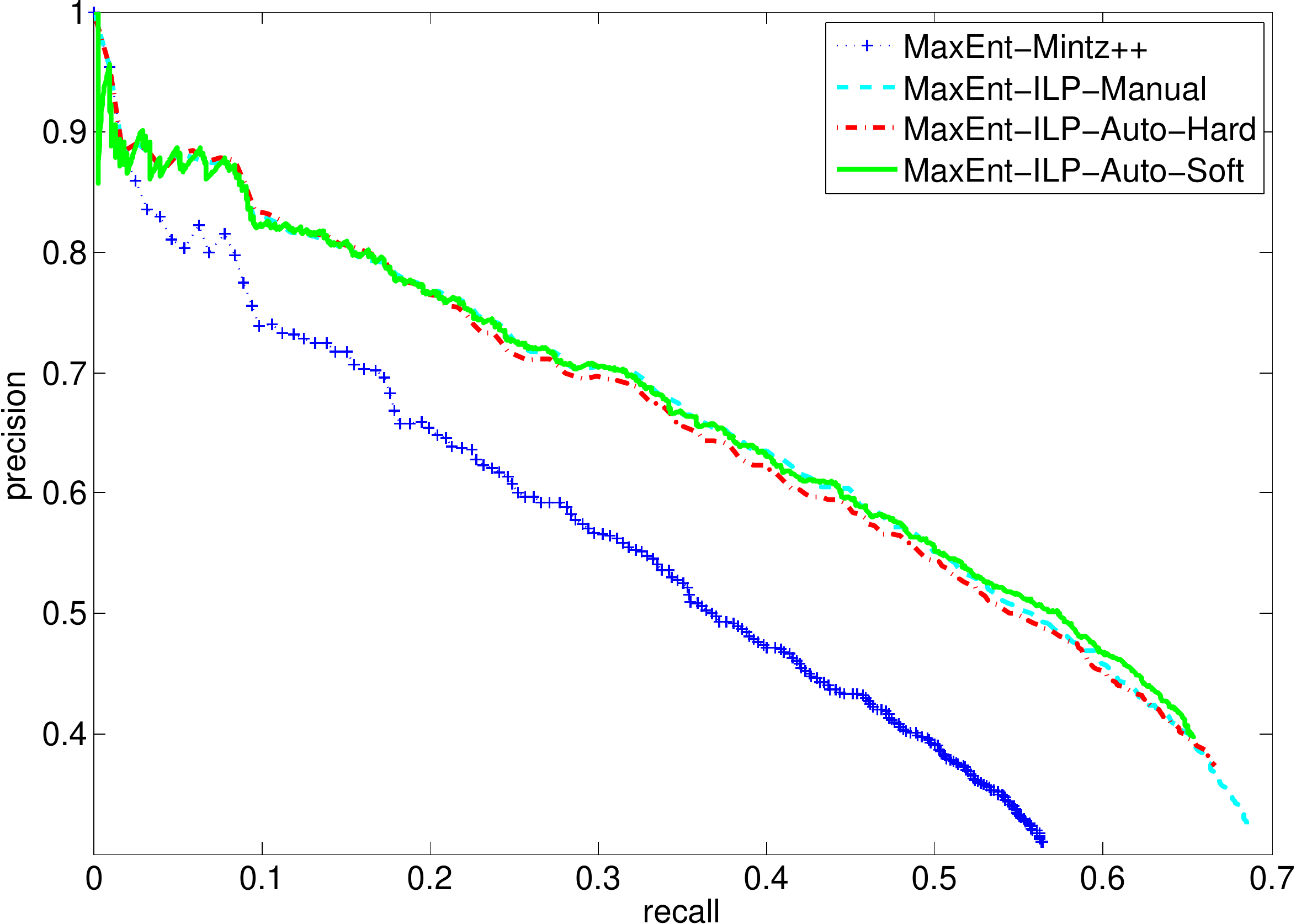}}
\caption{Performance of automatic clues and manual clues with MaxEnt extractor. }\label{fig:manvsauto}
\end{figure}

In Figure~\ref{fig:manvsauto}, we can observe that the results with automatically obtained clues, no matter in hard or soft style,
are comparable to those with manually refined clues. 
This indicates that the automatically collected clues can help to handle the inconsistencies among
local predictions as manually refined clues do, and are shown to be effective in our framework.

%
We also find that the soft style formulation  performs slightly
better than the hard style on both datasets.
We think the reason is that the automatically collected clues in a soft style formulation can take more unusual situations
into consideration, while the hard style is well-directed according to the annotators.

\paragraph{\textbf{Results on the Neural Network Extractors}}

\begin{figure}[h]
\centering
\subfigure[The DBpedia Dataset]{\label{fig:manvsautonn:a} 
\includegraphics[height=4cm]{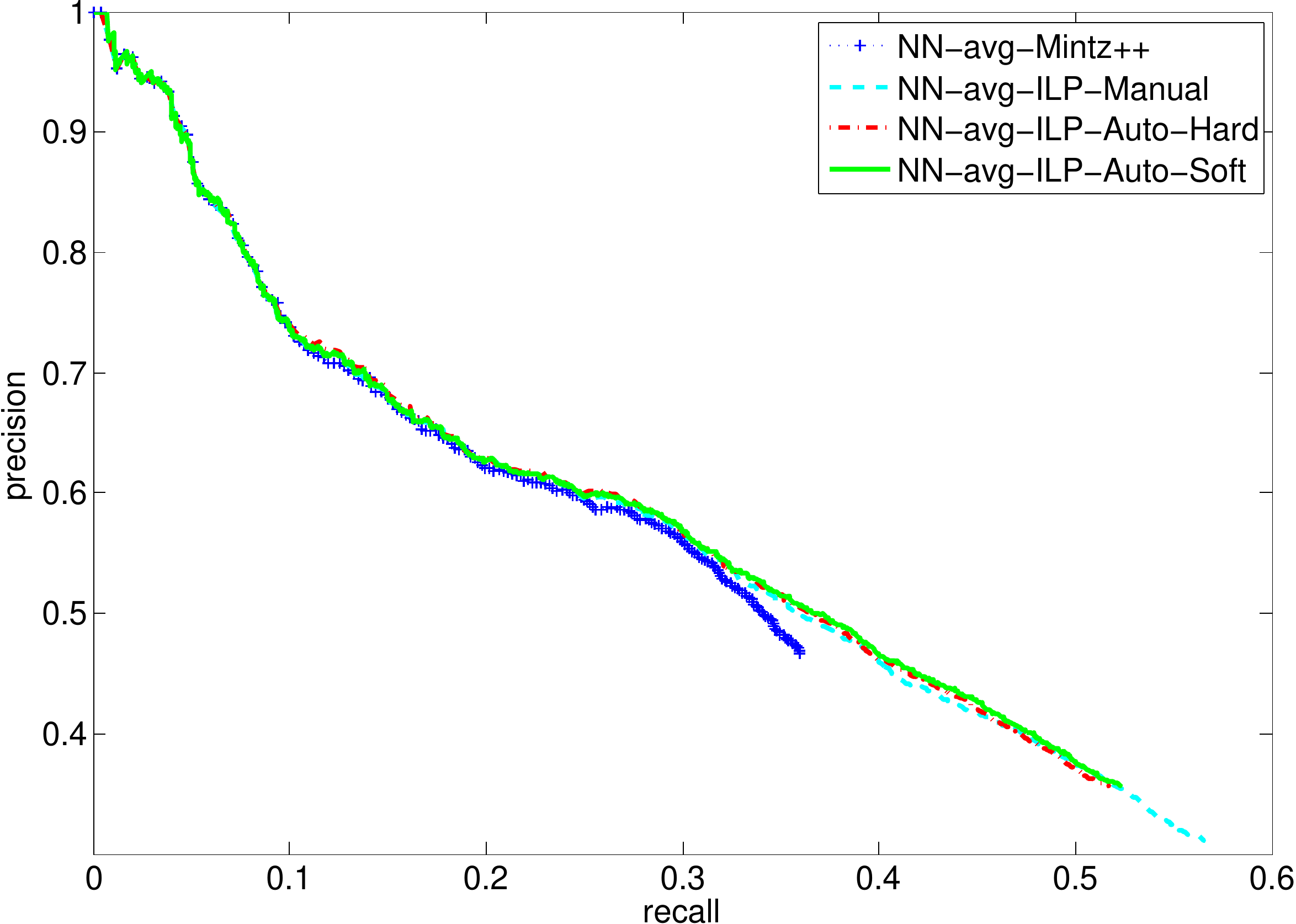}}
\subfigure[The Chinese Dataset]{\label{fig:manvsautonn:b} 
\includegraphics[height=4cm]{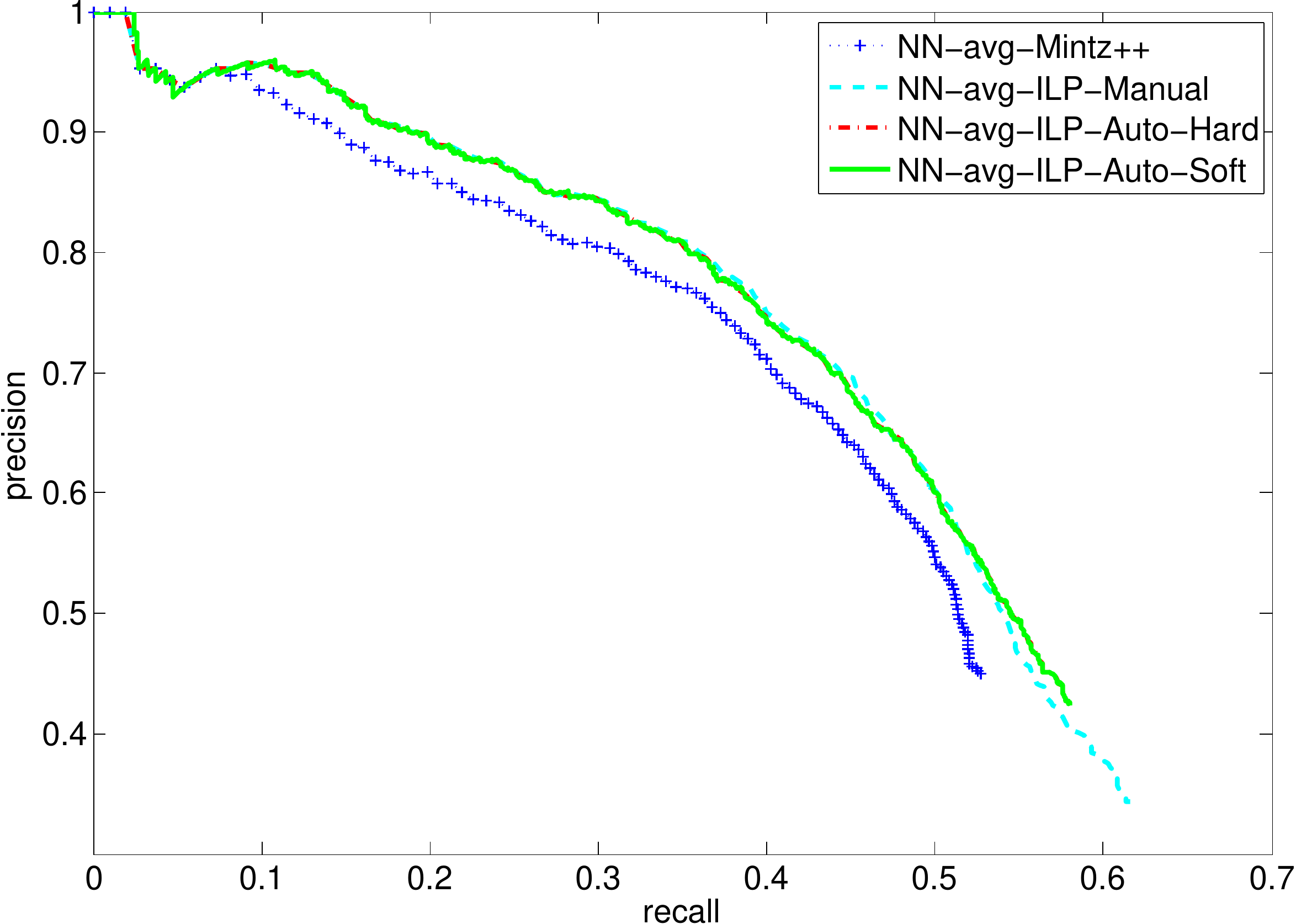}}
\subfigure[The DBpedia Dataset]{\label{fig:manvsautonn:c} 
\includegraphics[height=4cm]{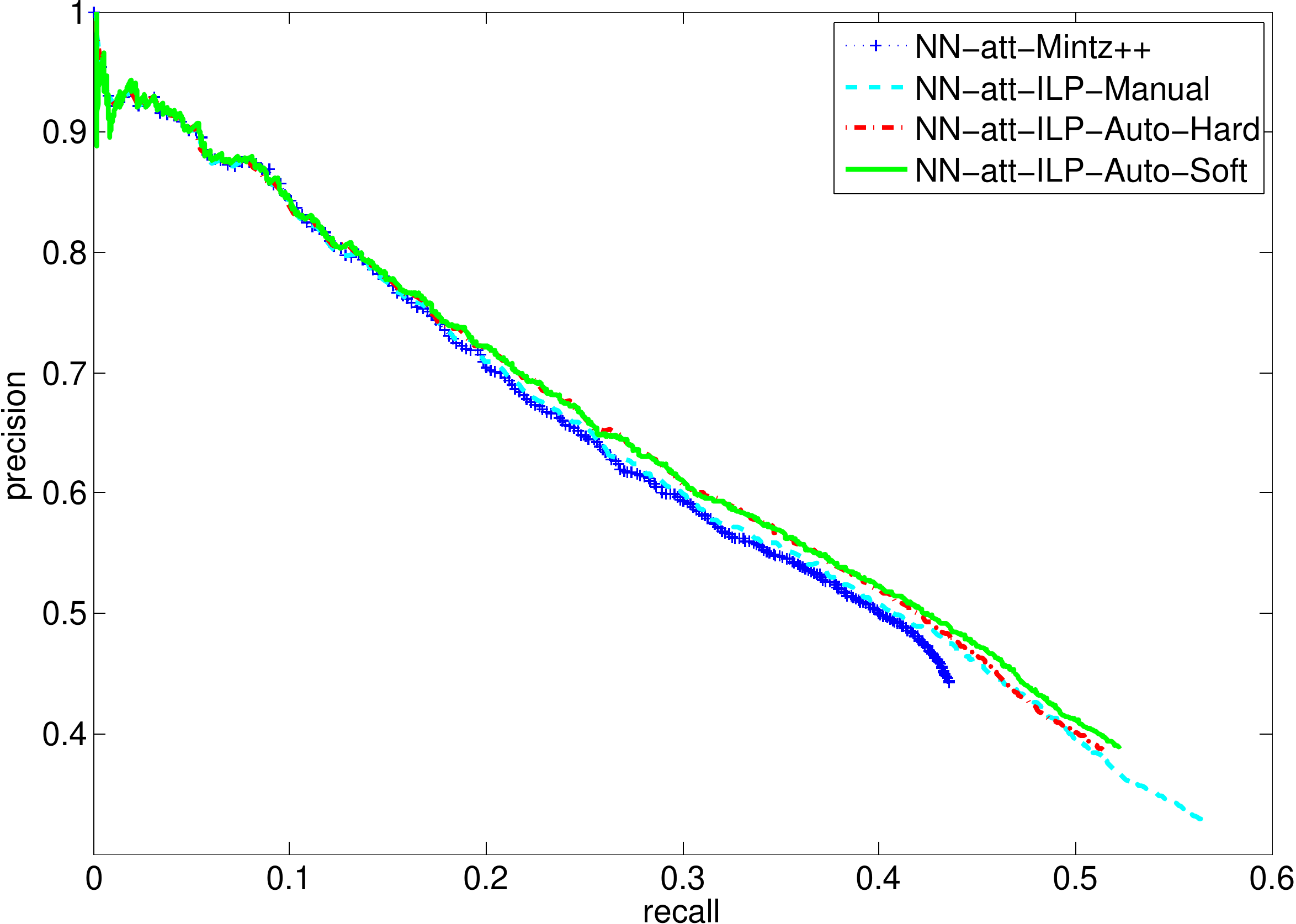}}
\subfigure[The Chinese Dataset]{\label{fig:manvsautonn:d} 
\includegraphics[height=4cm]{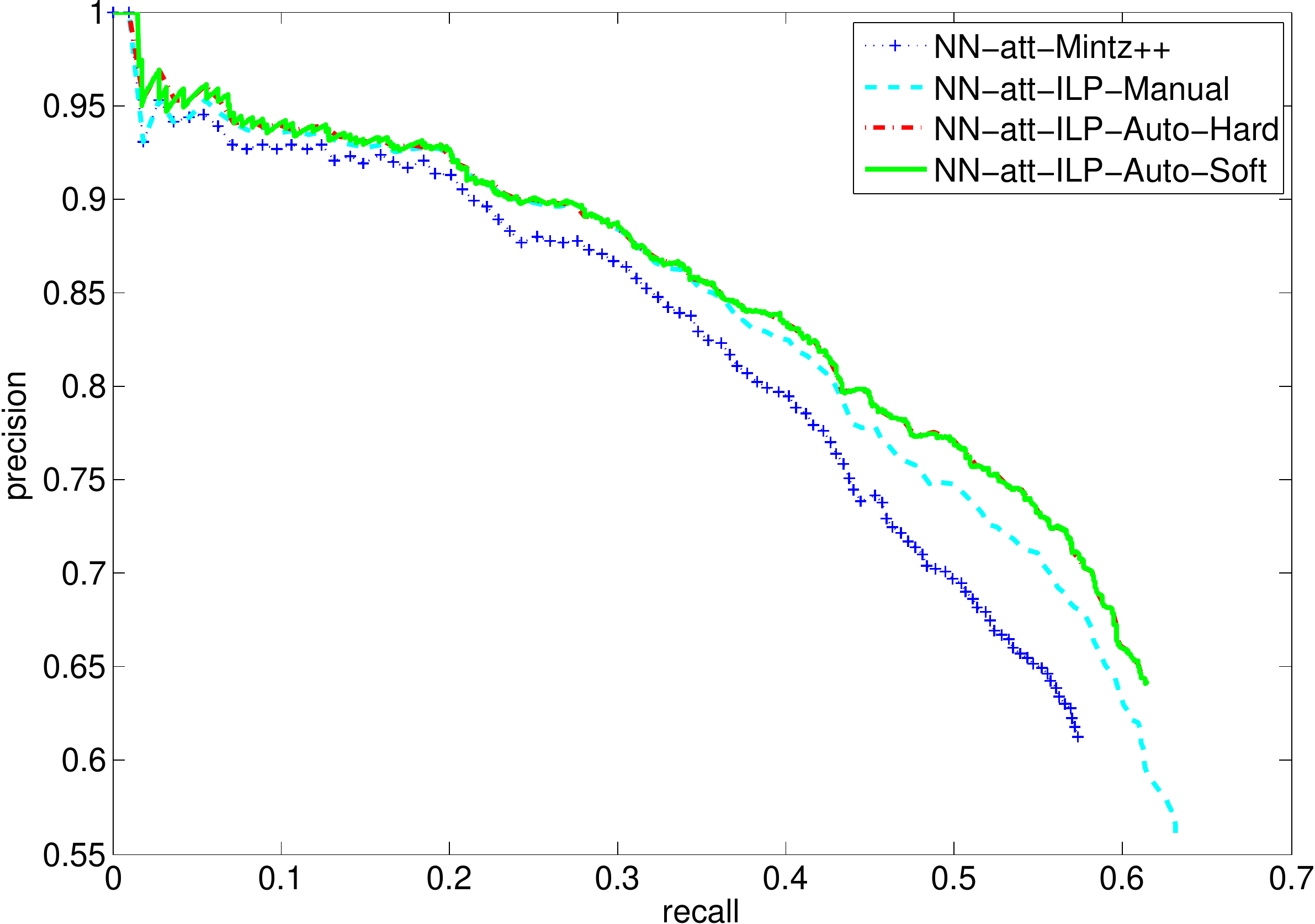}}
\caption{Performance of automatic clues and manual clues with neural network based extractors. }\label{fig:manvsautonn}
\end{figure}

As shown in Figure \ref{fig:manvsautonn},
in both datasets, the automatically collected clues can
improve the NN models better than, or comparably to, the manually obtained ones.
We also notice that in the Chinese dataset, the automatic clues can help \textit{NN-att} to obtain more improvement against manual clues.
After a careful check over the outputs, 
we find that in this dataset, most incorrect predictions 
in the output with manual clues belong to three relations,
where, unfortunately,
the manually obtained clues do not fully cover the clues about those relations, thus  cannot help find enough disagreements to generate useful constraints.
On the other hand, those involved relations are well covered by the automatically obtained clues.
This points out the main shortcoming for manually collected clues that it requires more deep human involvement,
and largely relies on the annotators' expertise.

By comparing the soft and hard style formulations with the automatic clues in Figure~\ref{fig:manvsautonn}, we can see
that both of them can effectively deal with the global inconsistencies among the output of NN extractors, and
perform comparably to each other.


\subsection{Comparing ILP with A Simple Rule-based Solution}
As we can see in many ILP solutions,
with the increasing number of entity pairs and candidate relations, the number of variables and constraints
encoded for ILP will increase dramatically.
As an alternative, one can also design a simple rule-based method to utilize our obtained clues.
For instance, we can design the selection rule as:
\textit{if two or more local predictions conflict with each other, we simply reserve the one with the
highest confidence}.

\paragraph{\textbf{Results on the MaxEnt Extractor}}
We first compare this simple rule-based strategy with ILP using MaxEnt model as the local extractor.
\begin{figure}
\centering
\subfigure[The DBpedia Dataset]{\label{fig:rulevssoft:a} 
\includegraphics[width=4.8cm]{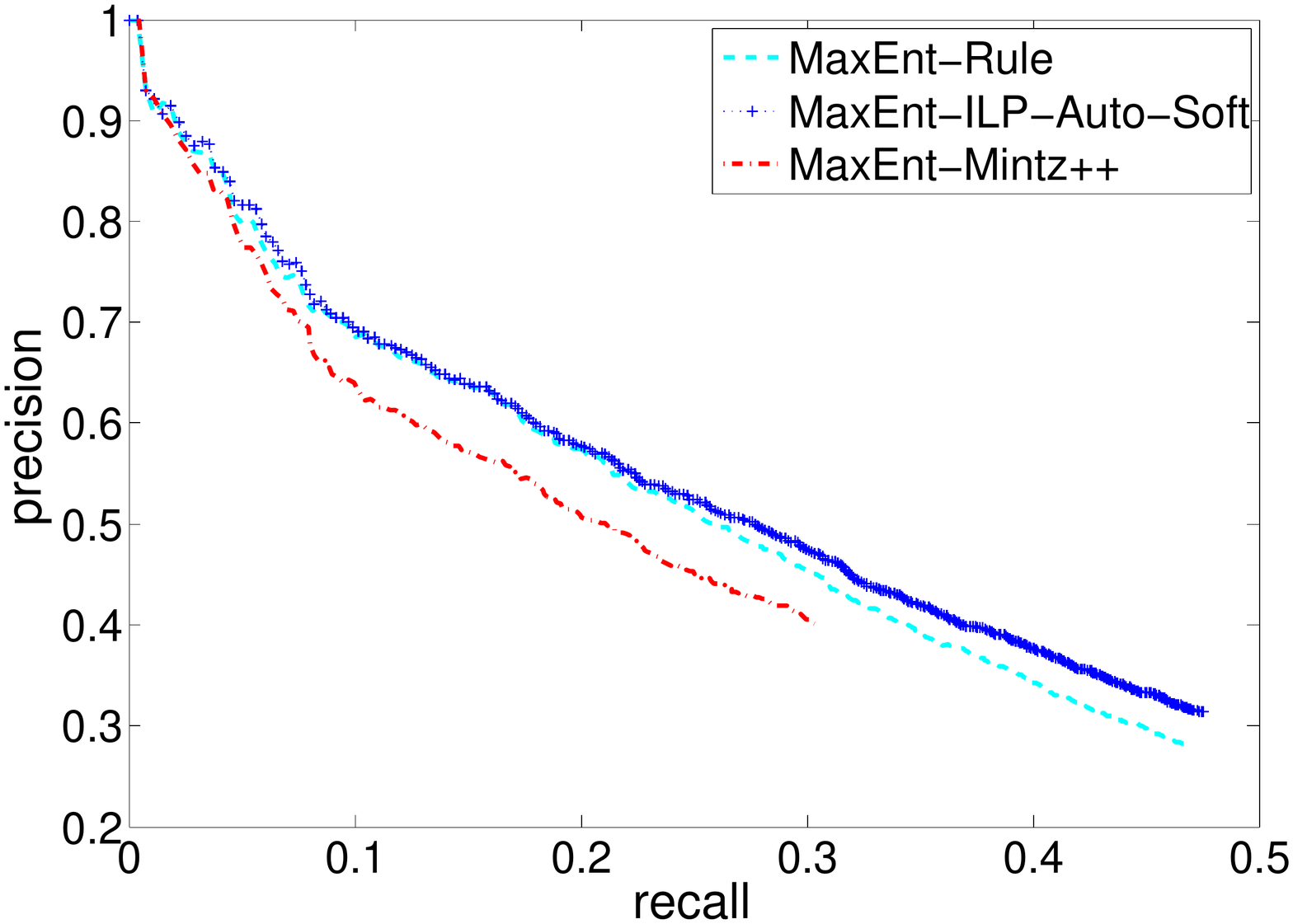}}
\subfigure[The Chinese Dataset]{\label{fig:rulevssoft:b} 
\includegraphics[width=4.8cm]{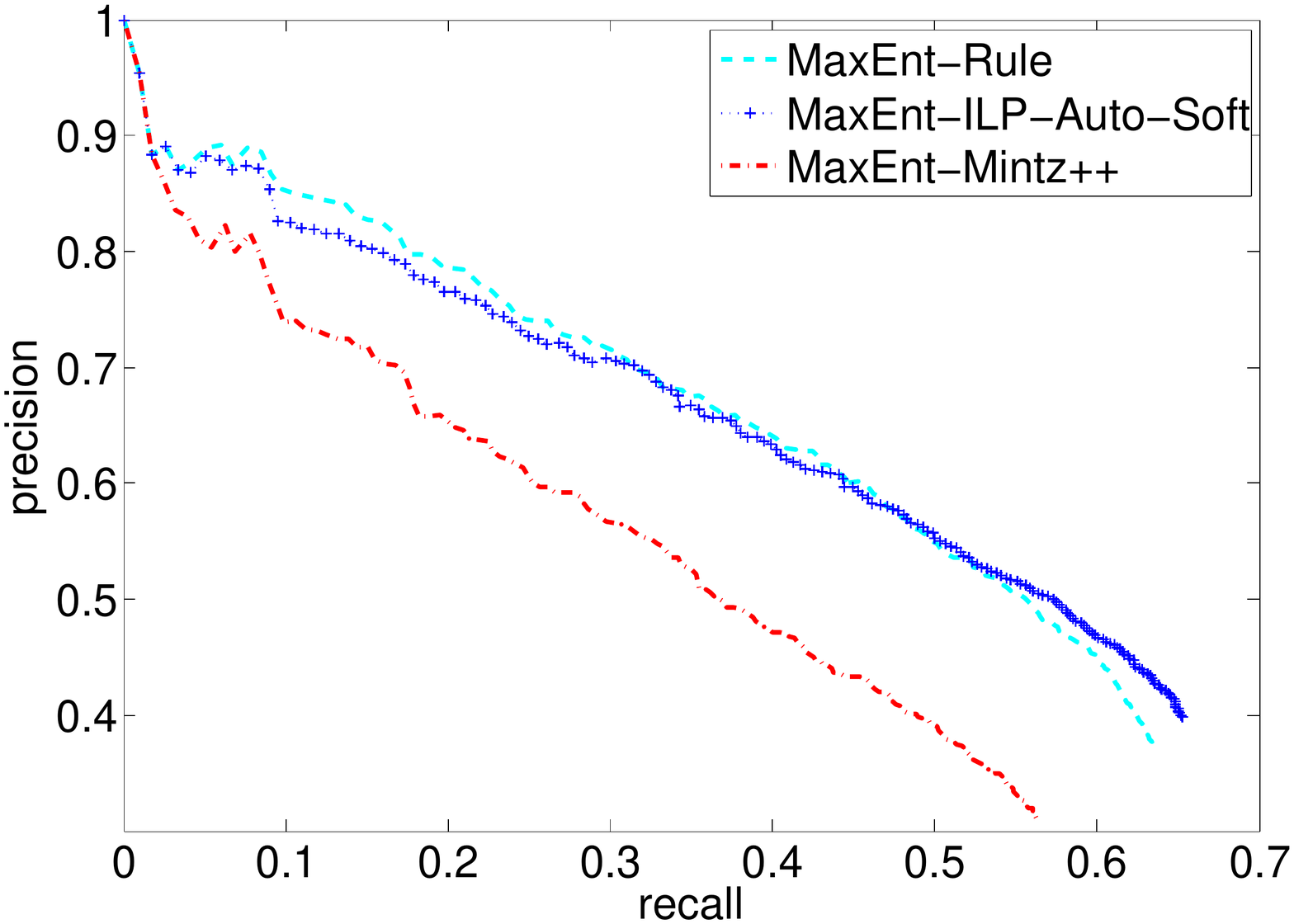}}
\caption{Performance of the simple rule-based strategy and the ILP solver with the MaxEnt extractor.}\label{fig:rulevssoft}
\end{figure}
As we can see in Figure~\ref{fig:rulevssoft}, both the simple rule-based strategy and the ILP framework significantly
improve over the baseline method in both datasets.
And, in the DBpedia dataset, our ILP solution performs slightly better than the rule-based strategy
when the recall is in the low region, and as the recall becomes higher, the improvement from the ILP solution becomes larger.
And, surprisingly, on the Chinese dataset, the rule-based strategy achieves slightly higher precision in the low recall
region, but when the recall becomes higher, the ILP solution obtains better performance again. 


\paragraph{\textbf{Results on the NN Extractor}}
\begin{figure}[h]
\centering
\subfigure[The DBpedia Dataset]{\label{fig:rulevssoftnn:a} 
\includegraphics[height=4cm]{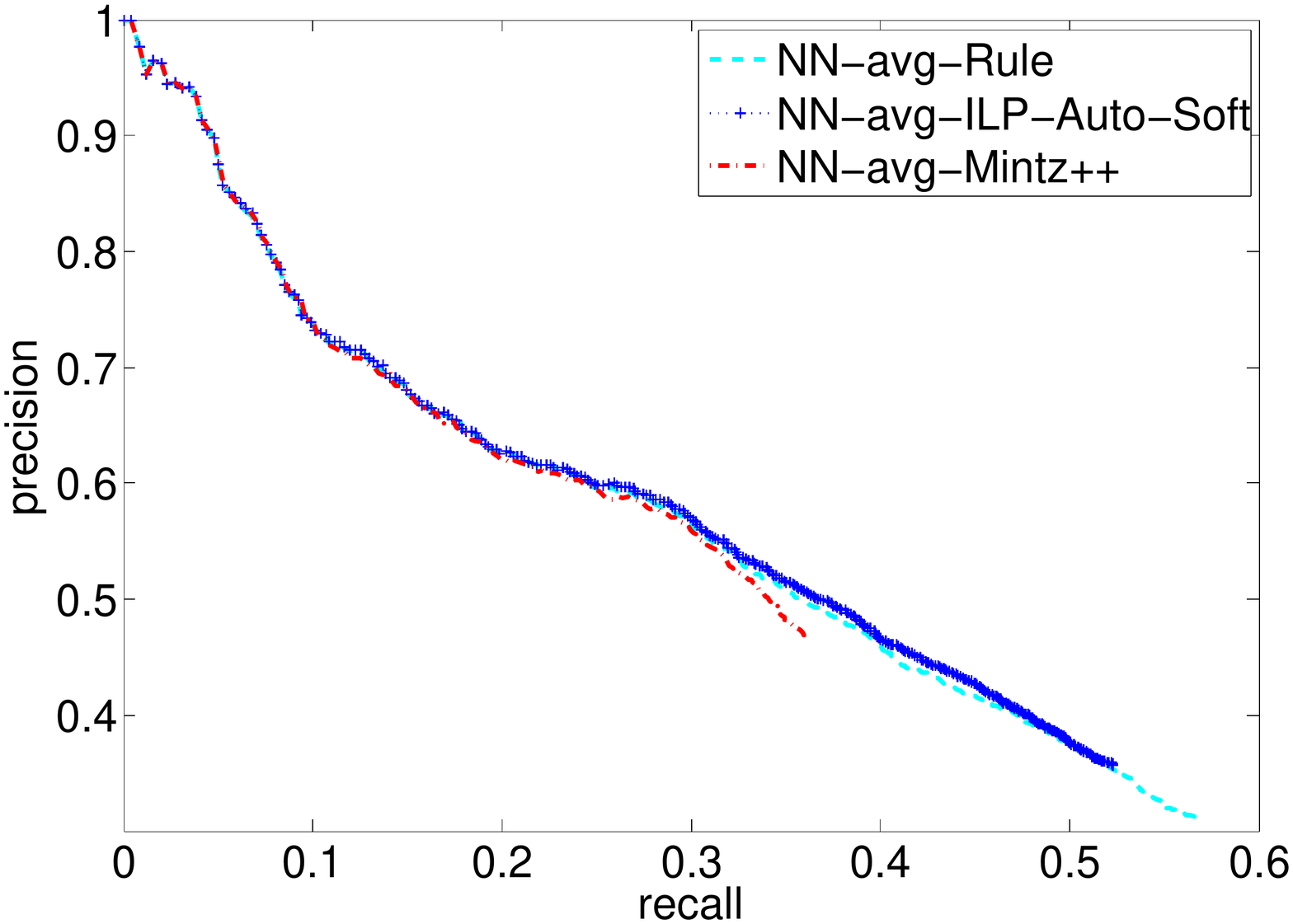}}
\subfigure[The Chinese Dataset]{\label{fig:rulevssoftnn:b} 
\includegraphics[height=4cm]{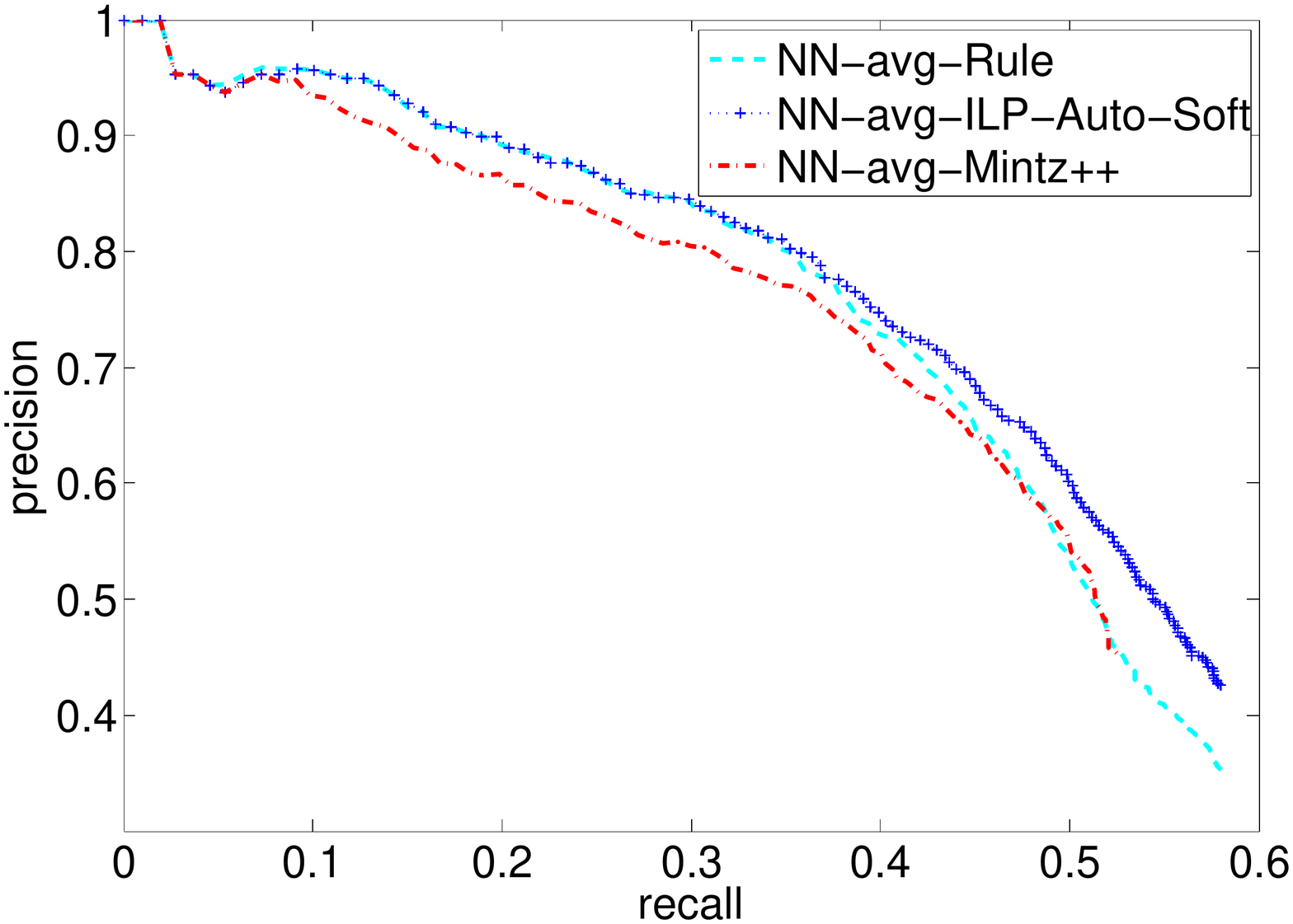}}
\subfigure[The DBpedia Dataset]{\label{fig:rulevssoftnn:c} 
\includegraphics[height=4cm]{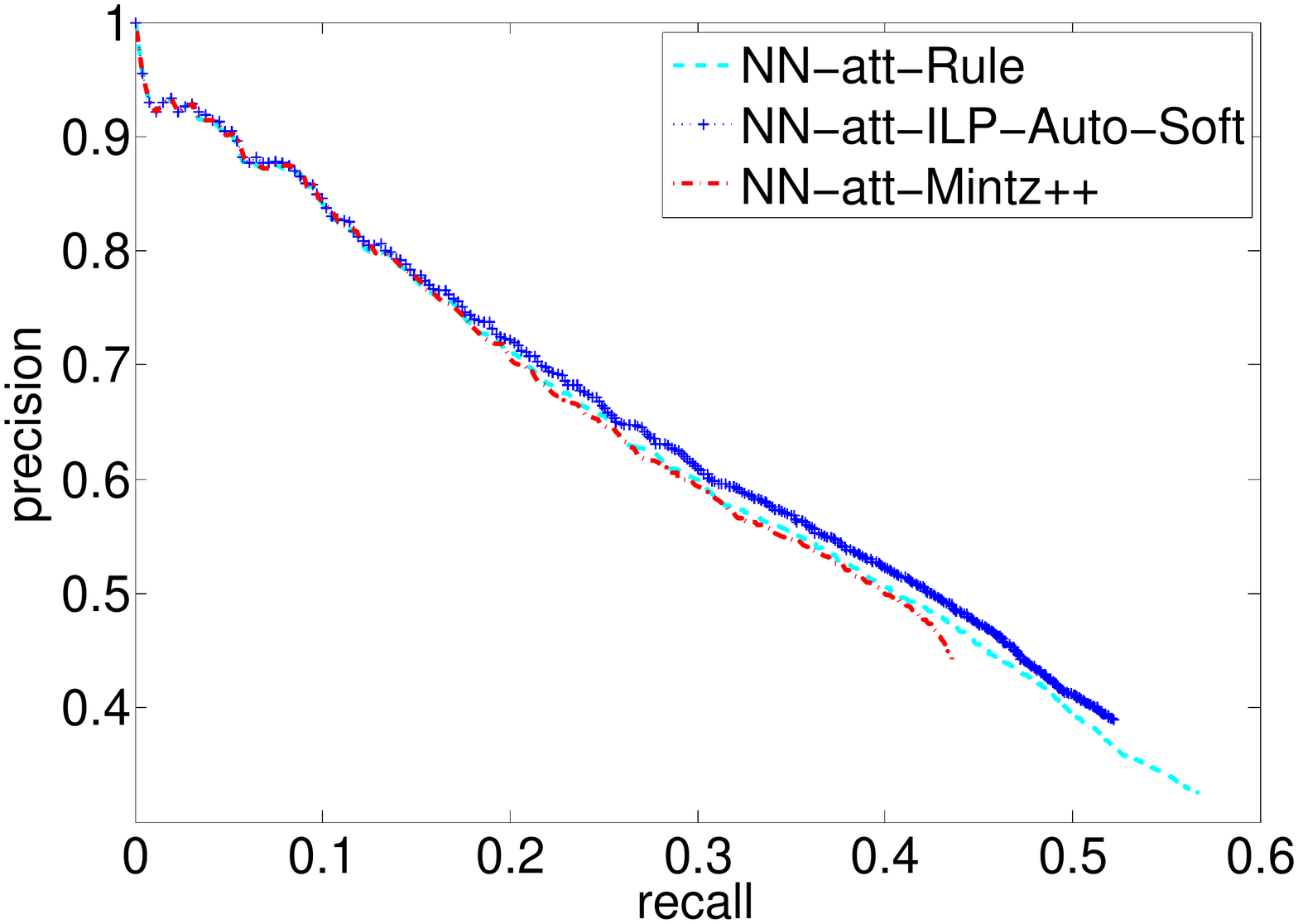}}
\subfigure[The Chinese Dataset]{\label{fig:rulevssoftnn:d} 
\includegraphics[height=4cm]{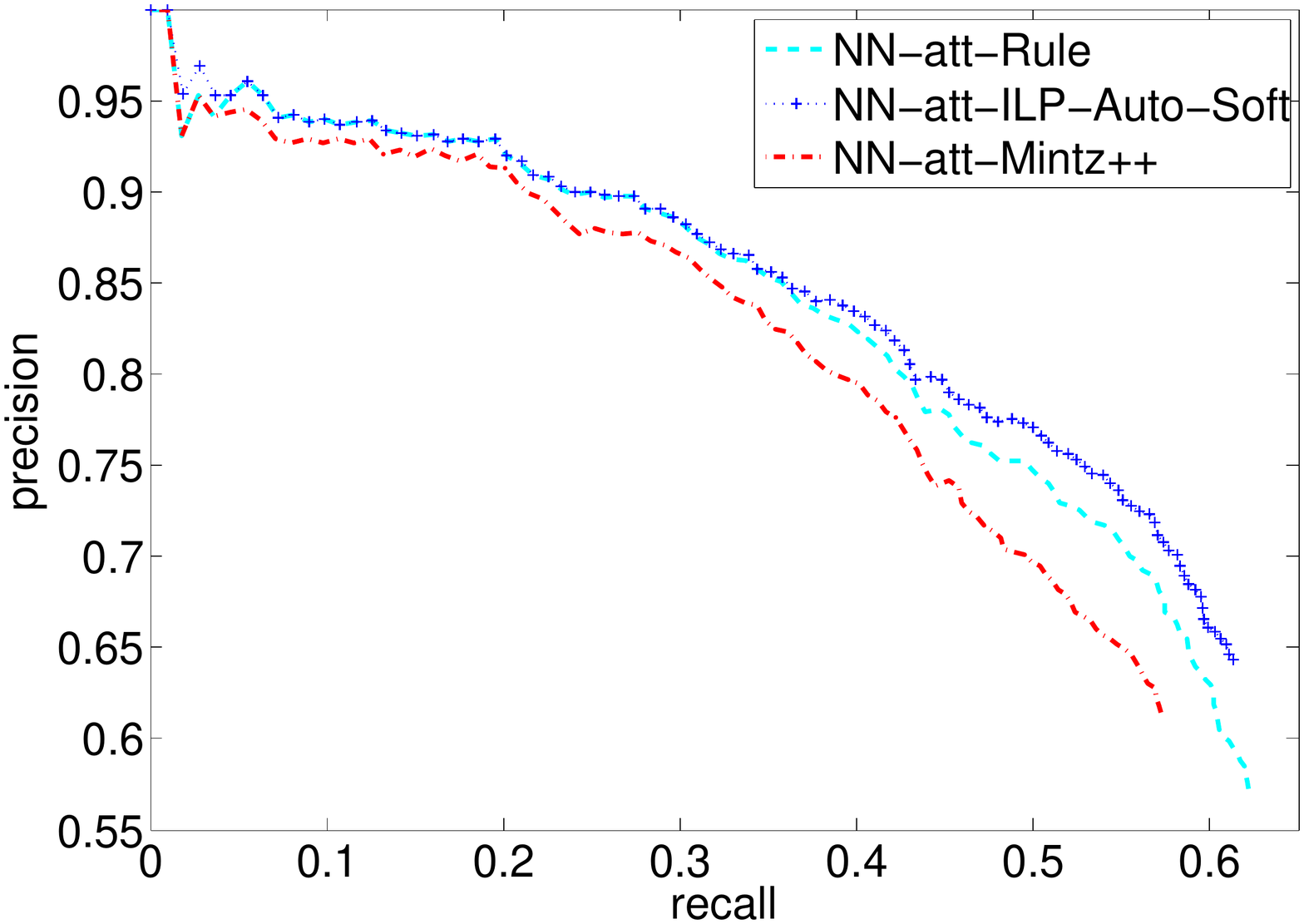}}
\caption{Performance of the simple rule-based strategy and the ILP  solver with the neural network based extractors. }\label{fig:rulevssoftnn}
\end{figure}
As is shown in Figure \ref{fig:rulevssoftnn}, when we use the NN models, both \texttt{avg} and \texttt{att}, as the preliminary extractor, 
on both datasets, our ILP based framework performs better than, or 
comparable to, the rule-based solution. All the results above indicate that compared with 
the simple heuristic technique, our ILP framework 
is more stable and works better in most situations, especially when the problem is 
complex and possibly hard to solve heuristically.

\subsection{Comparing with the State of the Arts}
In previous subsections, we have demonstrate the effectiveness of our framework with different local extractors and the global
clues, collected either manually or automatically. Next, we  compare our framework with both traditional and neural network state-of-the-art extraction models. Considering the fact that
the automatic clues performs consistently better than or comparably to the manual ones, we will use the framework with automatically collected clues for
detailed comparison and analysis.

\paragraph{\textbf{Traditional Extractors}}

\begin{figure}
\centering
\subfigure[The DBpedia Dataset]{\label{fig:autovsstate:a} 
\includegraphics[height=4cm]{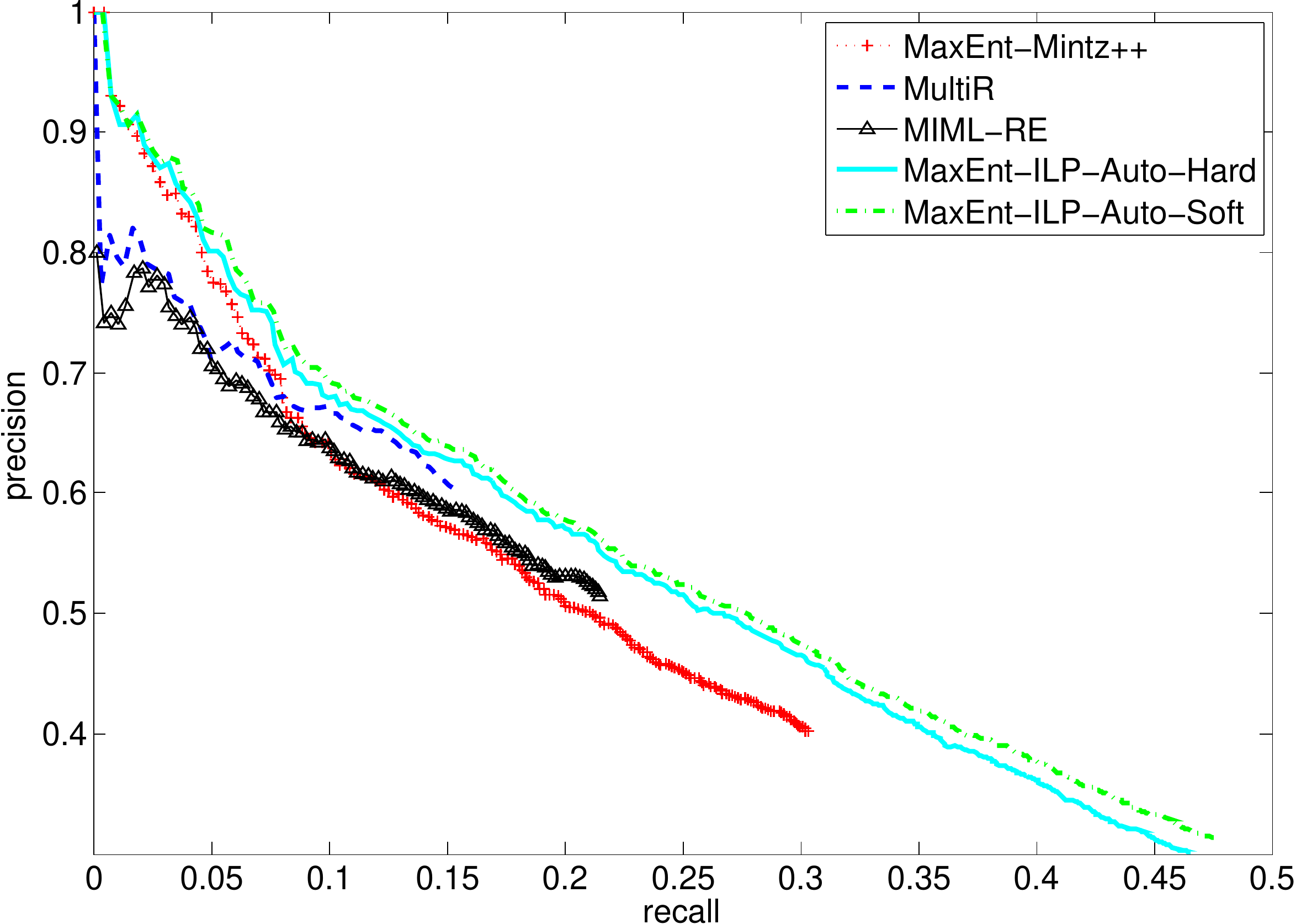}}
\subfigure[The Chinese Dataset]{\label{fig:autovsstate:b} 
\includegraphics[height=4cm]{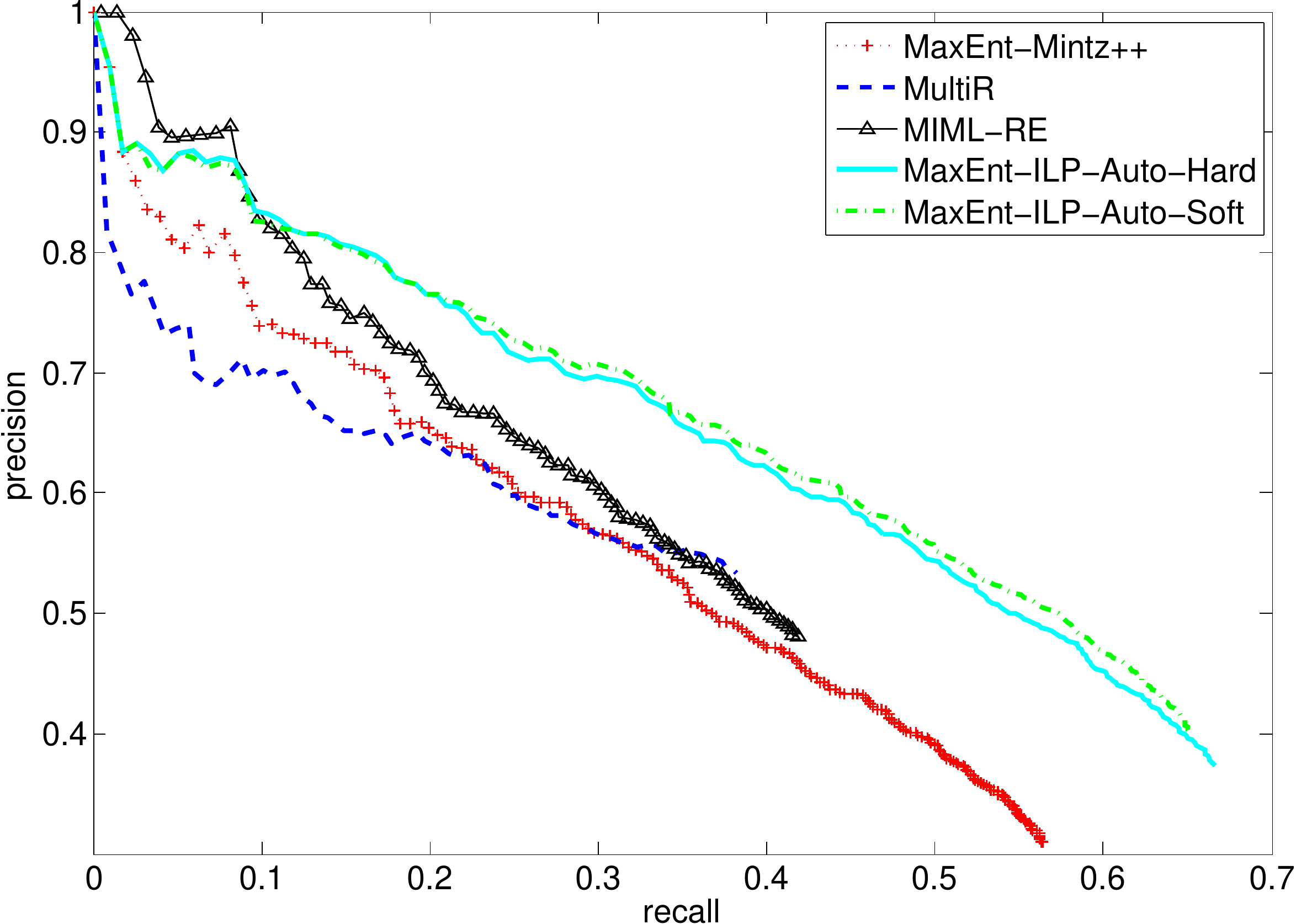}}
\caption{Performance of our framework and traditional state-of-the-art extractors. }\label{fig:autovsstate}
\end{figure}

We first compare our ILP optimized MaxEnt with
the traditional state-of-the-art extractors, MultiR and MIML-RE in Figure~\ref{fig:autovsstate}.

Compared with MultiR, our ILP framework obtains better results
in both datasets. Especially in the Chinese dataset, the precision is improved around 10-16\% at the same recall points.
Our proposed solution performs better compared to MIML-RE in the English
dataset, and outperforms MIML-RE in the Chinese dataset,
except in the low-recall part~($<$10\%) of the P-R curve. This shows that encoding the relation background
information into a simple MaxEnt model can outperform state-of-the-art traditional models.

\paragraph{\textbf{Neural Network Extractors}}

Regarding the neural networks category, we build our solutions by optimizing two sentence-level NN
extractors with our proposed ILP framework, and compare  with 
the \textbf{full} state-of-the-art NN extractors~\cite{linACL2016}. 
\begin{figure}[h]
\centering
\subfigure[The DBpedia Dataset]{\label{fig:autovsstatenn:a} 
\includegraphics[height=4cm]{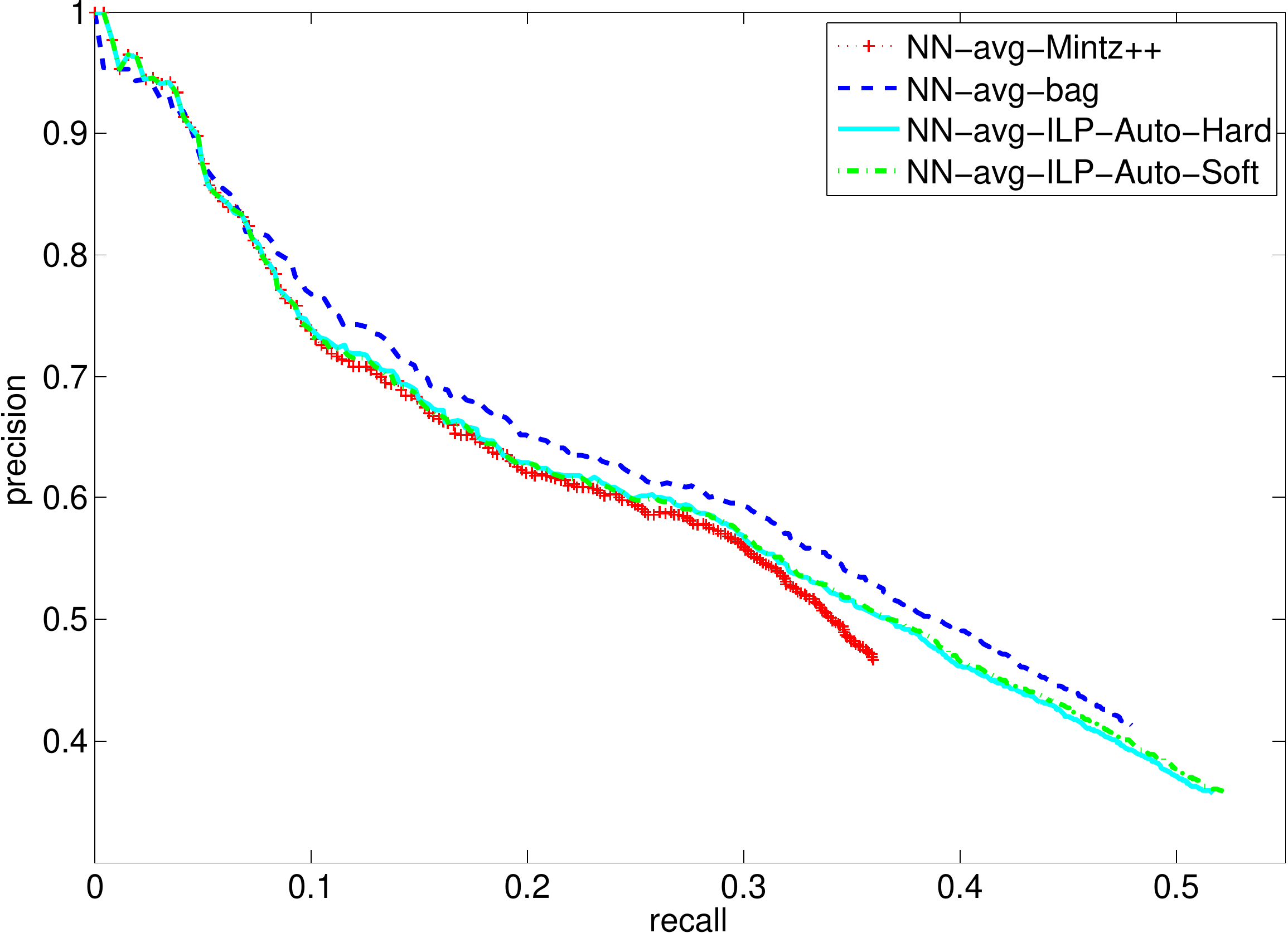}}
\subfigure[The Chinese Dataset]{\label{fig:autovsstatenn:b} 
\includegraphics[height=4cm]{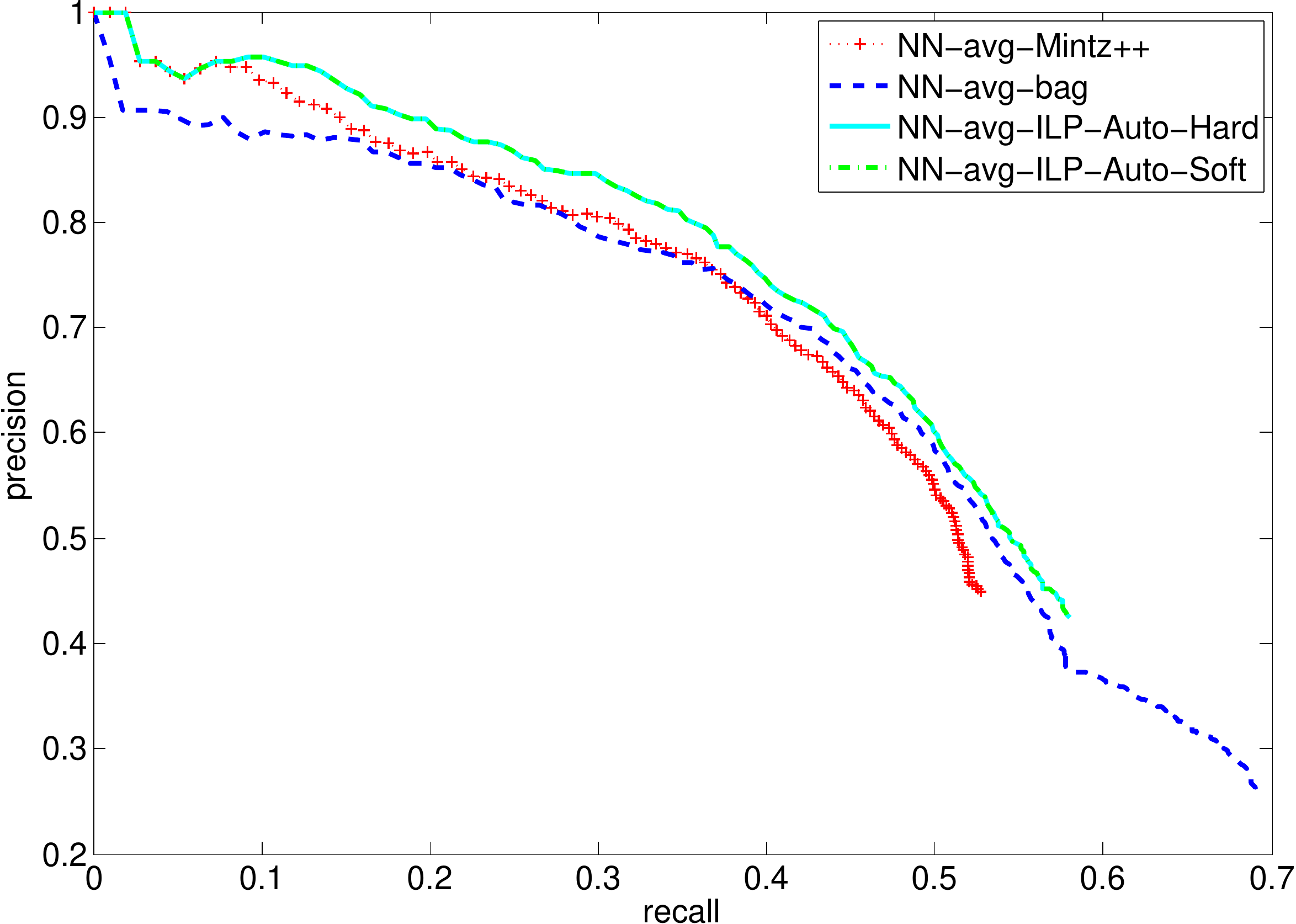}}
\subfigure[The DBpedia Dataset]{\label{fig:autovsstatenn:c} 
\includegraphics[height=4cm]{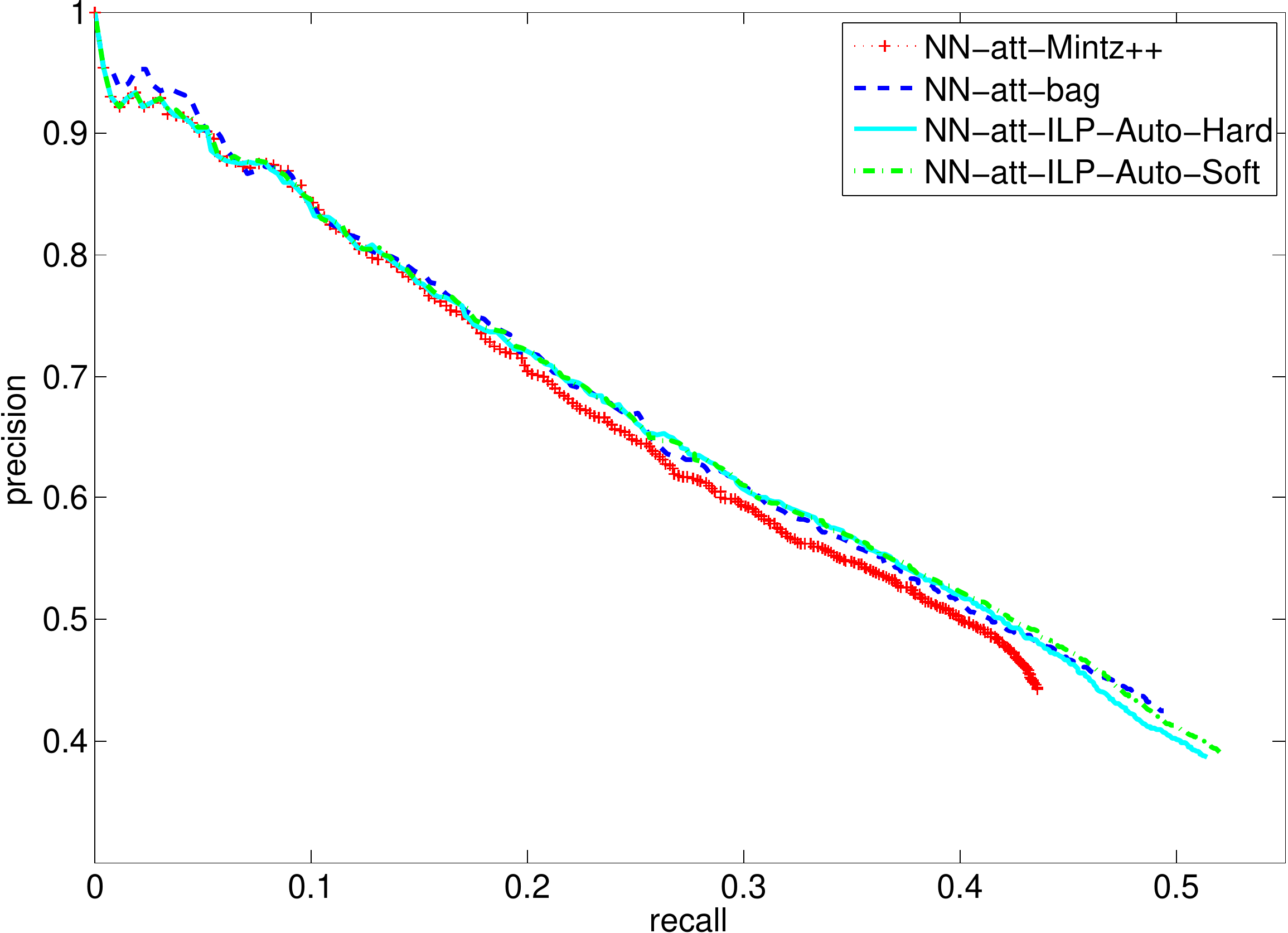}}
\subfigure[The Chinese Dataset]{\label{fig:autovsstatenn:d} 
\includegraphics[height=4cm]{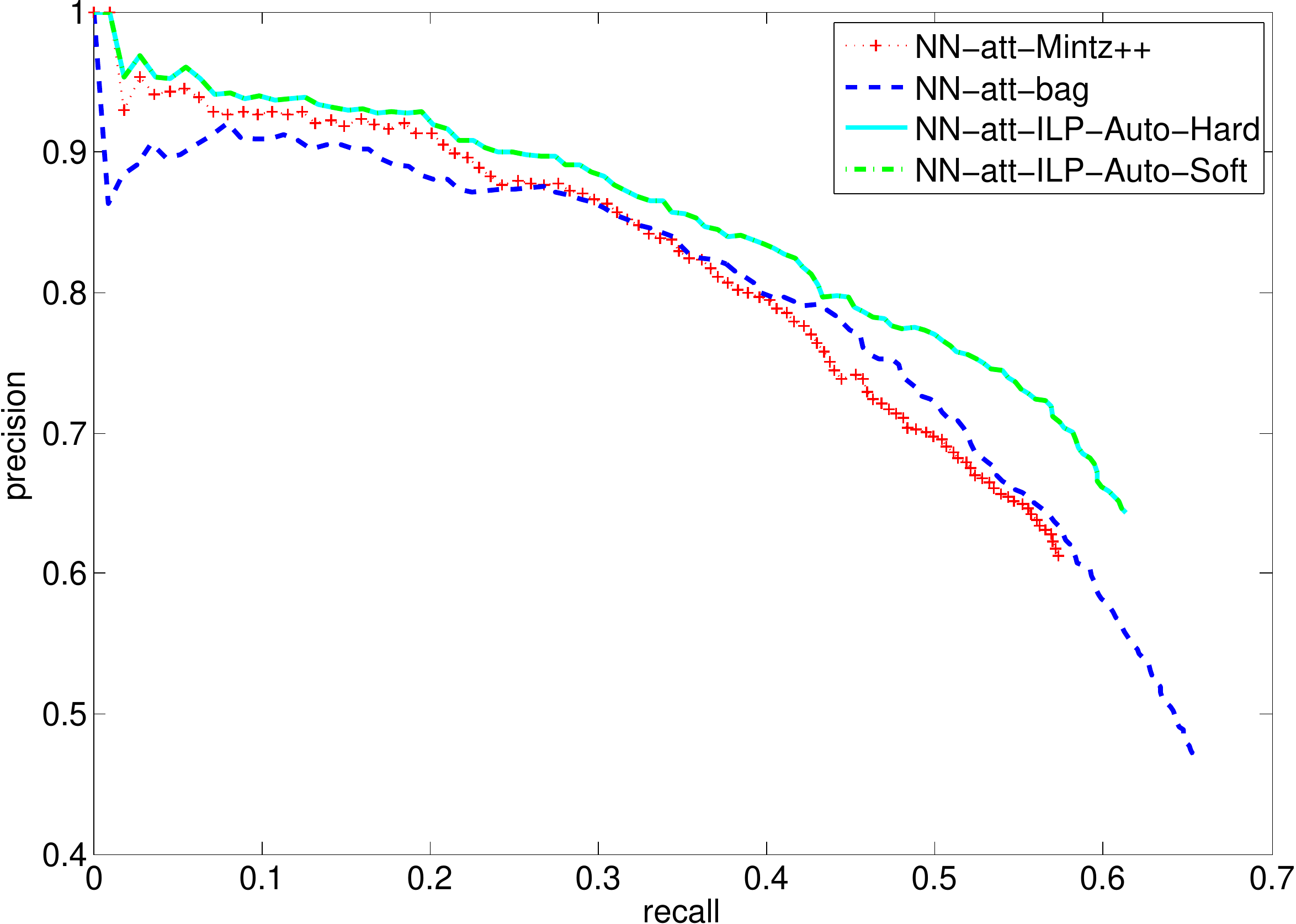}}
\caption{Performance of our framework and the state-of-the-art neural networks based extractors. }\label{fig:autovsstatenn}
\end{figure}

As shown in Figure~\ref{fig:autovsstatenn},
in the Chinese dataset,
our framework outperforms the original full
models, NN-avg-bag and NN-att-bag, significantly on both averaging network or attention network,
while in the DBpedia dataset, our framework performs comparably to the attention network, and
slightly worse than the averaging network. The main difference between our solution and those two state-of-the-art models is that
our framework focuses on exploring the global clues among different entity pairs and eliminating wrong predictions,
while the state-of-the-art NN extractors focus on how to better merge
different sentence-level mentions within an entity pair, e.g., averaging those mentions or attentively weighting them to represent the context between this entity pair. It would be an
interesting direction to investigate how to combine their advantages in a unified framework, which we leave for future work.

\subsection{Impacts of Different Constraints}
Our ILP based framework exploits two kinds of constraints, regarding the
\textbf{argument type inconsistencies} and \textbf{violations of arguments'
uniqueness}, respectively, to address
the global inconsistencies among local predictions, which are designed to implicitly encode relation requirements.
We thus investigate the impact of those constraints used in the ILP framework.

For the sake of clear and concise representation, we  follow \cite{Mihai2012}, to use
the peak F1 score (highest F1 score) as the evaluation criterion.

The baseline, \textit{No-Constraint}, is a variant of our framework without any constraints. It is different from
Mintz++, since we use the approach in Section~\ref{sec_candidate} to summarize the confidence scores of each relation and
obtain the top 3 relations as the final results. It can help us to collect more potentially correct results. Note that
No-Constraint has better or comparable F1 scores compared with Mintz++ in both datasets: in the DBpedia dataset Mintz++ is 34.7\%
and No-Constraint is 35.2\%, while in the Chinese dataset they are both 44.4\%.
As shown in Table~\ref{table_constraint}, in the DBpedia dataset, the
highest F1 score increases from 35.2\% to 38.3\% with the help of both kinds of clues,
while in the Chinese dataset, all constraints contribute to an improvement of  8.4\%. 

In the DBpedia dataset,  the constraints with respect to the implicit
argument type inconsistencies (SR+RO+RER) can improve the F1 score from 35.2\% to 37.7\%,
while in the Chinese dataset, the absolute improvement is as high as 5.0\%. 
The constraints targeting violations of arguments' uniqueness (OU+SU) can increase the F1 score by 1.5\% and
5.1\% on the DBpedia and Chinese dataset, respectively.
Using constraints derived from only one kind of clues can also improve
the performance, but not as well as using both of them.

We also examine how each sub-type of the constraints can improve the results.
As shown in Table~\ref{table_constraint}, the Sub-Rej (SR),
Rel-Obj (RO) and Rel-Entity-Rel (RER) constraints (obtained from the $\mathcal{C}^{sr}$, $\mathcal{C}^{ro}$ and $\mathcal{C}^{rer}$
clues in Section \ref{sec:hard}) can improve the F1 score by 1.1\%, 1.6\%, 1.2\%, respectively, in the
DBpedia dataset, and 0.9\%, 2.2\% and 3.6\%, in the Chinese dataset. As for the two
sub-types obtained according to arguments' uniqueness, the improvements of Obj-Unique (OU) and Subj-Unique (SU)
(obtained from the $\mathcal{C}^{ou}$ and $\mathcal{C}^{su}$
clues in Section \ref{sec:hard}) in the DBpedia and
Chinese datasets are 1.3\% and  0.6\%, 5.2\% and 0.0\%, respectively.
From the results we can see that
almost all sub-types of those constraints can contribute to the improvements
on both datasets.
Note that OU works better than SU, and SR leads to less improvement  compared to RO and ERE. This is mainly because
we have more OU constraints than SU, e.g.,   there are 1,372 OU constraints but only 358 SU constraints in the DBpedia dataset,
and more RO/ERE constraints than SR, e.g.,  we generate about 10,000 SU constraints, 24,000 RO constraints and 70,000 RER constraints in the DBpedia dataset.
This indicates that more constraints may help discover more disagreements among local predictions, thus may yield better results.

\begin{table*}
\centering \caption{\label{table_constraint}Results of different combinations
of constraints on the DBpedia dataset and the Chinese dataset.}
\setlength\tabcolsep{4pt}
\begin{tabular}{c|ccc|ccc}
\hline
\bfseries Method & \multicolumn{3}{c|}{\bfseries DBpedia} & \multicolumn{3}{|c}{\bfseries Chinese} \\
 & \bfseries P(\%) & \bfseries R(\%)  & \bfseries F1(\%) & \bfseries P(\%) & \bfseries R(\%)  & \bfseries F1(\%) \\
\hline
\itshape No-Constraint & 34.1 & 36.3 & 35.2 & 43.3 & 45.7 & 44.4 \\
\itshape Subj-Rel (SR) & 36.5 & 36.1 & 36.3 & 45.1 & 45.6 & 45.3 \\
\itshape Rel-Obj (RO) & \bfseries 37.0 & 36.6 & \bfseries 36.8 & 48.0 & 45.4 & 46.6 \\
\itshape Rel-Entity-Rel (RER) & 35.4 & 37.5 & 36.4 & 46.8 & \bfseries 49.3 & 48.0 \\
\itshape Obj-Unique (OU) & 35.3 & \bfseries 37.8 & 36.5 & \bfseries 50.3 & 48.8 & \bfseries 49.6 \\
\itshape Subj-Unique (SU) & 35.3 & 36.3 & 35.8 & 43.3 & 45.6 & 44.4 \\
\hline
\hline
\itshape SR+RO+RER & \bfseries 36.3 & \bfseries 39.1 & \bfseries 37.7 & 49.7 & \bfseries 49.1 & 49.4 \\
\itshape OU+SU & 35.7 & 37.7 & 36.7 & \bfseries 50.3 & 48.8 & \bfseries 49.5 \\
\hline
\hline
\itshape \bfseries All-Constraints & \bfseries 37.5 & \bfseries 39.1 & \bfseries 38.3 & \bfseries 52.8 & \bfseries 52.9 & \bfseries 52.8 \\
\hline
\end{tabular}
\end{table*}

\subsection{Other Factors} \label{sec_auto}  
Here, we discuss  four main factors in our framework, i.e., the size of the clues,
the number of candidate relations, the threshold for learning the clues, and
the weight of the soft penalty, which will affect the effectiveness or efficiency of our method.
For the sake of brevity, 
here we will take an ILP optimized MaxEnt model with automatically learnt
soft-style constraints (MaxEnt-ILP-Auto-Soft) as an example, and analyze its performance on the DBpedia dataset.

\paragraph{\textbf{The Size of Clues}}

We first investigate the impact of the size of clues. 
In the experiments, we add clues into the framework according
to their related relations' proportions in the local predictions,
and generate constraints based on those clues.
For example, in the DBpedia dataset, the two biggest relations,
\textit{Country} and \textit{birthPlace}, take up about 30\% in the local predictions.
We thus add the clues that are related to these two relations, and then move on to new clues
related to other relations according to those relations' proportions. 


\begin{figure}
\centering
\includegraphics[height=4cm]{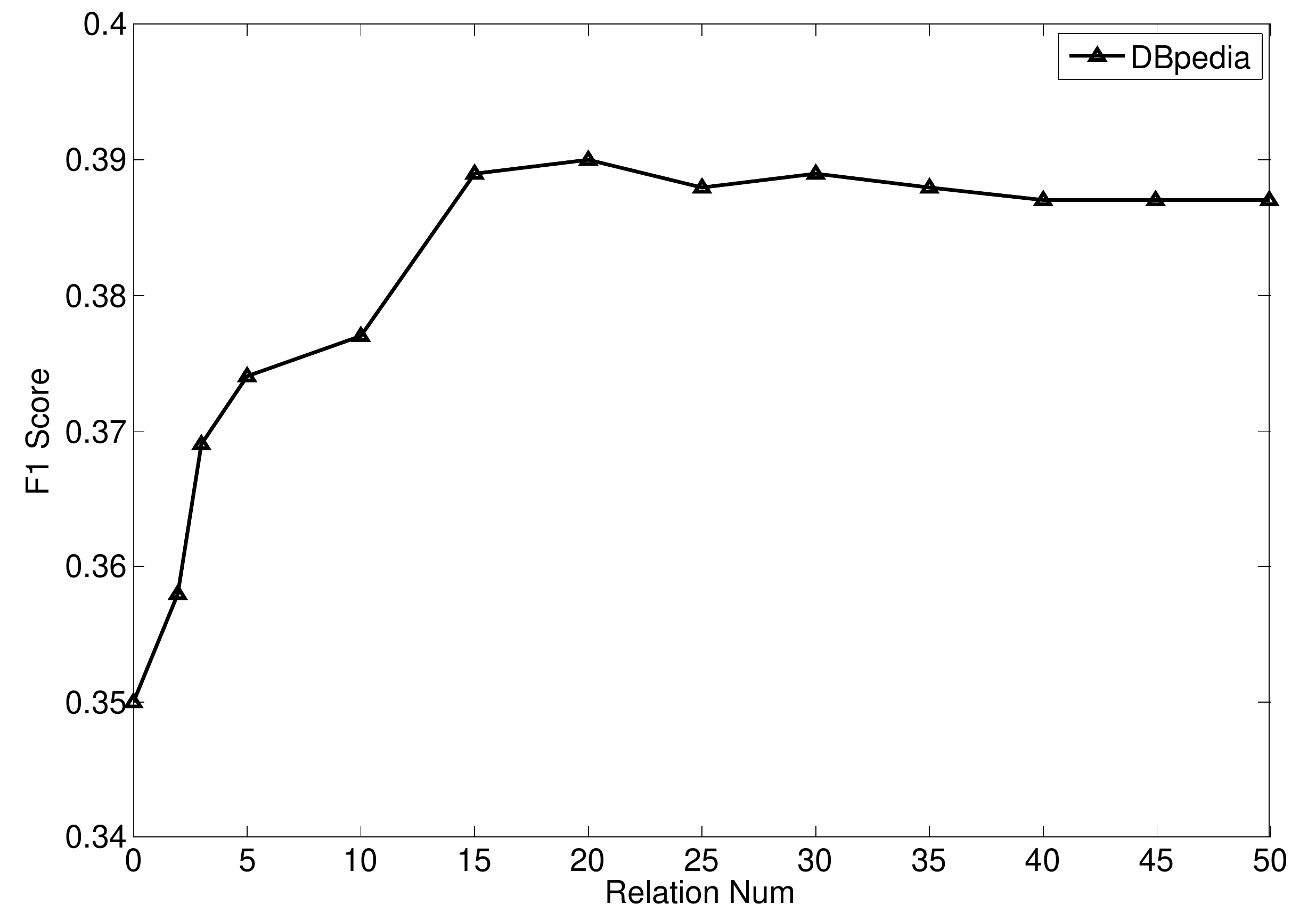}
\caption{The performance of MaxEnt-ILP-Auto-Soft in the DBpedia dataset under different number of relations involved in the constraints. X-axis is the number of relations, and Y-axis is the peak F1 score (highest F1 score).}\label{fig:softrelnum}
\end{figure}

As shown in Figure~\ref{fig:softrelnum}, the clues
related to more local predictions would potentially solve more inconsistencies, thus
are more effective. Adding the first two or three relations improves
the performance significantly, and as more relations are added, the
performances keep increasing until approaching the plateau.



\begin{figure}
\centering
\subfigure[Numbers of decision variables and constraints]{\label{fig:softcons:a} 
\includegraphics[height=4cm]{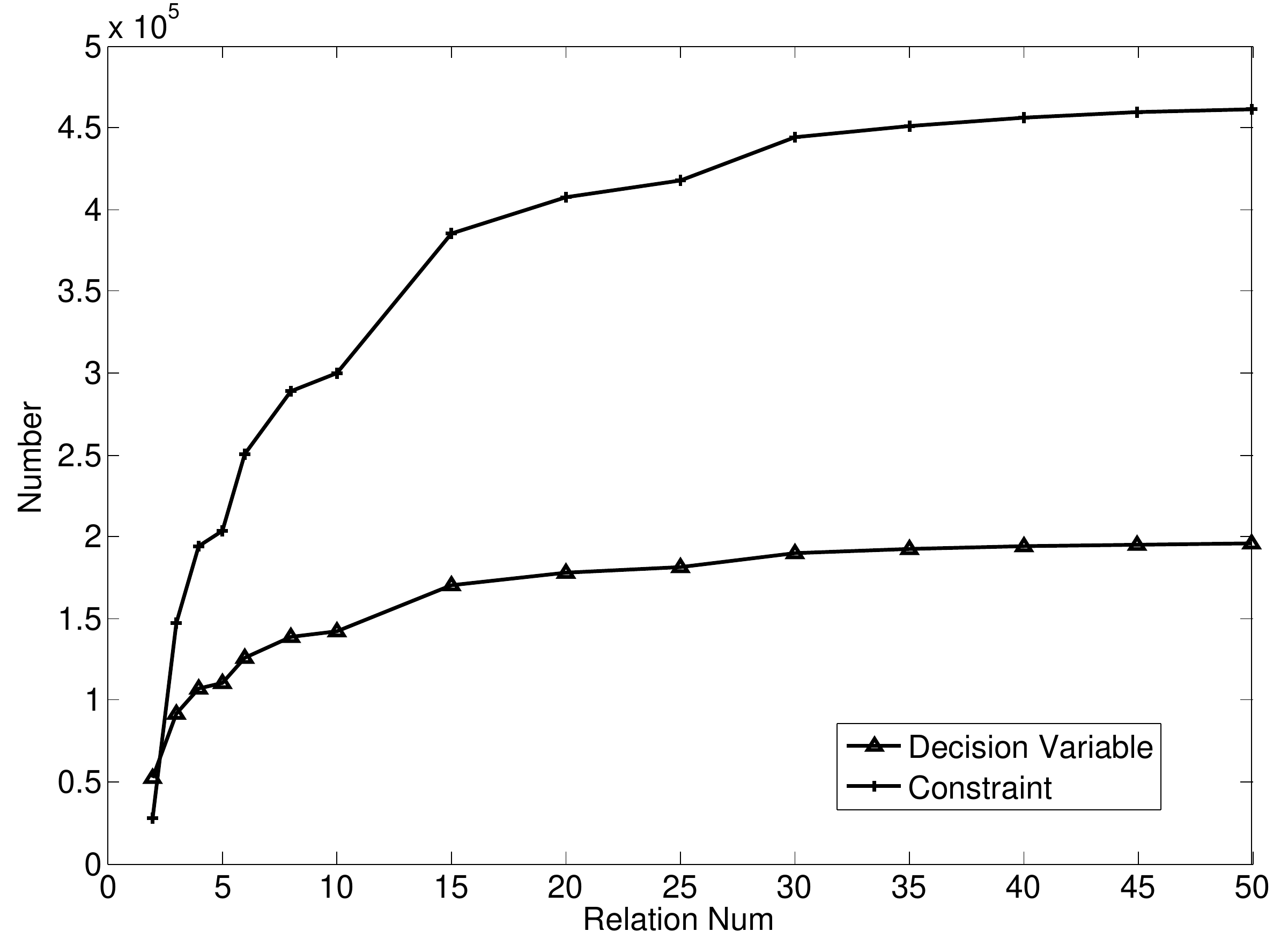}}
\subfigure[The running time of the model]{\label{fig:softcons:b} 
\includegraphics[height=4cm]{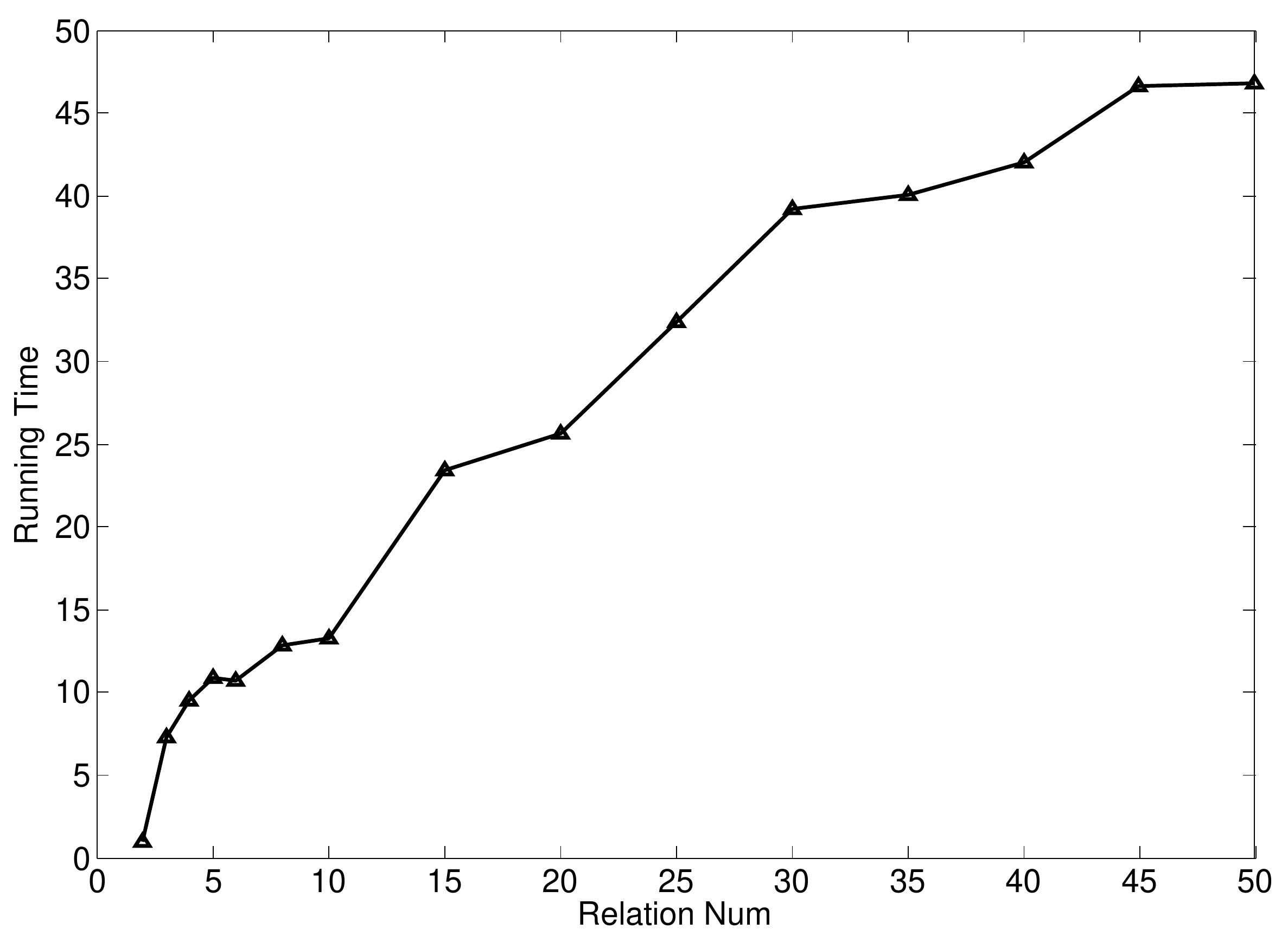}}
\caption{The efficiency of MaxEnt-ILP-Auto-Soft v.s. the number of the relations on the DBpedia dataset. In both (a) and (b), X-axis is the number of the relations. In (a), Y-axis is the number of the decision variables and the constraints, while in (b),  Y-axis is the running time of our model measured in seconds.}\label{fig:softcons}
\end{figure}

A related efficiency issue is that when we introduce more clues,  the numbers of the decision variables and the constraints, as well
as the running time of the model would increase, as illustrated in
Figure~\ref{fig:softcons:a} and Figure~\ref{fig:softcons:b}.
Specifically, the numbers of decision variables and  constraints
  dramatically increase at first, and then with a slower growth. The reason is straightforward:
we first add the relations with large proportions, which may bring more decision variables and
constraints.
As for the running time of the algorithm,
along with  more relations introduced,
the model becomes more complex, and the time cost keeps increasing.
Finally, the increase slows down and the running time almost stays still.
%

From the previous results we can observe that the running time of the optimization model is
mostly affected by the numbers of the constraints and decision variables introduced to the model.
More clues make the model perform better, but less efficient.
In practice, users can trade off between the performance and the efficiency of our framework. For example,
we can see from Figure~\ref{fig:softrelnum}  that
the performance does not improve much after incorporating more than 20 relations, 
which means that we can prune the clues unrelated to the top 20 relations and maintain a reasonable performance.


We also notice that, due to the capacity of the ILP solver, our model  cannot deal with too many
relations at the same time. If there are too many relations involved, e.g., thousands of relations,
there might be billions of constraints generated, and the running time will grow dramatically as well.
One possible solution is to limit the maximum number of  relations processed in one model. Then, how to  properly split thousands of relations into several optimization models to achieve the globally best performance
becomes an other interesting direction for future work.


\paragraph{\textbf{The Number of Candidate Relations}}

In our framework, we select the top $n$ local predictions for each entity pair as the candidates for the following optimization procedure. 
As shown in Figure \ref{fig:softrelnum}, we vary $n$ from 1 to 4 to show its impact to the extraction performance.
\begin{figure}
\centering
\includegraphics[height=4cm]{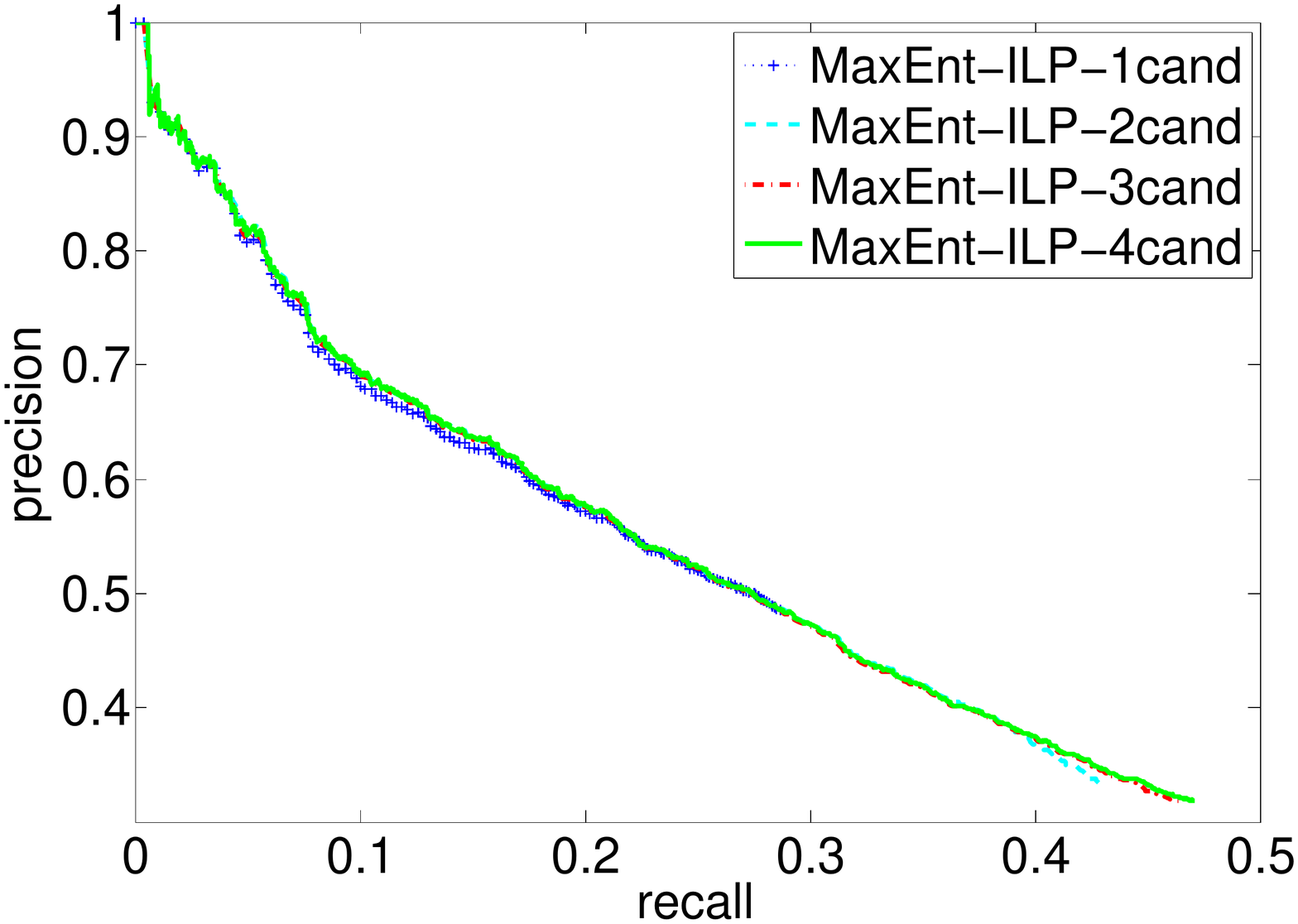}
\caption{The performance of MaxEnt-ILP-Auto-Soft v.s. the number of candidate relations on the DBpedia dataset.}\label{fig:softrelnum}
\end{figure}
We can observe that as $n$ increases from 1 to 3, the precision of our framework does not improve much, but the recall
increases dramatically. This indicates that selecting more candidates may help us collect more potentially correct
predictions. However, obtaining 4 candidate relations can only bring a small  improvement in recall compared to using 3 candidates.  On the contrary, it makes
the model more complex, since that will bring more constraints. To confirm, we investigate the local predictions  in different datasets,
and find that more than 80\% of the correct predictions are among the top 3. Thus, we finally select the top 3 predictions as the candidates
in our framework.

\begin{figure}
\centering
\includegraphics[height=4cm]{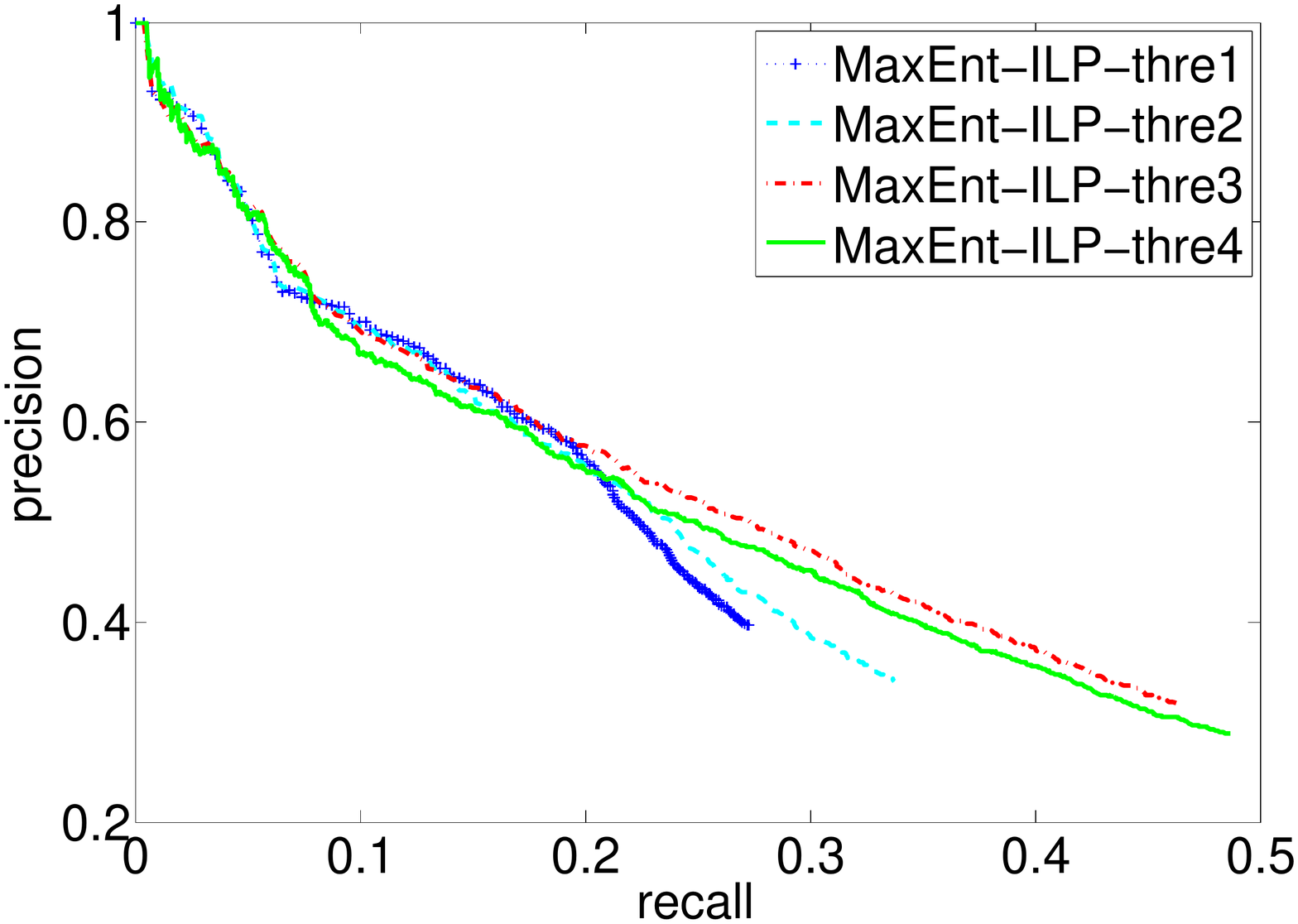}
\caption{The performance of MaxEnt-ILP-Auto-Soft v.s. the threshold of learning the clues on the DBpedia dataset.}\label{fig:autothresh}
\end{figure}

\paragraph{\textbf{The Threshold for Learning the Clues}}
As discussed in Section~\ref{conflict_gen}, when our model automatically collects the clues, it is crucial to set up a 
proper threshold $\kappa$ since it will determine the quality of the obtained constraints.
Intuitively, as the threshold $\kappa$ gets larger, the number of collected clues will increase, but often lead to more constraints of lower quality.
Figure \ref{fig:autothresh} shows the performance of our framework as $\kappa$ varies from -1 to -4 (represented as
MaxEnt-ILP-thre1, MaxEnt-ILP-thre2, MaxEnt-ILP-thre3 and MaxEnt-ILP-thre4). 
When comparing MaxEnt-ILP-thre1 and MaxEnt-ILP-thre2, we can see that their precisions are competitive
within the low-recall region. After the recall is higher than 0.2, the precision of MaxEnt-ILP-thre2 is higher than
MaxEnt-ILP-thre1, and the final recall of MaxEnt-ILP-thre2 is also better than MaxEnt-ILP-thre1.
We think the reason may be that the low quality clues learnt from $\kappa=-1$ often
hurt the correct predictions.
The performance of MaxEnt-ILP-thre3 is better than both MaxEnt-ILP-thre1 and MaxEnt-ILP-thre2, indicating that $\kappa=-3$ can
derive higher quality clues than -1 and -2. As the threshold is set to -4, the precision becomes lower almost at all recall points, since $\kappa=-4$ will lead to much fewer clues than MaxEnt-ILP-thre3, thus can only eliminate fewer incorrect predictions. 

\begin{figure}
\centering
\includegraphics[height=4cm]{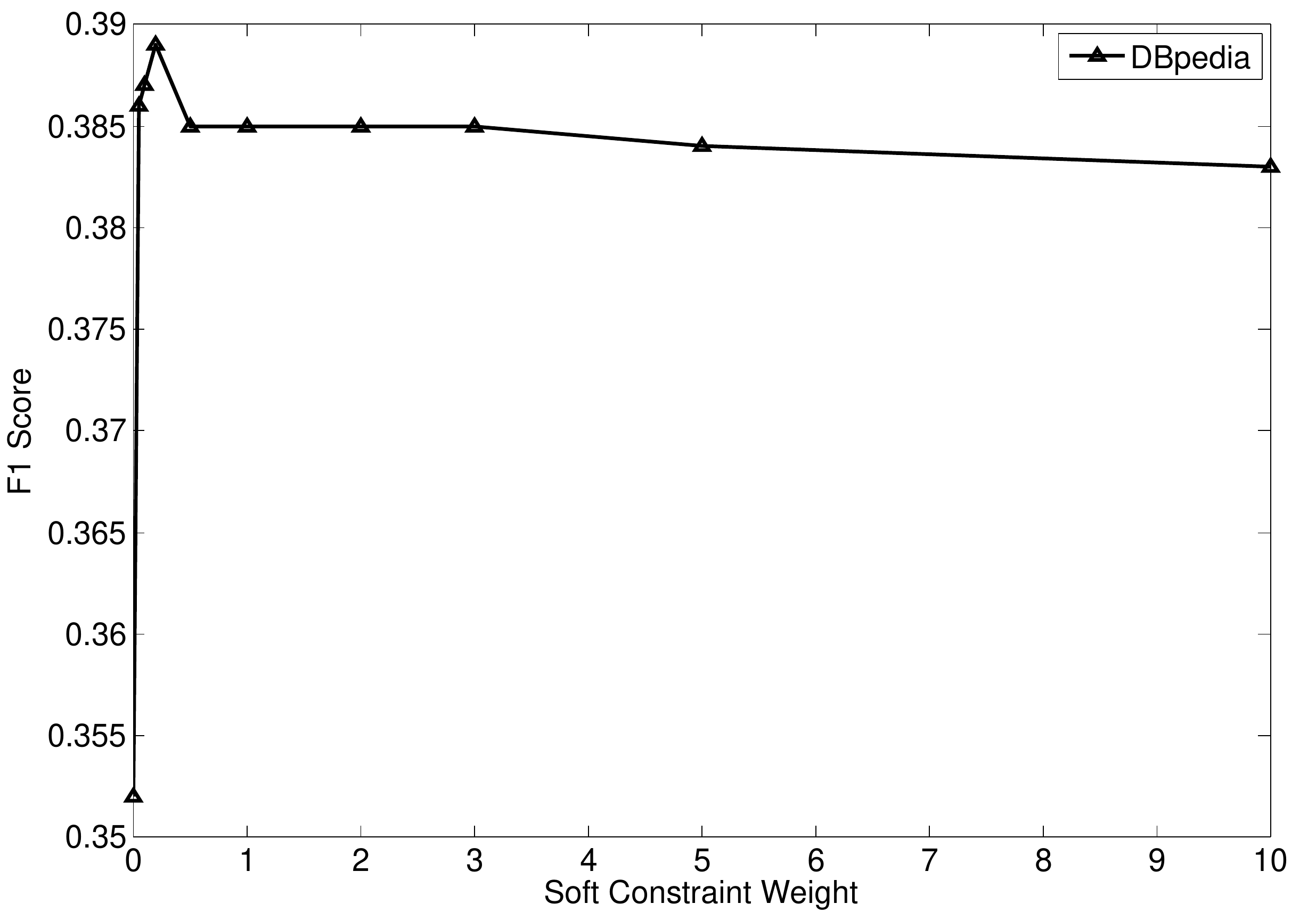} 
\caption{Performance of soft constraints v.s. the weight of the constraints. X-axis is the weight of the soft penalty, while Y-axis is the peak F1 score.}\label{fig:softweight}
\end{figure}

\paragraph{\textbf{The Weight for the Soft Penalty}}
There is a parameter $\alpha$ in the soft style formulation (Equation~\ref{penalty}),
which controls the importance of soft constraints.
Intuitively,
the larger $\alpha$ is, the more penalty we will obtain when we violate
the constraints. When the penalty is large enough, it will be very expensive to violate a constraint, and we will not benefit anything from violating
a constraint. That will  make a constraint work as a hard one. Figure \ref{fig:softweight} shows how
our framework performs as $\alpha$ increases from 0 to 10.

We can see that as $\alpha$ increases from 0, the performance first
improves significantly, and then almost stays still. This indicates the performance of the soft
constraints is not sensitive to the parameter $\alpha$.

\section{Conclusions and Future Work}\label{sec_conclusion}
In this paper, we explore the global clues derived from structured knowledge bases to help
resolve the disagreements among local relation predictions, thus reduce the
incorrect predictions and improve the performance of relation extraction.
Two kinds of clues, including implicit argument type information and argument
cardinality information of relations are investigated.
Those clues can be used to generate constraints in either hard style or
soft style, both of which can be explored effectively in a constrained optimization framework,
e.g., integer linear programming.
Our framework outperforms the state-of-the-art models if we can find such
clues in the knowledge base and they are applicable to the dataset. Furthermore, our framework is scalable for
various local extractors, including traditional models and the
modern neural network models.
Additionally, we show that the clues can be learnt automatically from the KB, and lead
to comparable or better performance to manually refined ones.

In the future, we will investigate how to incorporate new types of clues in our framework, and extend  
our framework to a large-scale relation extraction scenario.
In terms of modeling, we would like to study a unified framework to combine the global clues among different
entity pairs and the local information within an entity pair, which are currently treated in isolation.


\section*{Acknowledgements}

We would like to thank Dong Wang and Kun Xu for their helpful discussions and comments. This work was supported by the National Hi-Tech R\&D Program of China [Grant Number 2015AA015403]; the National Science Foundation of China [Grant Numbers 61672057 and 61672058]; and the IBM Shared University Research Award.

\section*{Appendix: Details about the DBpedia and Chinese Datasets}
\subsection{The DBpedia Dataset}
There are in total 51 relations used to construct our DBpedia dataset, as summarized in Table~\ref{relations_dbpedia}.

\begin{table*}
\centering \caption{\label{relations_dbpedia}The 51 relations used to build the DBpedia dataset.}
\setlength\tabcolsep{4pt}
\begin{tabular}{cccccccc}
\hline
region & residence & birthPlace & primeMinister & influenced \\
child & division & recordLabel & almaMater & deathPlace \\
country & formerTeam & currentPartner & associatedBand & successor \\
keyPerson & knownFor & locationCountry  & stateOfOrigin & product \\
location & coachedTeam & locatedInArea & foundationPerson  & type\\
hometown & leaderName & nationality & subsidiary & owner \\
garrison & team & locationCity  & foundationPlace & partner \\
occupation & vicePresident & regionServed & draftTeam & spouse  \\
 employer & debutTeam & parentCompany & headquarter  & capital \\
state  & president &  owningCompany  & influencedBy & city \\
  \multicolumn{2}{c}{associatedMusicalArtist}  & & &\\
\hline
\end{tabular}
\end{table*}

As discussed in Section~\ref{sec_constraint}, we mainly examine two categories of clues, \textit{implicity argument types inconsistancies} and \textit{violations of aruments' uniqueness}. The former could be roughly divided into three subcategories, $\mathcal{C}^{sr}$, indicating a pair of relations are inconsistent regarding their subjects, $\mathcal{C}^{ro}$, indicating the inconsistency regarding their objects, and $\mathcal{C}^{rer}$ referring the inconsistency in terms of one relation's subject and the other one's object.  The latter one, \textit{violations of aruments' uniqueness}, focused on the argument cardinality requirements for a relation. Specifically, $\mathcal{C}^{ou}$ means a relation expects unique object, while $\mathcal{C}^{su}$ indicates a relation requires unique subject for a given object.

We list the clues manually designed for the DBpedia dataset in Table~\ref{clues_dbpedia}.

\begin{table}
\centering \caption{\label{clues_dbpedia}The manual clues designed for the DBpedia dataset.}
\setlength\tabcolsep{4pt}
\begin{tabular}{c}
\hline
\rowcolor{gray!30}
{\bfseries $\mathcal{C}^{sr}$}\\
\hline
\small $<$country, capital$>$, $<$country, birthPlace$>$, $<$country, \small nationality$>$, $<$capital, deathPlace$>$, \\
\small $<$country, deathPlace$>$, $<$capital, birthPlace$>$,  $<$country, residence$>$, $<$team, location$>$,\\
\small  $<$state, birthPlace$>$, $<$state, deathPlace$>$,
\small $<$state, nationality$>$, $<$leaderName, residence$>$,\\
\small $<$city, hometown$>$, $<$capital, birthPlace$>$, $<$capital, deathPlace$>$,  $<$capital, state$>$\\
\small  $<$team, associatedMusicalArtist$>$, $<$country, hometown$>$\\
\hline
\rowcolor{gray!30}
{\bfseries $\mathcal{C}^{ro}$}\\
\hline
\small $<$country, capital$>$, $<$country, state$>$, $<$country, team$>$, $<$country, city$>$, \\
\small $<$country, state$>$, $<$city, state$>$, $<$nationality, capital$>$, $<$nationality, state$>$, \\
\small $<$nationality, city$>$, $<$city, team$>$, $<$leaderName, team$>$, $<$location, team$>$, \\
\small $<$state, team$>$, $<$almaMater, nationality$>$, $<$almaMater, city$>$, $<$almaMater, country$>$,\\
\small  $<$city, leaderName$>$, $<$nationality, team$>$,  $<$country, locationCity$>$, \\
\small $<$state, locationCity$>$, $<$locationCountry, locationCity$>$, $<$locationCountry, state$>$, \\
\small $<$nationality, locationCity$>$, $<$locationCountry, city$>$\\
\hline
\rowcolor{gray!30}
{\bfseries $\mathcal{C}^{rer}$}\\
\hline
\small  $<$country, country$>$, $<$capital, capital$>$, $<$team, team$>$, $<$country, birthPlace$>$,\\
\small  $<$city, birthPlace$>$, $<$region, birthPlace$>$, $<$country, deathPlace$>$, $<$city, deathPlace$>$, \\
\small $<$country, nationality$>$, $<$nationality, country$>$,  $<$city, region$>$, $<$country, city$>$, \\
\small $<$almaMater, nationality$>$, $<$country, state$>$, $<$city, capital$>$, $<$nationality, country$>$, \\
\small $<$locationCountry, country$>$, $<$country, residence$>$, $<$leaderName, team$>$,\\
\small $<$city, residence$>$, $<$president, team$>$, $<$team, nationality$>$, $<$team, debutTeam$>$\\
\hline
\hline
\rowcolor{gray!30}
{\bfseries $\mathcal{C}^{ou}$}\\
\hline
nationality, country, capital, state, stateOfOrigin, locationCountry\\
\hline
\rowcolor{gray!30}
{\bfseries $\mathcal{C}^{su}$}\\
\hline
capital, subsidiary\\
\hline
\end{tabular}
\end{table}

\subsection{The Chinese Dataset}

\begin{CJK}{UTF8}{gbsn}
As summarized in Table~\ref{relations_chinese}, there are 28 relations used to construct our Chinese dataset.
There are also two categories of clues manually annotated for the Chinese dataset, including 5 subcategories, as listed in Table~\ref{clues_chinese}.

\begin{table}[h]
\centering \caption{\label{relations_chinese}The 28 relations used to construct the Chinese dataset.}

\setlength\tabcolsep{4pt}
\begin{tabular}{cccccc}
\hline
所属洲 & 国籍 & 籍贯  &  著名景点  &   首都  &   所属地区\\
下辖地区  &   名人  &   法人  &   毕业院校  &   创建地点  &   总部所在地\\
 生产厂商  &   执政党  &   政党  &   现任领导人  &   校长  &   导演\\
 知名企业  &   作者  &   出生地  &   知名校友  &   主场  &   代表队员\\
 主演  & 主编    &   品牌  &   上市市场  & & \\
\hline
\end{tabular}
\end{table}

\begin{table}
\centering \caption{\label{clues_chinese}The manual clues designed for the Chinese dataset.}
\setlength\tabcolsep{4pt}
\begin{tabular}{c}
\hline
\rowcolor{gray!30}
{\bfseries $\mathcal{C}^{sr}$}\\
\hline
\small$<$国籍, 下辖地区$>$, $<$国籍, 所属洲$>$, $<$国籍, 首都$>$, $<$国籍, 现任领导人$>$,\\
\small $<$国籍, 下辖地区$>$, $<$国籍, 首都$>$, $<$首都, 籍贯$>$, $<$所属洲, 籍贯$>$, \\
\small$<$著名景点, 国籍$>$, $<$下辖地区, 国籍$>$, $<$所属地区, 国籍$>$, $<$著名景点, 籍贯$>$, \\
\small $<$籍贯, 所属洲$>$, $<$籍贯, 代表队员$>$, $<$籍贯, 执政党$>$, $<$籍贯, 所属地区$>$, \\
\small $<$国籍, 执政党$>$, $<$国籍, 创建地点$>$, $<$所属洲, 创建地点$>$, $<$著名景点, 创建地点$>$\\
\hline
\rowcolor{gray!30}
{\bfseries $\mathcal{C}^{ro}$}\\
\hline
\small $<$国籍, 所属洲$>$, $<$国籍, 毕业院校$>$, $<$国籍, 首都$>$, $<$国籍, 现任领导人$>$, \\
\small $<$国籍, 所属洲$>$, $<$著名景点, 所属洲$>$, $<$首都, 所属洲$>$, $<$籍贯, 毕业院校$>$, \\
\small $<$籍贯, 所属洲$>$, $<$首都, 毕业院校$>$, $<$执政党, 现任领导人$>$, $<$校长, 主场$>$, \\
\small $<$品牌, 主演$>$, $<$上市市场, 主演$>$, $<$校长, 导演$>$, $<$代表队员, 校长$>$,\\
\small  $<$代表队员, 导演$>$, $<$主演, 校长$>$, $<$生产厂商, 执政党$>$, $<$政党, 首都$>$\\
\hline
\rowcolor{gray!30}
{\bfseries $\mathcal{C}^{rer}$}\\
\hline
\small $<$所属洲, 所属洲$>$, $<$所属洲, 国籍$>$, $<$所属洲, 著名景点$>$, $<$所属洲, 首都$>$, \\
\small $<$所属洲, 所属地区$>$, $<$所属洲, 下辖地区$>$, $<$国籍, 国籍$>$, $<$国籍, 著名景点$>$, \\
\small $<$国籍, 所属地区$>$, $<$国籍, 下辖地区$>$, $<$著名景点, 所属洲$>$, $<$著名景点, 国籍$>$, \\
\small $<$著名景点, 著名景点$>$, $<$著名景点, 首都$>$, $<$著名景点, 所属地区$>$,  $<$首都, 所属洲$>$\\
\small  $<$著名景点, 下辖地区$>$, $<$首都, 国籍$>$, $<$首都, 首都$>$, $<$首都, 下辖地区$>$\\
\hline
\hline
\rowcolor{gray!30}
{\bfseries $\mathcal{C}^{ou}$}\\
\hline
\small 所属洲, 首都, 生产厂商, 主场, 政党, 创建地点, 国籍, 籍贯\\
\hline
\rowcolor{gray!30}
{\bfseries $\mathcal{C}^{su}$}\\
\hline
\small  首都, 执政党, 现任领导人, 品牌\\
\hline
\end{tabular}
\end{table}

\clearpage
\end{CJK}

\section*{References}

\bibliography{WPS_ref}  

\end{document}